\documentclass[%
 reprint,
superscriptaddress,
 aps,
 prx,
]{revtex4-2}
\usepackage{bm, bbold}
\usepackage{bbm}
\usepackage{subfigure}

\usepackage{sansmath}
\usepackage{tikz,pgfplots,pgfplotstable}
\usepackage[oldvoltagedirection]{circuitikz}
\usepackage{graphicx}
\usepackage{siunitx,amsmath,amssymb}
\usepackage[unicode,pdfusetitle,colorlinks,allcolors=blue]{hyperref}
\usepackage[capitalise]{cleveref}
\usepackage[english]{babel}
\usepackage[version=4]{mhchem}
\usepackage{float}
\usepackage{xcolor}
\usepackage{MnSymbol}

\usepackage{macros}

\makeatletter
\adddialect\l@en\l@english
\makeatother

\definecolor{fuchsia}{RGB}{255,0,255}

\definecolor{orang}{RGB}{255,130,0}

\usetikzlibrary{external,shapes,arrows,fit,positioning}
\usepgfplotslibrary{groupplots,colorbrewer}
\pgfplotsset{compat=1.18}
\pgfplotsset{cycle list/RdBu}
\pgfplotsset{cycle list/Set1}
\pgfplotsset{cycle list/Set2}
\pgfplotsset{cycle list/Paired}
\pgfplotsset{/pgfplots/layers/sensiblelayers/.define layer set={axis background, axis grid, main, axis ticks, axis lines, axis tick labels, axis descriptions, axis foreground}{/pgfplots/layers/standard}}

\pgfplotsset{
    colormap={spectral11}{
        rgb255=(158,1,66)
        rgb255=(213,62,79)
        rgb255=(244,109,67)
        rgb255=(253,174,97)
        rgb255=(254,224,139)
        rgb255=(255,255,191)
        rgb255=(230,245,152)
        rgb255=(171,221,164)
        rgb255=(102,194,165)
        rgb255=(50,136,189)
        rgb255=(94,79,162)
    }
}

\pgfplotsset{
    colormap={thermal}{
        rgb255=(  4, 35, 51)
        rgb255=(  5, 41, 67)
        rgb255=(  8, 46, 84)
        rgb255=( 14, 49,103)
        rgb255=( 25, 51,124)
        rgb255=( 39, 52,145)
        rgb255=( 57, 51,158)
        rgb255=( 73, 54,159)
        rgb255=( 86, 59,156)
        rgb255=( 98, 65,152)
        rgb255=(109, 70,148)
        rgb255=(120, 75,145)
        rgb255=(131, 80,143)
        rgb255=(142, 84,140)
        rgb255=(153, 88,137)
        rgb255=(165, 92,134)
        rgb255=(177, 95,130)
        rgb255=(189, 99,124)
        rgb255=(201,103,117)
        rgb255=(212,107,109)
        rgb255=(223,112,100)
        rgb255=(232,119, 90)
        rgb255=(240,127, 80)
        rgb255=(246,136, 72)
        rgb255=(249,147, 65)
        rgb255=(251,159, 61)
        rgb255=(252,172, 60)
        rgb255=(251,185, 62)
        rgb255=(249,198, 65)
        rgb255=(246,211, 71)
        rgb255=(242,225, 77)
        rgb255=(237,238, 84)
        rgb255=(232,250, 91)
    }
}

\tikzstyle{every picture}+=[font=\small\sffamily\sansmath]
\pgfplotsset{tick style={semithick,black}}
\newcommand{\product}[1]{\texttt{#1}}
\newcommand{\tnode}{thermodynamic neuron}

\definecolor{leftColor}{RGB}{94,79,162}
\definecolor{rightColor}{RGB}{244,109,67}

\definecolor{tiltColor}{RGB}{94,79,162}
\definecolor{barrierColor}{RGB}{102,194,165}
\definecolor{readoutColor}{RGB}{244,109,67}

\definecolor{joos}{RGB}{0,242,242}

\definecolor{jeremybejanin}{RGB}{242,0,0}

\begin{document}

\title{
A Blueprint for Equilibrium-Based Differentiable\\ Continuous-Variable Thermodynamic Computing}


\author{Owen Lockwood}
\email{owen@extropic.ai}
\affiliation{Extropic Corp., San Francisco, California 94111, USA}
\author{Jérémy Béjanin}
\affiliation{Extropic Corp., San Francisco, California 94111, USA}
\author{Joost Bus}
\affiliation{Extropic Corp., San Francisco, California 94111, USA}
\author{Christopher Chamberland}
\affiliation{Extropic Corp., San Francisco, California 94111, USA}
\author{Patrick Huembeli}
\affiliation{Extropic Corp., San Francisco, California 94111, USA}
\affiliation{Noumenal Labs Inc, Dallas, Texas 75229, USA}
\author{Frank Sch\"afer}
\affiliation{Extropic Corp., San Francisco, California 94111, USA}
\author{Guillaume Verdon}
\email{gill@extropic.ai}
\affiliation{Extropic Corp., San Francisco, California 94111, USA}
\affiliation{Department of Applied Mathematics, University of Waterloo, Ontario N2L 3G1, Canada}

\date{\today}

\begin{abstract}
To help address the escalating energy and latency demands of machine-learning workloads, we introduce a blueprint for an energy-efficient and fast thermodynamic computing stack that leverages stochastic analog processes in physical hardware. In this work, we focus on energy-based thermodynamic computing where the stochastic process is well described by Langevin dynamics with tunable energy potentials. The implementation of such potentials in physical hardware enables us to generate and sample from basic parameterized energy-based models. We demonstrate how to construct and train popular classes of machine learning models based on these hardware-native energy-based models, using the framework of probabilistic graphical models. We analyze the runtime and energy consumption of different models in this thermodynamic paradigm based on theoretical considerations and numerical studies. As a preliminary experimental realization of such hardware, we present our stochastic analog superconducting circuits driven by thermal noise. Together, these results outline a path toward energy-efficient thermodynamic hardware for probabilistic machine learning.
\end{abstract}


\maketitle


\section{Introduction}

In an era of dramatically increasing demand for compute powering machine learning based workloads, the energy and time that digital computers consume have become a problem in need of improvement~\cite{de2023growing, reuther2022ai, luccioni2024power, masanet2020recalibrating}. In this paper, we demonstrate a thermodynamic computing paradigm based on the equilibria of energy functions which undergo Langevin dynamics, and which has potential to impact the computing landscape through increased time and energy efficiency. Thermodynamic computing as a method is distinguished from other kinds of computing in that it harnesses stochastic fluctuations as a core computing resource.

In classical (i.e.~digital and deterministic) and quantum computing, much effort is put into the elimination of random fluctuations from within the system. In the classical case, this is because digital computers are engineered to operate deterministically, and any stochastic effects due to random perturbations must therefore be mitigated through hardware engineering or error correction. Classical computers are designed such that the flow of electrons within their circuits can be systematically mapped to the computation of Boolean functions. The entropy and heat produced by the non-reversible operations in the computer must be pumped out of the system; otherwise, the thermal fluctuations that are produced by computations would disrupt the deterministic, binary logic of the computer. Quantum computers~\cite{mohseni2024build, de2021materials, wendin2017quantum}, similarly, require keeping stochastic fluctuations at bay because such fluctuations cause quantum states to decohere and thereby to become unusable for computation. Quantum computing systems require substantial effort (e.g.~radiation shields, fridges, etc.) to minimize these fluctuations that threaten quantum coherency. Thus, classical and quantum computers, in different ways, depend on the elimination of stochastic fluctuations. In both cases, these require additional energy being used by the system.

Yet in modern algorithms, we often are forced to reintroduce stochastic fluctuations (as they rely on noisy gradients, Monte Carlo estimates, sampling-based inference, etc.), despite having engineered them out of the hardware to the best of our ability. Since modern digital methods must actively counter the thermodynamic nature of computation~\cite{wolpert2019stochastic, wolpert2024stochastic} to enable deterministic computations, this results in decades of research optimized for hardware that fights the physics of nature, rather than embraces it. Fundamentally, this results in an algorithmic lock-in that we believe must break~\cite{hooker2021hardware}. This lock-in is clearly revealed for algorithmic methods that require stochasticity and are naturally suited for physical systems, which are ubiquitous in machine learning fields (e.g.~Markov chain Monte Carlo sampling, diffusion modeling, etc.). Deterministic digital hardware is used for stochastic and machine learning routines, with stochasticity introduced via software,  architecture, and algorithm design, resulting in significant inefficiencies. While generating randomness itself is relatively inexpensive on modern hardware (and not the bottleneck of modern systems, as they have been designed in that way), using deterministic digital systems to represent and manipulate inherently probabilistic distributions is not. As the collapse of informal scaling laws, such as Moore's law, continues to play out, the design strategy of keeping stochasticity at bay cannot last forever. To overcome these limitations, a new computing hardware approach is required; one in which the physics of the substrate aligns with the dynamics of the computation rather than fighting it.

Thermodynamic computing is a paradigm that aims to resolve this incongruity and leverage thermal fluctuations as a resource for stochastic computation~\cite{conte2019thermodynamic}. This presents a way for probabilistic computing to be implemented more naturally in the stochastic dynamics of physical systems. In this approach, similar to quantum computing, measuring the final state gives rise to a distribution rather than a deterministic outcome. However, thermodynamic computing is distinctly different from quantum computing as it intentionally operates at finite temperature and does not seek to exploit quantum properties such as superposition or entanglement. The key feature of thermodynamic computing devices is that their components exist at the same scale as the relevant thermodynamic fluctuations, making them available as a resource, and that these fluctuations are used for inference and learning on-chip. Thermodynamic computing is especially interesting in the current context due to its natural connection to sampling-based methods, and in particular to the Energy-Based Models (EBMs) in machine learning research, which the dynamics of physical systems can naturally express and hence run potentially far more efficiently, on thermodynamic hardware.

In this paper, we demonstrate energy-based thermodynamic computing through a set of $M$ thermodynamic neurons, which are composable subsystems $\mathcal{S}$, each described by Langevin dynamics~\cite{langevin1908sur, crooks1999excursions}, such that the full state space is $\mathcal{X} = \mathcal{S}_1 \times \dots \times \mathcal{S}_M$. We first outline the concepts required for this equilibrium based computing to provide a proof of principle and discuss the computational primitives. We demonstrate one can tune the equilibration times and distributions via control of the potential and temperature. We then show how this approach can be used for the kind of probabilistic inference that is required for modern machine learning. We then introduce a framework based on probabilistic graphical models that couples these subsystems together to form expressive models whose energy function depends on the node and inter-node coupling potentials, both of which can be parametrized and controlled. We discuss how these building elements can be used to realize complex applications, including large scale machine learning models.
Finally, we present our preliminary findings on constructing a basic but essential building block for an analog thermodynamic computer based on superconducting hardware. Superconducting circuits allow for quadratic energy terms from inductance, as well as subsystems with double-well potentials and nonlinear couplings between them using the Josephson junction~\cite{golubov2004current}. Furthermore, on-chip dissipation in superconducting circuits is naturally very low and allows information processing at energy levels many orders of magnitude closer to the Landauer limit than classical computing~\cite{saira2020nonequilibrium}. Additionally, the associated temperature and energy regimes allow us to harness ambient thermal fluctuations, and do not require injecting noise artificially.

\section{Energy-based models from equilibria of physical systems}

\subsection{From the Gibbs distribution to energy-based models}

The core of thermodynamic computing is the Gibbs distribution (also called the Boltzmann distribution), 
\begin{equation}
    \pi_\theta({x}) = \frac{1}{Z_\theta} e^{-\beta E_\theta(x)},
    \label{eq:gibbs-distribution}
\end{equation}
describing the state space distribution of a physical system at thermal equilibrium with an energy function $E_\theta(x)$ parametrized by parameters $\theta$, where ${x} \in \mathcal{X}$ is the state described by a set of system variables, $\beta = (k_BT)^{-1}$ is the inverse temperature, and $Z_\theta = \int_{\mathcal{X}} e^{-\beta E_\theta(x)} d {x}$ is the partition function where the integral is taken over the space $\mathcal{X}$, the set of accessible states. 

The Gibbs distribution [Eq.~\eqref{eq:gibbs-distribution}] has inspired a class of machine learning models called energy-based models (EBMs)~\cite{huembeli2022physics, lecun2006tutorial, du2019implicit}. These physics-inspired models employ the notion of energy to define probability distributions based on Eq.~\eqref{eq:gibbs-distribution} by associating a scalar energy to a model variable configuration. Learning amounts to finding an energy function that associates low energy with probable data configurations and high energy with unfavorable configurations. This is achieved by finding the optimal set of parameters ${\theta}^*$. EBMs are highly flexible machine learning ans\"atze that have potent features like composability and the ability to do conditional inference through clamping. However, EBMs are difficult to scale on traditional digital hardware due to the computational cost of sampling routines dealing with the intractable normalization constant $Z_\theta$~\cite{song2021train}.

Due to their roots in physics, EBMs lend themselves to be implemented in analog hardware. Many ideas have been explored, such as Ising machines~\cite{jelinvcivc2025efficient, moy20221, inagaki2016coherent, mohseni2022Ising, aadit2022massively, singh2024cmos, laydevant2024training, chou2019analog}, probabilistic computers~\cite{camsari2017stochastic, camsari2019p, kaiser2021probabilistic, chowdhury2023full, niazi2024training, Freitas2026NEATRN}, quantum annealers~\cite{johnson2011quantum, king2022coherent, king2023quantum}, neuromorphics~\cite{shainline2017superconducting, kumar2025evaluation, kudithipudi2025neuromorphic, aimone2025neuromorphic}, and linear Ornstein–Uhlenbeck (OU) processors~\cite{coles2023thermodynamic, aifer2024thermodynamic}. However, some early ``thermodynamic" proposals~\cite{coles2023thermodynamic, melanson2023thermodynamic, aifer2024thermodynamic, aifer2024thermodynamic2, donatella2024thermodynamic} fail to fully leverage thermodynamic fluctuations, as they rely on artificially injected Gaussian noise. As a result, they are unlikely to provide meaningful speedups or energy savings.

EBMs have found applications in a variety of areas~\cite{osadchy2004synergistic, ranzato2006efficient, zhai2016deep} as they offer a number of advantages such as flexibility of parametrization, composability, ease of conditional sampling and more~\cite{du2019implicit}. Consider a parametrized energy function $E_\theta (x, y)$ over two states $x \in \mathcal{S}_1$ and $y \in \mathcal{S}_2$, which gives rise to a joint probability density 
\begin{equation}
     \pi_\theta({x, y}) = \frac{1}{Z_\theta} e^{-\beta E_\theta(x, y)},
\end{equation}
with normalization $Z_\theta = \int_{S_1 \times S_2} e^{-\beta E_\theta(x, y)} dx dy$. To sample from the conditional distribution $\pi_\theta(x|y) = \frac{1}{Z_\theta(y)} e^{-\beta E_\theta(x, y)}$, where $Z_\theta(y) = \int_{S_1} e^{-\beta E_\theta(x, y)} dx$, one can simply fix the value of $y$, and sample only the $x$ variables using the same method used to sample the joint distribution. This process, herein referred to as ``clamping", allows for efficient inference in scenarios where partial information is known or when exploring conditional relationships in the data. Conditional sampling also reveals the ease of composing EBMs. Multiple energy functions can be composed to form more expressive distributions directly in energy space. Through simple operations combining elementary energy functions (e.g., addition, subtraction, etc.), we can express conditional relationships between variables, such as logical AND/OR/NOT~\cite{liu2022compositional}. While the energy function in most ML cases is dimensionless, clearly physical energy functions are not. In this work we generally operate in a standard dimensionless setting (in which physical parameters are represented through dimensionless variables or incorporated into the energy function itself).

The standard approach to training EBMs is through maximum likelihood estimation (MLE). In MLE, the goal is to maximize the likelihood of the observed data under our model. Equivalently, we can minimize the negative log-likelihood. When there are no latent states, the objective is given by
\begin{equation}
    \label{eq:mle}
    \mathcal{L}(\theta) = \mathbbm{E}_{x \sim \pi_{\text{data}}} \left [ - \log \pi_\theta(x) \right ] ,
\end{equation}
where $\pi_{\text{data}}$ is the true data distribution, and $\mathbbm{E}$ denotes the expected value.

To optimize this objective for an EBM $\pi_\theta (x)$, a variety of techniques can be employed~\cite{song2021train}. Here, we focus on the contrastive divergence (CD) learning rule~\cite{hinton2002training, carreira2005contrastive, bengio2009justifying, sutskever2010convergence}, which is less commonly used in modern machine learning compared to score-based methods~\cite{song2020score}. This is largely due to the difficulty of sampling from high-dimensional models (a challenge which score matching sidesteps). In this work, we focus on hardware which accelerates the sampling subroutine, and thus do not have to rely on score-based methods and their approximations to the difficult to compute Hessian of energy. In CD-based learning rules, we compute the gradient via

\begin{equation} 
\label{eq:train-ebm-cd}
\begin{aligned}
\nabla_\theta \mathcal{L}(\theta) 
&= - \mathbbm{E}_{x \sim \pi_{\text{data}}} \bigl[ \nabla_\theta \log \pi_\theta (x) \bigr] \\
&= \mathbbm{E}_{x \sim \pi_{\text{data}}} \bigl[ \beta \nabla_\theta E_\theta (x) \bigr]
   - \mathbbm{E}_{x \sim \pi_\theta} \bigl[\beta \nabla_\theta  E_\theta (x) \bigr].
\end{aligned}
\end{equation}

In the equation above, the left term, $\mathbbm{E}_{x \sim \pi_{\text{data}}} \left[ \beta \nabla_\theta E_\theta (x) \right]$, is often called the ``positive phase" which is based on evaluating on the data (and is often much easier to compute) and the right, $\mathbbm{E}_{x \sim \pi_\theta} \left[\beta \nabla_\theta  E_\theta (x) \right]$, is called the ``negative phase" (and is usually the source of the vast difficulty for EBM training). In the case of latent variables, our objective reads
\begin{equation}
\begin{aligned}
\mathcal{L}(\theta) 
&=  \mathbbm{E}_{x\sim \pi_{\text{data}}} 
   \left[- \log \int_{\mathcal{S}_2} \pi_\theta (x, y) d {y} \right]
\end{aligned}
\end{equation}

We then obtain:
\begin{equation}
\label{eq:train-ebm-cd2}
\begin{aligned}
\nabla_\theta \mathcal{L}(\theta)
&= \mathbbm{E}_{x \sim \pi_{\text{data}}} 
   \mathbbm{E}_{y \sim \pi(\cdot|x,\theta)} 
   \bigl[\beta  \nabla_{\theta} E_{{\theta}}(x, y)\bigr] \\
&\quad - \mathbbm{E}_{(x, y) \sim \pi_\theta} 
   \bigl[\beta  \nabla_{\theta} E_{{\theta}}(x, y)\bigr]
\end{aligned}
\end{equation}

Equations~\eqref{eq:train-ebm-cd} and \eqref{eq:train-ebm-cd2} allow us to use standard stochastic gradient-based optimization methods to optimize the parameters $\theta$ of the EBM if we can approximate the expected values through sample averages.

\subsection{Sampling based on digital Langevin dynamics}

A common approach to sampling from EBMs is to simulate Langevin dynamics, which can be programmed to converge to a target distribution based on the gradient of the log probability~\cite{cheng2018convergence, cheng2018underdamped, zhang2022langevin, sun2023discrete}.

The celebrated Langevin Monte Carlo method~\cite{roberts1996exponential} represents such an approach to draw samples ($x \in \mathbb{R}^D$) from an EBM. The method is based on the concept of a gradient flow that brings the distribution $P(t)$ starting from an initial distribution $P_0 = P(t=0)$ progressively closer to the distribution $\pi_\theta$ with increasing time. Indeed, samples can be obtained from the path of the Langevin diffusion process, whose states $x(t)$ have distribution $P(t)$ and evolve according to the stochastic ordinary differential equation
\begin{align}
\label{eq:overdamped_langevin}
    \gamma_i d x_i = -\frac{\partial U_\theta(x)}{\partial x_i} dt + \sqrt{\frac{2 \gamma_i}{\beta}} dW_t^{(i)},
\end{align}
where $dW_t^{(i)}$ are the increments of independent Wiener processes, $i$ indexes the $D$ degrees of freedom, and we typically set $\gamma_i = 1$. To draw samples numerically, we discretize this diffusion process, e.g., via the Euler-Maruyama scheme or custom higher-order schemes. This leads to a biased numerical sampling scheme (that for low-order methods can be fixed via a Hastings correction).

Despite many ongoing efforts and the strong potential benefits, sampling from EBMs [and therefore also training EBMs using the gradient estimators in Eqs.~\eqref{eq:train-ebm-cd} and \eqref{eq:train-ebm-cd2}] is extraordinarily difficult for high-dimensional models. Fundamentally, this difficulty comes from sampling, as in general, sampling in high-dimensional models is done via iterative locally informed proposals (such as digital Langevin dynamics), which means that the probability of overcoming barriers and valleys in this landscape can be very low, resulting in inefficient iterative sampling. Thus, it is natural to ask, rather than crudely and inefficiently simulating these dynamics numerically, can we build an analog physical device which exhibits controllable Langevin dynamics to efficiently and speedily sample? This idea represents the core backbone of our energy-based thermodynamic paradigm.

\subsection{Sampling based on physical Langevin dynamics}

Consider the dynamics of a physical system with $D$ degrees of freedom (again, $x \in \mathbb{R}^D$) where inertia plays a significant role, and the motion occurs in the presence of damping and random fluctuations. Such stochastic processes commonly occur in statistical mechanics when the time evolution of degrees of freedom in an energy landscape influenced by thermal noise is of interest, and are characterized by the underdamped Langevin equation, which is a stochastic ordinary differential equation of the following form 
\begin{align}
    \begin{split}
        dx_i &= \frac{p_i}{m_i} dt \\
        dp_i &= -\left(\frac{\partial U_\theta(x)}{\partial x_i} + \frac{\gamma_i}{m_i} p_i\right)dt + \sqrt{\frac{2 \gamma_i }{\beta}} dW_t^{(i)},
    \end{split}
    \label{eq:underdamped_langevin}
\end{align}
where $x_i(t)$ and $p_i(t)$ are the position and conjugate momentum of the $i$-th degree of freedom at time $t$, respectively, $U_\theta(x)$ is the potential energy function describing the energy landscape, $\gamma_i$ is the damping coefficient that quantifies resistance to motion, and $m_i$ is the mass. The term $\sqrt{\frac{2 \gamma_i }{\beta}}$ ensures that the system satisfies the fluctuation-dissipation theorem, linking damping and noise in thermal equilibrium~\cite{balakrishnan1979fluctuation}. We will later present our superconducting hardware realization, which is a tunable double-well system engineered to be operated in the thermodynamic domain where its time evolution can be modeled by underdamped Langevin equations.

The Langevin equation~\eqref{eq:underdamped_langevin} corresponds to a Fokker-Planck equation which is a partial differential equation describing the normalized probability density of $x$ and $p$ at time $t$ by $P(x, p, t)$ that can generally be written in operator form as $\frac{\partial}{\partial t} P(x, p, t) = \mathcal{L}^* P(x, p, t)$, where $\mathcal{L}^*$ is the adjoint operator~\cite{sarkka2019applied}.
In the case of underdamped Langevin dynamics, this Fokker-Planck equation is also called the Klein-Kramers equation, and the adjoint operator acts as
\begin{multline}
    \label{eq:FokkerPlanck}
    \mathcal{L}^* (\cdot) = \\
    \left[\sum_{i=1}^D \left( -\frac{p_i}{m_i} \frac{\partial}{\partial x_i} + \frac{\partial}{\partial p_i} \Bigl(
      \frac{\partial U}{\partial x_i} 
      + \frac{\gamma_i}{m_i} p_i\Bigr) 
       + 
      \frac{\partial^2}{\partial p_i^2} 
        \frac{\gamma_i}{\beta} 
       \right) \right](\cdot)
\end{multline}
The Fokker-Planck equation provides a deterministic way to describe the time evolution of the probability distribution associated with the stochastic process. This is useful for understanding ensemble behavior rather than individual sample paths of the Langevin equations~\eqref{eq:underdamped_langevin}. For instance, we can use Eq.~\eqref{eq:FokkerPlanck} to identify steady-state distributions, mean first-passage, and escape times.

In particular, it is well known that the Fokker-Planck equation associated with underdamped Langevin dynamics has the equilibrium solution~\cite{risken1989fokker}
\begin{align}
    \pi_\theta(x,p) = \frac{e^{-\beta E_\theta(x, p)}}{Z_\theta},
\end{align}
where $\log Z_\theta$ can be interpreted as the free energy (when multiplied by $-\beta^{-1}$) and $E_\theta(x, p) =  \frac{p^2}{2m} + U_\theta(x)$. Marginalization over $p$ leads to
\begin{align}
    \pi_\theta(x) = \frac{e^{-\beta E_\theta(x)}}{Z_\theta},
    \label{eq:steady_state_marginalized}
\end{align}
where $E_\theta(x) = U_\theta(x)$, and which is also the equilibrium solution of the overdamped Langevin equation~\eqref{eq:overdamped_langevin} corresponding to the limit of Eq.~\eqref{eq:underdamped_langevin} with small mass and large damping. 

Equation~\eqref{eq:steady_state_marginalized} represents the backbone of an energy-based thermodynamic computer. Namely, the steady-state distribution of physical Langevin dynamics is a Gibbs distribution with energy potential $U_\theta(x)$. If we can design a physical system with a sufficiently controllable energy potential landscape, so that the time and energy consumption for physical equilibration and readout is smaller than for digital MCMC sampling, we expect computational benefits from such a thermodynamic computing device. The key attributes of such a system would be thermalization time (the amount of time it takes to go from the initial condition to the steady state) and thermalization energy (the amount of energy required to go from the initial condition to the steady state). This idea harkens back to early ideas of thermodynamic computation as computation driven by analog noise, such as ``Brownian computers"~\cite{bennett1982thermodynamics}.

\subsection{Computing expected values}

In addition to (conditional) sampling from energy functions, we consider averaging as another important building block to further increase the number of possible operations that we can perform with our hardware. This not only helps provide intuition into the energy functions we are working with, but since we expect thermodynamic computers to operate in extremely low energy regimes, we might also expect to gain energy advantages from approximate averaging of deterministic operations. Expected values of EBMs representing a conditional distribution give rise to parametrized deterministic functions of interest
\begin{align}
    \label{eq:expected_value}
    f(z) = \mathbbm{E}_{y \sim \pi_\theta(\cdot|z)} \left [y \right ],
\end{align}
where we have clamped $z$. Such an expectation can be achieved by measuring the state $y$ over an extended time or averaging the samples from repeated runs. This can be achieved through analog to digital converters, but this often comes with engineering downsides. We discuss later how such an operation could be achieved without digital conversions. For now, consider an estimator based on repeated runs.  

For a set of \(N\) samples \(\{y_i\}_{i=1}^N\), where each \(y_i \in \mathbb{R}^d\), the mean estimator is computed as 
\begin{equation}
    \bar{y}_N = \frac{1}{N} \sum_{i=1}^N y_i,
\end{equation}
resulting in a vector \(\bar{y}_N \in \mathbb{R}^d\) that approximates the expected value \(\mathbbm{E}[y]\). The sample covariance matrix is
\begin{equation}
    \hat{\Sigma} = \frac{1}{N-1} \sum_{i=1}^N (y_i - \bar{y}_N)(y_i - \bar{y}_N)^\top,
    \label{eq:covariance_matrix}
\end{equation}
where \(\hat{\Sigma} \in \mathbb{R}^{d \times d}\). Its diagonal elements are the sample variances of the components of \(y\), while its off-diagonal elements are cross-covariances. For independent samples with population covariance \(\Sigma\), 
\begin{equation}\label{eq:cov_est}
    \text{Cov}(\bar{y}_N) = \Sigma/N,
\end{equation}
which can be estimated by $\hat{\Sigma}/N$.

\subsection{Energy-time-precision trade-offs}

The Monte Carlo standard error scales as $O(N^{-1/2})$. Unlike traditional floating-point computations where precision is fixed by the number format (e.g., Float32 or Float16), our approach provides (almost) continuously tunable precision (assuming we are operating with an averaging component, there are many cases as we will see later in which we do not need averaging). For computational operations whose output covariance does not depend on the encoded mean, the absolute precision is independent of the magnitude of the computed value: changing $\mathbbm{E}[y]$ does not change the standard error. Relative precision is naturally not always magnitude-independent. The number of samples can be adjusted at runtime to meet the required absolute or relative error tolerance. This enables adaptive precision, with more samples allocated to critical computations and fewer samples used where lower precision is acceptable. Halving the error requires four times as many independent samples, obtained either sequentially at increased latency or through four times as many parallel samplers increased area cost. Practical implementations must carefully balance these factors based on application requirements.

Landauer's principle states that erasing a bit of information must dissipate at least $k_BT \ln(2)$ energy, where $k_B$ is Boltzmann's constant and $T$ is the temperature. This limit need not be in terms of energy alone but can be expressed in terms of any conserved quantity~\cite{vaccaro2011information}. It represents the absolute minimum energy required for non-reversible computation (reversible computation can be performed in finite time with zero error and zero energy dissipation~\cite{fredkin1982conservative}). Landauer's limit is the lower limit of the energy consumed by an irreversible computation, and at room temperature is approximately $3\times10^{-21}$ joules. This limit is many orders of magnitude lower than what is currently used in deterministic digital computing systems~\cite{shankar2023energy}. In digital systems, every operation is precisely defined and executed, leading to highly accurate and reproducible results. This precision, however, comes at a significant energy cost. Each operation, whether it is a simple bit flip or a complex arithmetic calculation, requires a substantial amount of energy compared to this theoretical minimum. Such energy expenditure is necessary to maintain the system's state against thermal fluctuations and to ensure the reliability of each operation. In contrast, computation based on thermodynamic principles could potentially operate with energy consumption much closer to Landauer's limit.

In practice, computation cannot simultaneously achieve high precision, high speed, and low power consumption. We can immediately identify some of the key tradeoffs of thermodynamic computation: increasing precision comes at the cost of increased power dissipation or reduced speed; faster computation requires either reduced precision or increased power dissipation; and operating with less energy necessitates either slower or less precise computation. It is also worth pointing out that the system cannot be changed arbitrarily fast because the thermalization time, $\tau_{\text{therm}}$, will put a limit on the maximum speed. Therefore, $\tau_{\text{therm}}$ represents the fastest timescale at which our system can reach equilibrium, setting a natural speed limit for our computations. Of course, in theory we can tune the thermalization time by carefully selecting parameters in the Langevin equations [Eq.~\eqref{eq:underdamped_langevin}]. However, any physical implementation will bound the parameter space and result in some platform-dependent thermalization time.

This discussion can be formalized using the concepts of thermodynamic~\cite{nakazato2021geometrical, reilly2025physical} and information-geometric speed limits~\cite{lahiri2016universal, ito2018stochastic, ito2020stochastic, ito2024geometric}. The dissipation and work tradeoffs for training EBMs have also been discussed \cite{hnybida2025minimal}. These limits characterize the trade-offs between entropy production and the transition time required for a system to evolve from an initial probability distribution to a final one, where the distance between the two distributions is measured in terms of the Wasserstein distance. \citet{ito2024geometric} provides a unified theory, referred to as geometric thermodynamics for the Fokker-Planck equation, further connecting these works to optimal transport~\cite{villani2009optimal, klinger2025minimally}.

After completion of this work, we became aware of Ref.~\cite{rolandi2026energy}, which studies energy-time-accuracy trade-offs and optimal driving protocols rather than the equilibrium model constructions considered here.

Next, we discuss how one can (approximately) compute $\tau_{\text{therm}}$ for a given physical realization described by the Langevin equations~\eqref{eq:underdamped_langevin}.

\subsection{Estimating thermalization time}
\label{sec:thermalization_time}

For simple energy functions, we can analytically compute the characteristic timescales of equilibration via the smallest nonzero eigenvalue (the spectral gap) of the Fokker-Planck operator [Eq.~\eqref{eq:FokkerPlanck}]. In the case of (overdamped) Gaussian potentials, the spectral gap is known to be proportional to $\frac{1}{\gamma}$. If the potential $U_\theta(x)$ has multiple wells, equilibration involves rare transitions over barriers. The equilibration time will be dominated by the Kramers escape time, which exhibits an exponential dependence on well heights~\cite{kramers1940brownian, buttiker1983thermal}.

For more generic potentials, estimating $\tau_{\text{therm}}$ becomes more difficult~\cite{risken1985eigenvalues}. We can still numerically approximate $\tau_{\text{therm}}$, by discretizing over the states and computing the eigenvalues of the Fokker-Planck operator matrix. Another approach is to numerically integrate the Langevin equations~\eqref{eq:underdamped_langevin} for many different noise realizations. We then use these trajectories to track how the probability distribution $P(x, p, t)$ converges to the equilibrium Boltzmann distribution $\pi_\theta$ using metrics such as the total variation distance or the KL divergence between $P(x, p, t)$ and $\pi_\theta$.

\subsection{Estimating thermodynamic work}
\label{sec:work}

To characterize the energy dynamics of a thermodynamic computer, we consider time-dependent potentials $U_\theta(x(t), \lambda(t))$, where $\lambda(t)$ is a control parameter. Such time dependence arises, for example, when adjusting the bias of a Gaussian potential toward a target value, or when coupling modes. Changing $\lambda(t)$ performs stochastic control work on the system. The work and heat may have either sign on individual trajectories, as may the system entropy change; the second law constrains the average total entropy production of the system and environment. With work defined as being done on the system, the trajectory-level protocol work is~\cite{sekimoto2010stochastic, ray2023gigahertz, wimsatt2021harnessing, saira2020nonequilibrium}
\begin{align}
\label{eq: work integral}
     W(\tau) = \int_0^{\tau} \frac{\partial U_\theta(x(t), \lambda(t))}{\partial \lambda} \frac{d \lambda}{dt} dt,
\end{align}
which we approximate along trajectories obtained by numerically solving the Langevin equations~\eqref{eq:underdamped_langevin}. We evaluate the distribution $P(W(\tau))$ and its mean over many trajectories in our numerical studies. Evidently, $W(\tau) = 0$ when the potential is time-independent after initialization.

For work defined as being done on the system, $\Delta F = F(\tau) - F(0)$ is the minimum average work required in a reversible isothermal transformation between the equilibrium states associated with the endpoint control values. Equivalently, the maximum reversible work extractable from the system is $-\Delta F$. For an equilibrium-initialized ensemble, Jarzynski's non-equilibrium work relation~\cite{jarzynski2012equalities} connects the equilibrium free energy difference to the protocol work $W(\tau)$ [Eq.~\eqref{eq: work integral}]
\begin{align}
\label{eq: free energy change}
    \Delta F = - \frac{\log \mathbbm{E} \left[ \exp(-\beta W(\tau)) \right ]}{\beta} , 
\end{align}
where the expectation is over complete stochastic trajectories initialized from equilibrium~\cite{shiraishi2023introduction}. Practically, this expectation is approximated through sampling. When the dissipated work is large or broadly distributed, a significant number of samples may be required~\cite{vaikuntanathan2011escorted}, because the exponential average in Jarzynski's relation gives substantial weight to rare trajectories with low work~\cite{park2003free}.

The trajectory-level dissipated work is $W_{\mathrm{dis}} = W - \Delta F$, which can be negative for individual trajectories. Jarzynski's equality and Jensen's inequality imply the ensemble inequality
\begin{align}
    \langle W_{\mathrm{dis}} \rangle = \langle W \rangle - \Delta F \geq 0.
\end{align}
In the following, we use $P(W)$ and $\langle W \rangle$ as idealized work metrics. They are meant to be bounds on the end-to-end energy consumed by the physical system (but omit readout, calibration, refrigeration, etc.).

\section{Elemental potentials}
\label{sec:potentials}

This section examines the elemental potentials that we use to generate probabilistic graphical models. These potentials are ``elemental" because they form the building blocks for many more complicated functions and are representative of simple implementations of quadratic and quartic potentials of oscillators. We consider basic single-particle, coupling, and many-particle potentials $U_\theta$. We conduct analysis of the properties of these potentials based on the approaches outlined in the previous section. Some of these building blocks resemble those presented in Ref.~\cite{whitelam2024thermodynamic}; however, our focus is exclusively on equilibrium regimes. 

\subsection{Single-particle potentials}

The two main classes of single-particle potentials we focus on are Gaussian (also called single-well or quadratic), where the force is affine in $x$ (here, $x \in \mathbb{R}$) , and nonlinear (also called quartic or double-well) potentials. 

The energy function of a Gaussian oscillator takes the form: 
\begin{equation}
    \label{eq:gauss_general}
    U_\theta^{\rm{sw}}(x) = \frac{1}{2 \sigma} (x - \mu)^2,
\end{equation}
where the parameters $\theta = (\mu, \sigma)$ allow us to tune the mean and variance of the potential. In particular, clamping can be achieved using Eq.~\eqref{eq:gauss_general}, where the mean corresponds to the clamped value and the variance is minimized as much as possible. In mechanical oscillators, $\frac{1}{\sigma} = m\omega^2$ is a product of mass $m$ and frequency squared $\omega^2$.

\begin{figure}[!htbp]
    \centering
    \includegraphics[width=0.95\linewidth]{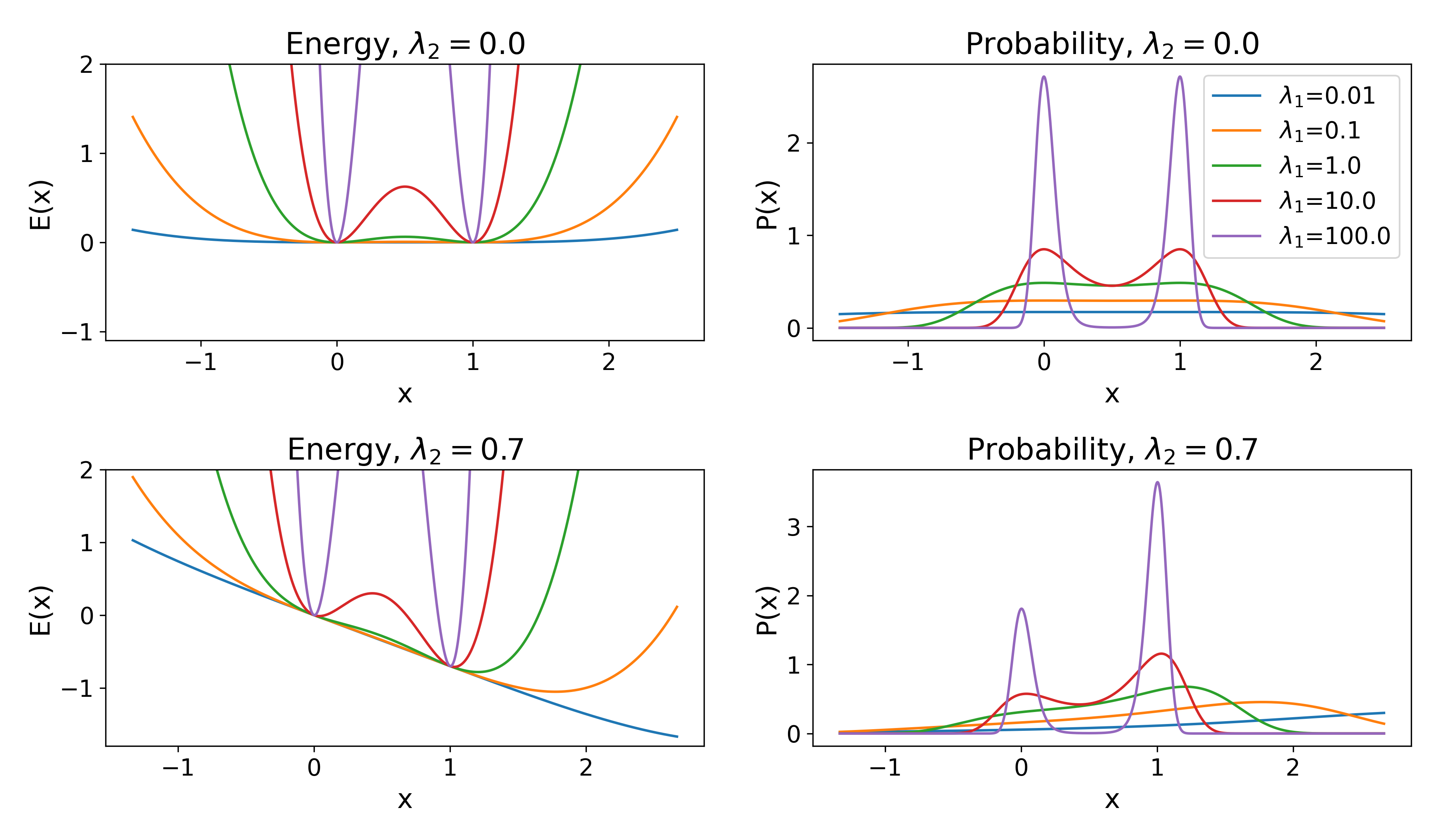}
    \caption{Comparison of energy curves and probability distributions for the double-well potential [Eq.~\eqref{eq:dw}] for varying choices of $\lambda_1$ and a bias value of $\lambda_2=0$ and $\lambda_2=0.7$. The bias tilts the double-well potential towards one side and increases the probability to be in one of the two wells.}
    \label{fig:double-well_energy_and_prob}
\end{figure}

The double-well potential is defined as
\begin{align}
    U_\theta^{\rm{dw}}(x) =  \lambda_1 x^2 (x - 1)^2 - \lambda_2 x,
    \label{eq:dw}
\end{align}
where the parameters $\theta = ( \lambda_1, \lambda_2)$ determine the height of the energy barrier separating the two wells and the tilt of the potential, which skews the relative depths of the wells, respectively. Double-well potentials are ubiquitous in physics, especially quantum physics~\cite{schafer2020spectral, borah2021measurement, wu2022nonequilibrium}. Figure~\ref{fig:double-well_energy_and_prob} illustrates these operations. Note the increased probability of measuring the position $x$ of the oscillator in one of the two wells depending on the tilt. Unlike Gaussian oscillators, thermal fluctuations play a crucial role in the double-well system, enabling transitions between the two wells. At zero temperature, such transitions occur only if the oscillator starts with sufficiently high momentum. 

If several independent single-particle potentials are considered (i.e., the energy function is a sum of single-particle potentials), the spectral-gap relaxation timescale is set by the slowest independent component. The thermalization time, $\tau_{\rm{therm}}$, of Gaussian potentials is on the order of nanoseconds under reasonable superconducting hardware assumptions (100~fF, 100~pH, 20~k$\Omega$, operating at 150~mK, see Sec.~\ref{sec:superconducting}). Changes in operating parameters significantly impact this time, making it highly dependent on the hardware implementation. In particular, for the double-well potential, $\tau_{\rm{therm}}$ depends exponentially on $\lambda_1$ (as discussed previously in Sec.~\ref{sec:thermalization_time}). 

\subsection{Coupling potentials}

Naturally, non-interacting potentials can only get so far, and coupling potentials together is a next step. A simple version of a coupled potential is a Gaussian coupled to a double well,
\begin{align}
    U_\theta^{\rm{sig}}(x, z) =  \lambda_1 x^2 (x - 1)^2 - z x + z^2,
    \label{eq:dw2}
\end{align}
which closely resembles the single-particle double-well potential [Eq.~\eqref{eq:dw}]. Here, $z$ represents another oscillator. If $z$ is sufficiently heavy (or if $z$ is a very low variance Gaussian), we can treat it as effectively static and use the same approximations as we did for single-particle potentials where $z$ is just treated as a constant (this will be important, as discussed in Sec.~\ref{sec:mpp}). As a pedagogical (and later relevant) example, let us consider computing the expected value for an EBM based on Eq.~\eqref{eq:dw2} as before and assume $z$ is effectively static. This potential allows us to program the commonly used sigmoid activation $\sigma_{\rm{ML}}$:
\begin{align}
    \label{eq:sigmoid}
    f(z) &= \lim_{\lambda_1 \to \infty} \mathbbm{E}_{x \sim \pi_\theta(\cdot|z)} \left [x \right ]
     = \frac{1}{1 + \exp(-z)}
     = \sigma_{\rm{ML}}(z).
\end{align}
It is important to note that the well height, $\lambda_1$, in double-well potentials represents a tradeoff. To recover the deterministic sigmoid function, one may want to maximize the well height. However, recall that this comes at the cost of reduced computation speed, as the thermalization time increases exponentially with well height. As expected, a smaller $\lambda_1$  results in a shorter  $\tau_{\rm{therm}}$ (allowing us to draw the first and subsequent samples with less waiting time, then converging to the sampled mean via $O \left (\frac{1}{\sqrt{N}} \right )$ for $N$ samples), but at the cost of higher error (since it is a worse approximation). These values are based on the same hardware assumptions as before (and computed with the eigenvalues of the overdamped Fokker-Planck operator). The eigenvalues were computed numerically, using SciPy~\cite{virtanen2020scipy}. 


We can also engineer time-dependent coupling potentials between two oscillators
\begin{align}
    \label{eq:coupling_potential}
    U_{\theta}^{\rm{c}}(x, y)= \lambda(t) (x - y)^2,
\end{align}
where the coupling is given by 
\begin{align}
    \lambda(t) = \lambda_c\left[\sigma_{\rm{ML}}(k(t-t_{\rm on}))-\sigma_{\rm{ML}}(k(t-t_{\rm off}))\right],
\end{align}
where $\theta = (\lambda_c, k, t_{\rm on}, t_{\rm off})$, $\lambda_c$ is the coupling strength, and $k$ determines the timescale for the coupling. This protocol switches the coupling on and (optionally) subsequently off. Suppose we have two single-particle potentials,  $U_\theta(x)$  and $U_\theta(y)$, along with a coupling potential $U_{\theta}^{\rm{c}}(x, y)$. To achieve an optimal energy-time trade-off, we typically aim to vary the coupling on the same timescale as the equilibration times dictated by the single-particle potentials.

\begin{figure}[htbp]
    \centering
    \subfigure[Work distribution of two coupled Gaussians.]{
        \includegraphics[width=0.45\linewidth]{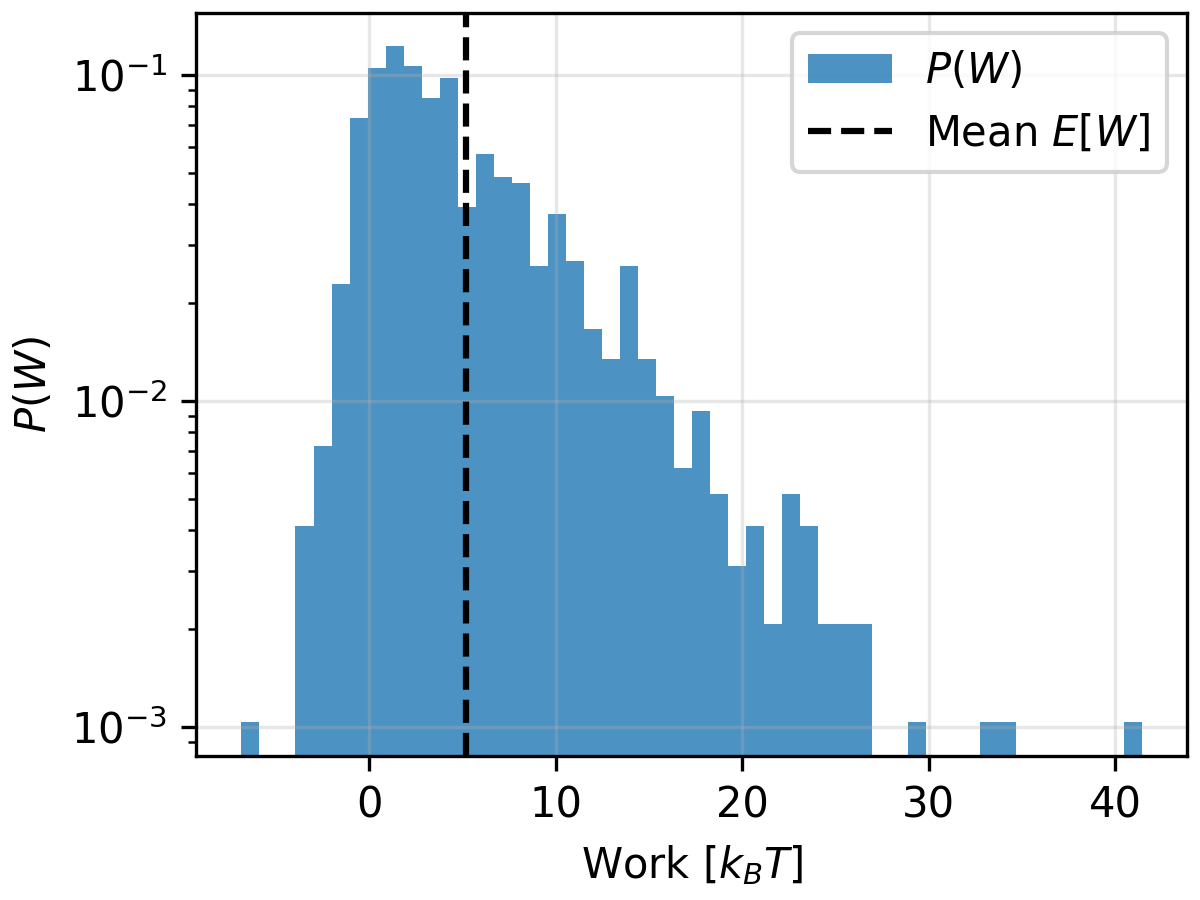}
        \label{fig:work_coupled_gaussians}
    }
    \hfill
    \subfigure[Work distribution of a Gaussian coupled to a double-well.]{
        \includegraphics[width=0.45\linewidth]{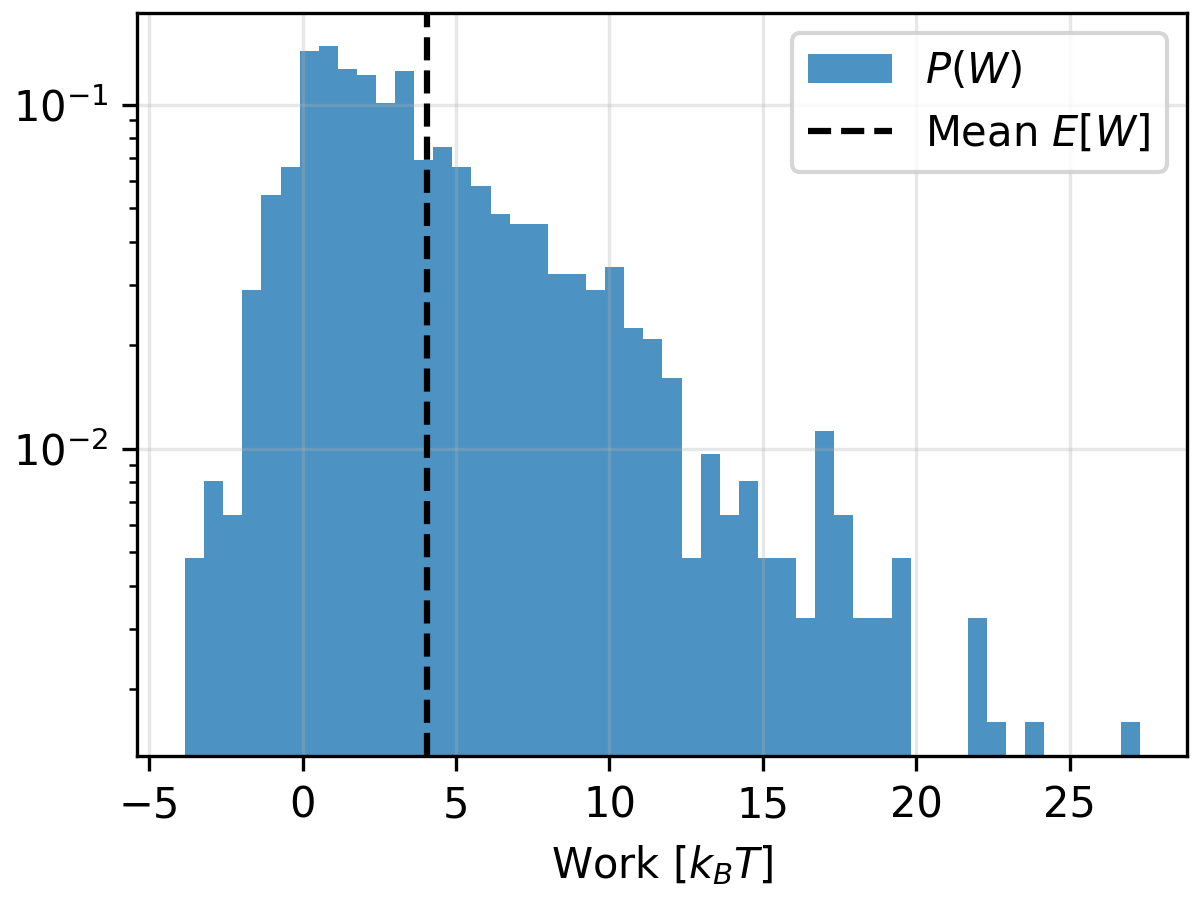}
        \label{fig:work_coupled_gaussian-dw}
    }
    \caption{Distributions of the work for the coupling in Eq.~\eqref{eq:coupling_potential}. Dashed lines indicate sample means. We use $N=1000$ trajectories, $t\in[0,100]$, $\lambda_c=100$, $t_{\rm on}=100/3$, and $t_{\rm off}=200/3$. Initial positions are sampled uniformly from $[-1,1]$ and initial momenta from $[-0.3,0.3]$. The two distributions are similar for this parameter choice.}
    \label{fig:work_coupled}
\end{figure}

Figure~\ref{fig:work_coupled} compares the work distribution in two different cases of coupled oscillators. These figures were created via numerical integration using diffrax~\cite{kidger2021on} with higher order SDE solvers~\cite{foster2023high, foster2021shifted}. Notably, under the idealized model, the mean work is of order $k_BT$ (this excludes the end-to-end energy consumed by the complete physical system). For scale, in units of $k_BT$, the Landauer limit is $\ln 2\approx0.7$. These work values are not directly comparable with reported end-to-end energies for digital computers~\cite{reuther2022ai, choi2014algorithmic, garcia2019estimation}. Chaining low-work potentials may reduce one contribution to computation energy~\cite{whitelam2025training}.

We can also consider other simple coupling potentials that independently couple independent Gaussian oscillators $x \in \mathbb{R}^D$ to other oscillators 
\begin{align}
\label{eq:vector-add}
     \begin{split}
         U_\theta^{\rm{add}}(x, y, z) =  \nu_1 \sum_{i=1}^D \frac{1}{2} x_i^2 -  \nu_2 \sum_{i=1}^D x_i (y_i + z_i)
     \end{split},
\end{align}
where $y,z \in \mathbb{R}^D, \theta = (\nu_1, \nu_2)$ which results in the expected value of $x$ being $ y + z $ (for equal $\nu$), 
\begin{align}
\label{eq:mvp}
     \begin{split}
         U_\theta^{\rm{MVP}}(x, z) =  \frac{\nu}{2} \sum_{i=1}^D  x_i^2 -  x^\top W z  
     \end{split},
\end{align}
where $x \in \mathbb{R}^D$, $z \in \mathbb{R}^{M}$, $W \in \mathbb{R}^{D\times M}$, $\theta=(\nu, W)$, which results in $\mathbbm{E}[x \mid z]=Wz/ \nu$, and hence gives the matrix-vector product $Wz$ when $\nu=1$ (in order to practically apply this operation, we must avoid precomputing the product and just programming it as a Gaussian tilt; and still compute the specific couplings digitally, which asymptotically scales the same as matrix classical operations \cite{borle2018analyzing}).

These coupling potentials can enable an analog form of averaged measurement of an oscillator $x$, an important building block previously referenced in the discussion of methods for computing expected values [cf.~Eq.~\eqref{eq:expected_value}]. For example, an oscillator $y$ can be coupled to a Gaussian oscillator $x$ via Eq.~\eqref{eq:coupling_potential} and made nearly static at the value of $x$ at a time $t$ when $x$ and $y$ are again decoupled. This operation allows us to read out and store the value of an oscillator $x$ into $y$, which conceptually resembles digital ``sample-and-hold'' devices. In detail, this involves two steps. First, perform $N$ coupling operations (with $N$ independent oscillators $\{y^{(i)}\}_{i=1}^N$  waiting $N\tau_{\text{therm}}$ for $N$ samples, note that we have to wait that long because the $N$ oscillators are coupled to a single input, and thus even in parallel they will experience the same position of the input) based on the coupling potential $U_\theta^c$ [Eq.~\eqref{eq:coupling_potential}]. Specifically, we can allow $y_i$ to equilibrate to the value of $x$ without perturbing it, then to rapidly increase the effective mass $y_i$ once it is coupled to the oscillator being ``measured". This can be achieved by carefully selecting the parameters in the Langevin Equations~\eqref{eq:underdamped_langevin}. In theory, the coupling can be done very rapidly. Second, couple these oscillators to another oscillator $y^{(N+1)}$ using the potential in Eq.~\eqref{eq:vector-add} with $\frac{\nu_2}{\nu_1} = \frac{1}{N}$ and large $\nu_1$. We refer to the oscillators that implement this averaged measurement operation as estimation oscillators (or relay oscillators~\cite{relay_gadget_analo_USPTO, relay_gadget_multi_well_USPTO}).

\subsubsection{Many-particle potentials}\label{sec:mpp}

Hardware implementations may also enable the coupling of multiple oscillators. These many-particle potentials are a specific kind of coupled potential that couple many particles together. An intuitive example is
\begin{multline}
     U_\theta^{\rm{soft}}(x,z) = \\ 
     \lambda_1 \sum_{i=1}^D x_i^2 (x_i -1)^2 - \sum_{i=1}^D x_i z_i + \lambda_2 \left(\sum_{i=1}^D x_i -1 \right)^2,
\label{eq:softmax}
\end{multline}
where $\theta = (\lambda_1, \lambda_2)$, $x \in \mathbb{R}^D$, and $z \in \mathbb{R}^{D}$. This potential is built on top of the sigmoid potential given by Eq.~\eqref{eq:dw2}. While this potential may look somewhat arbitrary, its expectation yields the softmax function, and we present it as an example of many-particle potentials (there are certainly many others one could consider) and its importance will be highlighted in Sec.~\ref{sec:diff_exp_value}. Although these potentials imply all-to-all connectivity, physical implementations that do not support this can reduce the degree of connectivity by introducing additional degrees of freedom as has been done in other regimes~\cite{thermo_transformer_USPTO,sajeeb2025scalable}.

Working with the potentials introduced in this section offers a set of building blocks that are potentially extremely fast, operating at or below digital clock cycles for entire thermalization processes, and highly energy-efficient, functioning at many orders of magnitude lower energy than digital computers (subject to the hardware implementation of the thermodynamic computer). However, implementing these building blocks as described in this section is not necessarily trivial.

\section{Thermodynamic deep learning and differentiable programs via expected values}\label{sec:diff_exp_value}


Let us now demonstrate how carefully designed energy landscapes can implement basic computational operations through statistical averaging of their thermal equilibrium states. We have already seen an important example in this context: using Eq.~\eqref{eq:sigmoid}, we demonstrated how to implement a sigmoid activation function. More generally, by working with expectations rather than individual samples, we can recover (almost) deterministic operations similar to common digital subroutines while potentially consuming less energy (recall that we can reduce the sample error $O(\frac{1}{\sqrt{N}})$ by drawing $N$ samples which takes time $N\tau_{\text{therm}}$). 

Consider a traditional multilayer perceptron (MLP) composed of stacked layers of functions given by $f(z) = \sigma_{\rm{ML}}(Wz + b)$. These functions can be further decomposed into matrix-vector multiplication, vector addition, and the application of an activation function. The final layer is often a softmax operation that converts output values into probabilities.

Let us first consider the matrix-vector multiplication $x = f(z) = W z$ where $W \in \mathbb{R}^{m \times n}$ and $z \in \mathbb{R}^{n}$. The energy function presented in Eq.~\eqref{eq:mvp} has an equilibrium distribution given by
\begin{align}
     \pi(x|z) = \mathcal{N}\left (Wz, \mathbb{I} \right),
\end{align}
for $z$ clamped and $\nu = 1$. Thus, measuring the expected value of $x$ through estimation oscillators allows us to approximately compute the matrix-vector product $Wz$. 

Similarly, the energy function presented in Eq.~\eqref{eq:vector-add} has the equilibrium distribution
\begin{align}
     \pi(x|y, z) = \mathcal{N}\left (\frac{\nu_2}{\nu_1}(y+ z), \frac{1}{\nu_1} \mathbb{I} \right),
\end{align}
for $y$ and $z$ clamped. Measuring the expected value of $x$ using estimation oscillators allows us to add the values of $y$ and $z$ for $\nu_1 = \nu_2$. It might be tempting to program a linearly tilted Gaussian and simply supply $Wz$ as the tilt instead. However, computing  $Wz$ purely digitally eliminates any advantage of thermodynamic computation. These tilts and interactions must occur within the thermodynamic computer to preserve its benefits.

\begin{figure}[htbp]
    \centering
    \includegraphics[width=0.95\linewidth]{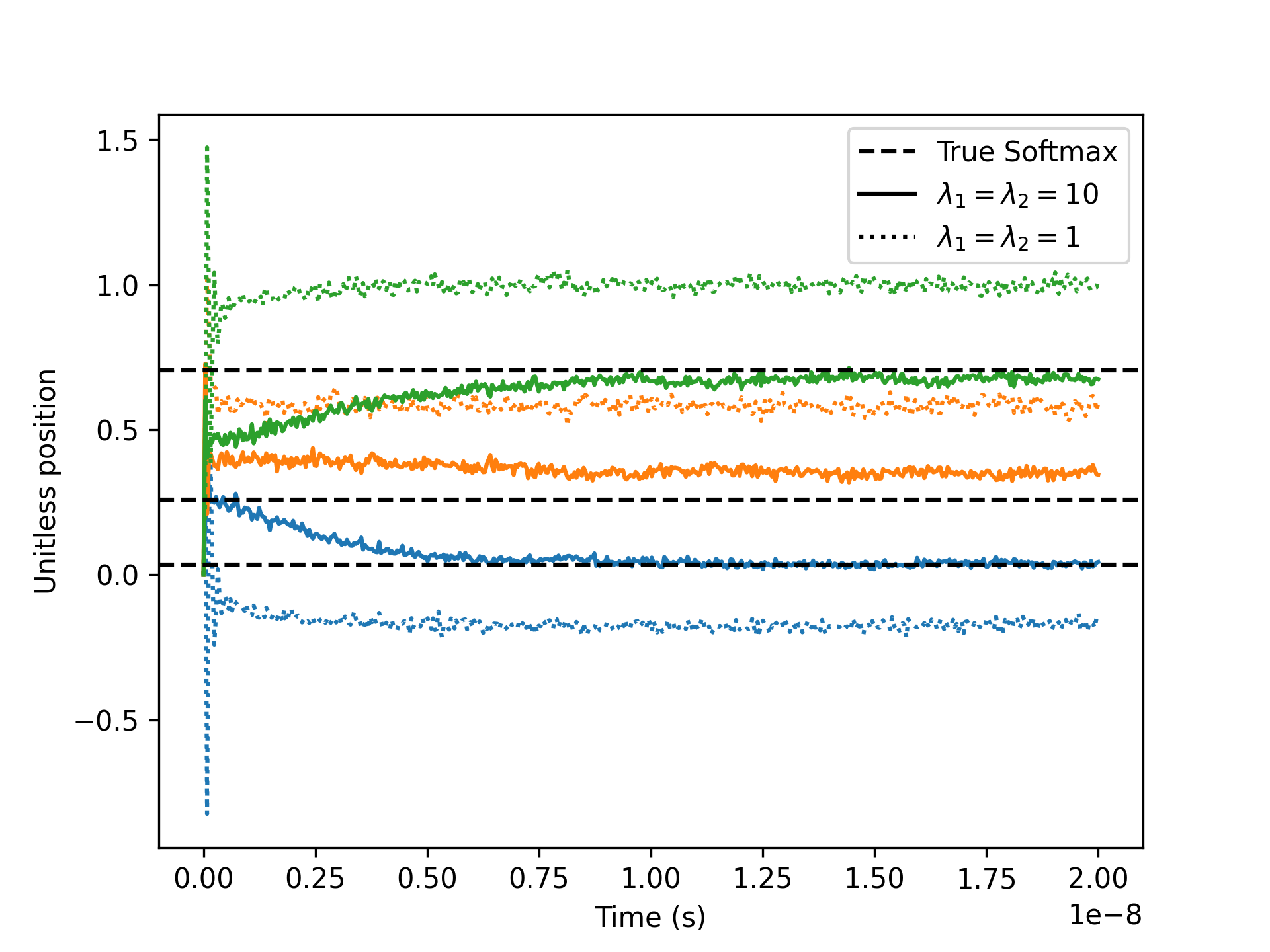}
    \caption{Convergence of the expectation of many trajectories from numerical integration of the softmax potential at different $\lambda$ values.}
    \label{fig:softmax-traj}
\end{figure}

Finally, the energy function  in Eq.~\eqref{eq:softmax}  allows us to approximate a softmax operation. In particular, the parameter $\lambda_2 \to \infty$ enforces the simplex constraint $\sum_i x_i=1$. Then, the parameter $\lambda_1$ lets us turn the continuously valued oscillator values $x$ into one-hot vectors with $\lambda_1 \rightarrow \infty$. In this hard-constraint limit, the one-hot state $e_i$ has energy $-z_i$, so the equilibrium probability over one-hot states is proportional to $e^{z_i}$. In total, we have
\begin{align}
    f_i(z) &= \lim_{\lambda_1, \lambda_2 \to \infty} \mathbbm{E}_{x \sim \pi_\theta(\cdot|z)} \left [x_i \right ] = \frac{\exp(z_i)}{\sum_{j=1}^D \exp(z_j)}.
    \label{eq:softmax2}
\end{align}
Even with $\lambda_1$ and $\lambda_2$ as small as 10, we observe relatively good convergence to the true deterministic softmax. As we will see, low approximation quality is sufficient for training machine learning models. The approximation quality is illustrated in Figure~\ref{fig:softmax-traj}. As expected, larger values of $\lambda_1$ and $\lambda_2$ lead to longer equilibration times, while smaller values of $\lambda_1$ and $\lambda_2$ result in poorer approximations of the softmax. In particular, the outputs may not be perfectly normalized but still maintain a similar structure - i.e., the relative ordering of values remains the same, though their magnitudes vary. Convergence times are highly hardware-dependent, but under superconducting hardware assumptions from before, one can achieve high speed. 

With these potentials in mind, there is a clear approach for how to na\"ively evaluate machine learning models on a thermodynamic computer. First, the initial data is clamped, then information propagates through each of these potentials, followed by an estimation oscillator, before the output is read. One could also consider a relaxed version of these models, where the blocks are directly coupled without estimation oscillators. In this approach, distributions would be propagated forward based on individual samples rather than expectations. This may lead to different training dynamics or trade-offs, which we leave as an open area for exploration.

\subsection{Gradient computation}\label{sec:grad-comp}

Forward evaluation of ML models, while an extremely important step on its own~\cite{agrawal2023sarathi, yuan2024llm}, is only half the picture. A key attribute of our systems is that we can compute gradients as well. Differentiability~\cite{bolte2020mathematical} and backpropagation~\cite{lecun1988theoretical} are the backbone of modern ML training and both continue to be an area of active research~\cite{baydin2018automatic, margossian2019review, schafer2021abstractdifferentiation, moses2021reverse, arya2022automatic, arya2023differentiating}. 

At first glance, it seems like the above deterministic approximations trivially admit differentiation rules in the case where they are truly deterministic. The derivatives are known for the operations they are approximating and training could be conducted in the same way as on a digital computer (but with the outputs of the function/inputs to the next determined by the samples from the thermodynamic computer). However, this relies on the assumption that our approximations are quite close to the deterministic operations. As shown previously, this is not the case, and additional tests showed that these errors compound quickly making training of machine learning workflows based on these approximations unwieldy. Additionally, this relies on the assumption that our deterministic distribution is a known and easy to differentiate distribution. Since one is working in the energy function landscape (and is likely constrained in terms of what energy functions their substrate provides), we focus on a rule which only requires knowledge of the energy function (and its derivative). This allows for a general purpose approach that works with any energy function and doesn't rely on the strength of the deterministic approximation. As we will see later, in Sec.~\ref{sec:thermoformer}, this helps chart a path toward components used in larger models.


Given the EBM structure of our thermodynamic computing paradigm, we can derive additional derivative rules that might be more suitable for implementation on thermodynamic hardware (without resorting to digital simulations)~\cite{mean_field_USPTO, mean_field_PTC}. These rules complement Eqs.~\eqref{eq:train-ebm-cd} and \eqref{eq:train-ebm-cd2}. 
Consider an EBM with dimensionless energy function $E_\theta(y|x)$ and partition function $Z(x,\theta)$. The expected value of $y$ is:
\begin{equation}
\mathbbm{E}[y]= \int y p(y|\theta,x) dy = \int y \frac{e^{-E_\theta(y|x)}}{Z(x,\theta)} dy.
\end{equation}
Taking the derivative of the $i$th component with respect to a parameter $\theta_j$:
\begin{equation}
\frac{\partial \mathbbm{E}[ y_i ]}{\partial \theta_j} = \int y_i \frac{\partial}{\partial \theta_j} \left(\frac{e^{-E_\theta(y|x)}}{Z(x,\theta)}\right) dy.
\label{eq:derivative_start}
\end{equation}
Using the quotient rule, we have:
\begin{multline}
     \frac{\partial}{\partial \theta_j} \left(\frac{e^{-E_\theta(y|x)}}{Z(x,\theta)}\right) = \\
   -p(y|\theta,x)\left(\frac{\partial E_\theta(y|x)}{\partial \theta_j} + \frac{1}{Z(x,\theta)}\frac{\partial Z(x,\theta)}{\partial \theta_j}\right).   
   \label{eq:quotient rule}
\end{multline}
Substituting Eq.~\eqref{eq:quotient rule} back into Eq.~\eqref{eq:derivative_start} leads to
\begin{align}
\frac{\partial \mathbbm{E}[ y_i ]}{\partial \theta_j} &= -\int y_i p(y|\theta,x) \left(\frac{\partial E_\theta(y|x)}{\partial \theta_j} - \mathbbm{E}\left [ \frac{\partial E_\theta(y|x)}{\partial \theta_j}\right ] \right) dy \notag\\
&= -\left(\mathbbm{E}\left[ y_i\frac{\partial E_\theta(y|x)}{\partial \theta_j}\right] - \mathbbm{E}[ y_i ] \mathbbm{E} \left [ \frac{\partial E_\theta(y|x)}{\partial \theta_j}\right ] \right).
\end{align}
The Jacobian is thus given by the negative cross covariance:
\begin{equation}
\frac{\partial \mathbbm{E}[ y ]}{\partial \theta} = -\text{Cov}\left(y, \frac{\partial E_\theta(y|x)}{\partial \theta}\right).
\label{eq:cov_theta}
\end{equation}
This form connects parameter gradients to statistical correlations in the system, providing a method for computing derivatives through sampling. Similarly, we find
\begin{equation}
\frac{\partial \mathbbm{E}[ y ]}{\partial x} = -\text{Cov}\left(y, \frac{\partial E_\theta(y|x)}{\partial x}\right).
\label{eq:cov_x}
\end{equation}

The Jacobians in Eqs.~\eqref{eq:cov_theta} and \eqref{eq:cov_x} can be used to forward- and back-propagate derivative information. This chain rule applies to a computation graph in which the mean of one block is clamped as the deterministic input of the next block. For instance, in a computation graph $x \to \mathbbm{E}[ y^{(1)}]  \to \dots \to \mathbbm{E}[ y^{(l)} ]\to \mathbbm{E}[ y^{(l+1)} ] \to \dots \mathbbm{E}[ y^{(L)} ]$,
with $y^{(L)} \in \mathbb{R}$,
we can compute the gradient $\nabla_{\theta^{(l)}} \mathbbm{E}[ y^{(L)} ]$ with respect to parameters $\theta^{(l)}$ in the $l$th EBM as
\begin{equation}
\nabla_{\theta^{(l)}} \mathbbm{E}[ y^{(L)} ] = \frac{\partial \mathbbm{E}[ y^{(L)} ]}{\partial \mathbbm{E}[ y^{(L-1)} ]} \dots \frac{\partial \mathbbm{E}[ y^{(l+1)} ]}{\partial \mathbbm{E}[ y^{(l)} ]} \frac{\partial \mathbbm{E}[ y^{(l)} ]}{\partial \theta^{(l)} }. 
\end{equation}
The involved covariance matrices must be estimated from samples, which has important implications on error propagation. The potentials themselves are approximating deterministic functions, and we also have sample error on the forward pass, and this also introduces sample error on the backwards pass. As we discuss in the next sections, these error rates are manageable with a reasonable number of samples. The covariance-form estimator of Eqs.~\eqref{eq:cov_theta} and \eqref{eq:cov_x} bears some similarity to equilibrium propagation~\cite{scellier2017equilibrium, kendall2020training, stern2021supervised}, which computes gradients in physical networks from the difference between clamped and free phases. These ideas were also explored in Ref.~\cite{massar2025equilibrium} and connects the covariance to the Hessian. Additionally, there may even be advantages to training under noisy/probabilistic/sample-based conditions, as these are used in a variety of traditional machine learning methods from reinforcement learning~\cite{fortunato2018noisy, plappert2017parameter, eberhard2023pink} to uncertainty quantification~\cite{gal2016dropout, gal2017concrete, lockwood2022review} to improving learning dynamics~\cite{srivastava2014dropout, kingma2015variational, shen2017continuous, molchanov2017variational}. 


\section{Modularity \& scalability via thermodynamic hypergraphical models}

\subsection{Representing complex distributions}

Having outlined the construction of basic energy models that allow tuning of the equilibrium distribution and enable sample generation, these models can now be used as primitives for more complex operations. Just as we created interacting nonlinear potentials, it is also possible to couple many different oscillators together. The exact degree of coupling and connectivity, however, may depend on the specific hardware implementation. 

Specifically, one could leverage the composability features of EBMs to construct probabilistic graphical models (PGMs)~\cite{koller2009probabilistic}. We can see that the previously discussed class of averaged models are effectively a special case of these graphical models, specifically directed acyclic graphs (DAGs). PGMs provide a powerful framework for representing complex probability distributions through graph structure. Using the factor graph formalism, we can represent both directed and undirected graphical models~\cite{loeliger2004introduction}, where models are expressed as bipartite graphs $\mathcal{G} = (\mathcal{V}_v, \mathcal{V}_f, \mathcal{E})$. In this representation, variable nodes correspond to the random variables in our distribution while factor nodes represent functions of these variables. This framework allows us to decompose joint distributions into factors of EBMs with $f(i)$ being the set of factors (energy functions) that are connected to variable node $i$, and $n(a)$ denoting the set of variables incident to factor $a$. The Markov blanket of variable $i$ is the set of neighboring variables that share at least one factor with $i$, $\mathrm{MB}(i) = \left(\bigcup_{a \in f(i)} n(a)\right)$ excluding $i$. This results in representation via
\begin{align}
    \pi(x_{\mathcal{V}_v}) 
    &= \frac{1}{Z} \prod_{a \in \mathcal{V}_f} \exp\left[-E_a(x_{n(a)})\right] \\
    &= \frac{1}{Z}\exp\left[-\sum_{a \in \mathcal{V}_f} E_a(x_{n(a)})\right],
\end{align}
where $E_a$ represents the energy function of factor $a$ (note that we drop the parameters $\theta$ merely for simplicity of notation). The normalization constant $Z$ ensures the distribution integrates to 1:
\begin{equation}
    Z = \int \prod_{a \in \mathcal{V}_f} \exp\left[-E_a(x_{n(a)})\right] \prod_{i \in \mathcal{V}_v} d x_i.
\end{equation}

A classic example that illustrates both single-variable factors and pairwise interactions is the Ising model~\cite{cipra1987introduction}. In this model, each variable represents a spin $s^{(i)} \in \{-1,+1\}$, and the probability distribution takes the form:
\begin{equation}
\label{eq:ising-dist}
    \pi(s^{(1)},\dots,s^{(N)}) = \frac{1}{Z}\exp\left(\sum_i h_i s^{(i)} + \sum_{\langle i,j \rangle} J_{ij}s^{(i)}s^{(j)}\right),
\end{equation}
where $h_i$ represents the local magnetic field at site $i$ (single-variable factors) and $J_{ij}$ represents the coupling between spins (pairwise factors). This naturally decomposes into factors:
\begin{equation}
    E_i(s^{(i)}) = -h_is^{(i)} \quad \text{and} \quad E_{ij}(s^{(i)}, s^{(j)}) = -J_{ij}s^{(i)}s^{(j)}.
\end{equation}
Since factors can connect any subset of variables, including single variables, this also notably results in a hypergraph (a graph in which a single edge can connect more than two nodes) structure. Factors connected to single variables often represent prior distributions or local constraints, while factors connecting multiple variables capture interactions. 

This factor graph framework unifies various types of graphical models. Directed graphical models correspond to products of local conditional probability factors, while undirected models correspond to products of positive compatibility functions or energy factors, as exemplified by the Ising model discussed above. The distinction lies in the semantics and normalization of the factors, not simply in whether a factor is symmetric. The hypergraph structure emerges naturally when factors connect arbitrary sets of variables, enabling the representation of complex higher-order dependencies.

This modular structure makes the model's dependency patterns explicit while allowing us to build complex systems from simpler components. The resulting framework not only facilitates efficient inference algorithms that exploit the graph structure~\cite{drton2017structure} but also provides natural pathways for parallel computation and scalable learning. This not only enables us to increase our model's expressivity using basic building blocks, but also might be necessary for the next era of machine learning~\cite{du2024compositional}, as composition and modularity play an increasingly important role in the next era of scaling.

\subsection{Sampling}

Having outlined how to construct factor graphs by combining smaller EBMs as factor building blocks, we can move on to a central challenge in probabilistic modeling, which is computing quantities of interest from our models after they have been specified or trained. This process, known as model evaluation and prediction, can take several forms depending on our goals. We might wish to compute marginal distributions $\pi(x^{(i)})$ for individual variables, useful for understanding the behavior of specific components of our system. Or we might need conditional probabilities $\pi(x^{(i)}|x^{(j)})$ to make predictions about some variables given observations of others. Often, we seek maximum probability configurations, which represent the most probable states of our system. These inference tasks become intractable to compute exactly as our systems grow, due to the exponential growth of the state space and the high-dimensional integrals or sums involved. The factor graph structure, however, suggests natural approaches for approximate inference that exploit the locality of interactions in our models.

Gibbs sampling emerges naturally from the factor graph structure as a method to generate samples from the joint distribution~\cite{gibbs_sampling_USPTO}. The key insight is that, while sampling from the full joint distribution is difficult, sampling a single variable conditioned on all others is often straightforward due to the local structure of factors. This leads to an iterative algorithm that updates variables one at a time, exploring the probability space through a carefully constructed random walk. Although there are regimes in which Gibbs sampling can be advantageous to do digitally~\cite{robert1999monte}, within a single connected thermodynamic chip, the advantage of using something like Gibbs sampling is an open question. It may depend on the ease of hardware measurement, clamping, etc., in addition to whether Gibbs sampling offers any advantages in cases where you have access to a chip that efficiently computes Langevin dynamics over the entire graph (in general, the theoretical effectiveness of Gibbs sampling outside of specific types of graphs remains an open question~\cite{de2015rapidly}). However, in cases of multi-chip models, Gibbs sampling can be a valuable tool, as these chips may have purely digital interconnects.

The conditional distribution for each variable takes a particularly simple form in factor graphs:
\begin{equation}
    \pi(x^{(i)}|x^{(/i)}) \propto \exp\left(-\sum_{a \in f(i)} E_a(x_{n(a)})\right),
\end{equation}
where $x^{(/i)}$ denotes all variables except for $x^{(i)}$ and the sum runs only over factor nodes $a$ that include variable $i$. This local computation makes each update step efficient, as we need only consider factors directly connected to the variable being updated. Traditional Gibbs sampling is entirely made of iterative loops of clamping and sampling, although there are many variants that could potentially be applied to multichip models~\cite{gonzalez2011parallel, terenin2020asynchronous, daskalakis2018hogwild}. In order to draw a sample $x_i \sim \pi(\cdot|x^{(/i)})$ digitally, any sampling method can be used (either exact conditionals if they are known, or a standard MCMC algorithm, in which case this is known as Metropolis-within-Gibbs). For thermodynamic chips, this allows for a hierarchy of sampling, as each variable could in itself be representing a distribution, enabling Gibbs sampling within each Gibbs step one level up. The modular sampling is naturally available at the hardware design level, but can also be controlled through software (since the level at which clamping/sampling is done can be programmed). 
Although we focus on Gibbs sampling in this work, there are other methods of prediction on graphical models, for example, our deterministic program DAGs are conceptually similar to mean field belief propagation~\cite{yedidia2000generalized}.


\subsection{Training}

Training factor graphs extends the methods we discussed in the previous section to handle multiple interacting EBMs. A key difference is that our models often include both observed (visible) and unobserved (hidden) variables. Hidden variables are powerful tools that can capture underlying structure in our data, mediate long-range dependencies, or represent latent factors in our system~\cite{bishop1998latent}. This is equivalent to training according to Eq.~\eqref{eq:train-ebm-cd2}. Computing these expectations now requires marginalizing over hidden variables. The training algorithm typically alternates between inference (or state estimation) and parameter updates.

A key advantage of factor graphs is their modular nature, which one could exploit during training. For large graphs, one might first train individual factors or small subgraphs independently, then fine-tune the full model jointly, and finally, iteratively scale through curriculum learning. The initialization of both parameters and hidden variables can significantly impact training success. Where possible, we can initialize factors based on domain knowledge or pre-train them on simpler tasks. For hidden variables, initialization strategies often depend on their intended role in the model,  they might be initialized randomly, or based on prior knowledge about the structure we wish to capture.

\section{Toward on-chip thermodynamic self-learning}\label{sec:selflearning}

Thus far, we have treated the model parameters $\theta$ as digitally stored values updated by an external optimizer, while the thermodynamic substrate supplies samples for fixed $\theta$ (cf.~Sec.~\ref{sec:grad-comp}). A natural next step is to ask whether learning itself can be carried out on-chip by promoting parameters to physical degrees of freedom that evolve stochastically, alongside the visible and latent variables. This section sketches one concrete route: a timescale-separated Langevin system in which the fast variables $(x,z)$ rapidly equilibrate for quasi-static $\theta$, while $\theta$ drifts under an effective force that encodes the learning signal.

This idea of inducing learning via adding terms to the Hamiltonian is not dissimilar from recent work on self-training of Ising models~\cite{debos2025learningmultifieldcoherentising}. This connection between self-learning and thermodynamic computation also runs much deeper than practical algorithms \cite{lloyd2025thermodynamics+}. Other approaches have been proposed for local learning of out-of-equilibrium models~\cite{bosch2025local}, and self-learning machines have been proposed for physical devices~\cite{lopez2023self}.

The following Langevin equations, cast in dimensionless form (see Appendix~\ref{app:dimensionless}), describe the dynamics of the system in which there are three sets of equations for each of the $\{x, z, \theta\}$:
\begin{align}
\begin{split}
    d \{x, z, \theta\}_i &= p_{\{x, z, \theta\}i} dt  \\
    d p_{\{x, z, \theta\}i} &= \Bigl[-\partial_{\{x, z, \theta\}}U(x, z, \theta) - \zeta_i p_{\{x, z, \theta\}i}\Bigr]dt \\ &+ \sqrt{2\zeta_i}\xi_i(t) dt.
\end{split}
\end{align}

In the above equations, the \(x_j\) are the visible variables, the \(z_\ell\) are the latent variables, and the \(\theta_i\) are the parameters.
It is assumed that the dynamics of the parameters is much slower than those of the variables.
In this timescale-separated scenario, a Born-Oppenheimer-like approximation holds, in which effective equations of motion for the (slow) parameter can be derived~\cite[Sect.~8.3]{risken1989fokker}.
To lowest order in the ratio of timescales, they are given by
\begin{align}\label{eq:langevin-nonlinear}
\begin{split}
    d \theta_i &= p_i dt,
    \\
    d p_i &= \left ( F^\text{BO}_i(\theta) - \zeta_i p_i \right) dt  + \sqrt{2\zeta_i} \xi_i(t) dt,
\end{split}
\end{align}
In these effective equations of motion, there appears the effective Born-Oppenheimer force
\begin{equation} \label{eq:born-oppenheimer-force}
    F^\text{BO}_i(\theta) \coloneq - \int \rd^{d_z} z \ \rd^{d_x} x \ \frac{\re^{-U(x, z, \theta)}}{Z(\theta)} \pderiv{U(x, z, \theta)}{\theta_i},
\end{equation}
where $Z(\theta) \coloneq \int \rd^{d_z} z \ \rd^{d_x} x \ \re^{-U(x, z, \theta)}$.
This expression for $F^\text{BO}(\theta)$ affords a clear intuitive picture: the force on the parameter degrees of freedom \(\theta_i\) is the (negative) gradient of the potential, averaged over the instantaneous equilibrium for the fast variables \(x_j\) and \(z_\ell\).
Note that $F^\text{BO}(\theta)$ comprises both real forces acting on the parameter degrees of freedom \(\theta_i\), and additional contributions owing to the couplings with \(x_j\) and \(z_\ell\).

The key observation is that $F^\text{BO}(\theta)$ provides a means of wielding the kinematics of a physical system to measure the terms in the learning rule of Eq.~\eqref{eq:train-ebm-cd2}, provided one uses a sufficiently short measurement window during which $\theta$ remains approximately constant.
Indeed, if both \(x_j\) and \(z_\ell\) are unclamped, then $F^\text{BO}(\theta)$ yields \(
    -\mathbbm{E}_{(x, z) \sim \pi_\theta} 
    \bigl[ \nabla_{\theta} E_{{\theta}}(x, z) \bigr]
\), whereas if \(x_j\) is clamped to a specific datum sample \(x\), $F^\text{BO}(\theta)$ yields \(
   -\mathbbm{E}_{z \sim \pi(\cdot|x,\theta)} 
   \bigl[\nabla_{\theta} E_{{\theta}}(x, z)\bigr]
\). Thus if we can construct such a system, we have a means for estimating the gradients on the chip allowing for more efficient, self-learning like training.

We now expand on how one may go about this.
As noted above, we limit ourselves to a short time window during which the (slow) parameters \(\theta\) do not have time to change significantly.
With this assumption, we are free to linearize the dependence of $F^\text{BO}(\theta)$ on \(\theta\), which yields equations of motion that are exactly solvable (specifically, those of an Ornstein–Uhlenbeck process)~\cite[Sect.~3.2]{risken1989fokker} and hold at early times:
\begin{align}\label{eq:langevin-linear}
\begin{split}
    d \theta_i &= p_i dt,
    \\
    dp_i &= \left ( F^0_i + \sum_j J^0_{ij} \bigl(\theta_j - \theta^0_j\bigr) - \zeta_i p_i \right )dt + \sqrt{2\zeta_i} \xi_i(t) dt,
\end{split}
\end{align}
where \(F^0 \coloneq F^\text{BO}(\theta^0)\) and \(J^0_{ij} \coloneq \left. \pderiv{F_i(\theta)}{\theta_j} \right|_{\theta^0}\) is the Jacobian of the force at the initial position.
It is \(F^0\), the effective force at the initial position \(\theta^0\), that we will estimate from the early-time dynamics. See Appendix~\ref{app:full_expressions} for more details.

We now describe two more integrated hardware architectures that might yield a path of self-learning, a natural gradient descent (NGD) approach, and a more fully analog NGD approach via relay oscillators. We consider these approaches more ambitious and speculative, but could yield ideas for future research directions.

Natural gradient descent (NGD)~\cite{martens2020new} preconditions the gradient with the inverse of the information metric, accounting for the curvature of the statistical manifold and yielding updates that are invariant to parameterization. For energy based models, we can define the Fisher information matrix via

\begin{align}\label{eq:fi-metric}
    \mathcal{I}^{\mathrm{FIM}}_{jk}(\theta) &= \mathbbm{E}_{x \sim \pi_\theta}\bigl[\partial_j E_\theta(x)\,\partial_k E_\theta(x)\bigr] \notag \\
    &\quad - \mathbbm{E}_{x \sim \pi_\theta}\bigl[\partial_j E_\theta(x)\bigr]\,\mathbbm{E}_{y \sim \pi_\theta}\bigl[\partial_k E_\theta(y)\bigr],
\end{align}
where $\partial_j \equiv \partial/\partial\theta_j$ and $E_\theta(x)$ is the energy function.  The NGD update rule is then
\begin{equation}\label{eq:ngd-update}
    \theta_{t+1} = \theta_t - \eta_t\bigl(\mathcal{I}^{\mathrm{FIM}}\bigr)^{\!+} \nabla_\theta \mathcal{L}(\theta_t),
\end{equation}
where $\eta_t$ is the learning rate and $(\cdot)^+$ denotes the pseudoinverse.

The key point, building on the Born-Oppenheimer approximation, is that both the gradient vector and all entries of the Fisher matrix can be estimated from rapid momentum measurements on the synapse oscillators of the thermodynamic chip~\cite{gen_1_5_USPTO}. Specifically, repeated momentum readouts at intervals $\delta t$ yield time-averaged estimates of the energy gradients $\partial_j E_\theta$; pairwise products of these measurements then provide the second-moment terms needed for $\mathcal{I}^{\mathrm{FIM}}_{jk}$. In this hybrid protocol, the matrix inversion of Eq.~\eqref{eq:ngd-update} is performed on an external device, the gradient information is merely read off.

A more ambitious approach is to avoid the digital readout and inversion by encoding components directly in auxiliary oscillators. We can imagine a protocol such as the following~\cite{gen_1_75_USPTO, gen_2_USPTO}. First, a set of estimator oscillators is coupled to the energy-gradient observables $\partial_j E_\theta$. Intermediate estimation oscillators sample the energy-gradient products at different times, and a final estimation oscillator, coupled to all intermediates, approximates the expectation value $\mathbbm{E}[\partial_j E_\theta \, \partial_k E_\theta]$. Separate estimation oscillators store the individual gradient expectations $\mathbbm{E}[\partial_j E_\theta]$. A third layer of estimation oscillators, coupled to the outputs of the previous two layers, encodes the full $(j,k)$ component of the Fisher metric via Eq.~\eqref{eq:fi-metric}. A block-diagonal approximation to the Fisher matrix can also be constructed layer by layer, potentially reducing the number of estimation oscillators. If one coupled parameter oscillators to the estimation network through a potential, whose equilibrium dynamics shift each parameter by an amount proportional to the corresponding entry of $(\mathcal{I}^{\mathrm{FIM}})^{+} \nabla_\theta \mathcal{L}$, this would yield the NGD update of Eq.~\eqref{eq:ngd-update} without any digital matrix inversion.

\section{Example architectures \& applications}\label{sec:examples}

Before getting into a physical implementation of the basic building blocks, we highlight a few example applied demonstrations of what energy-based thermodynamic chips could be used for. We also present some further arguments on the time/energy advantages that could exist. We provide a collection of theoretical outlines and numerical simulations to highlight the scope of applicability of these blocks we have detailed above. While we detail the hardware implementation of one block in Sec.~\ref{sec:superconducting}, the focus in this section is on what one could achieve with many of these hardware blocks. These workflows will vary and depend on hardware, but show the generalizability and promise of DAGs and PGMs as expressive models. In this section, to illustrate the enormous scope of possible graphical models that can be assembled using these building blocks, we demonstrate a broad class of more complex tools that can rely on the simple primitives that we demonstrated above, namely: Gaussian PGMs, Gaussian mixture models, Hidden Markov models, continuous Ising models, and a thermodynamic version of the popular transformer model~\cite{vaswani2017attention}. The following numerics were built on JAX~\cite{jax2018github, deepmind2020jax} and equinox~\cite{kidger2021equinox}.

\subsection{Gaussian probabilistic graphical model}

Even when using only simple Gaussian potentials, powerful models can be constructed. Gaussian PGMs originally gained prominence due to the popularity of algorithms such as Gaussian belief propagation~\cite{bickson2008gaussian, su2015convergence, ortiz2021visual}, and recently we have seen their resurgence at the intersection of Gaussian graphical models and neural networks~\cite{satorras2021neural, liang2021neural, patwardhan2022distributing,  ortiz2023gaussian, liang2023neural}. Gaussian graphical models are among the simplest examples of PGMs possible within our framework. One can simply couple together Gaussian oscillators (with any degree of connectivity that their hardware supports, adding latent variables to make up for the sparsity), and train them using the standard visible or hidden contrastive losses presented above. 

\begin{figure}[htbp]
    \centering
    \includegraphics[width=0.9\linewidth]{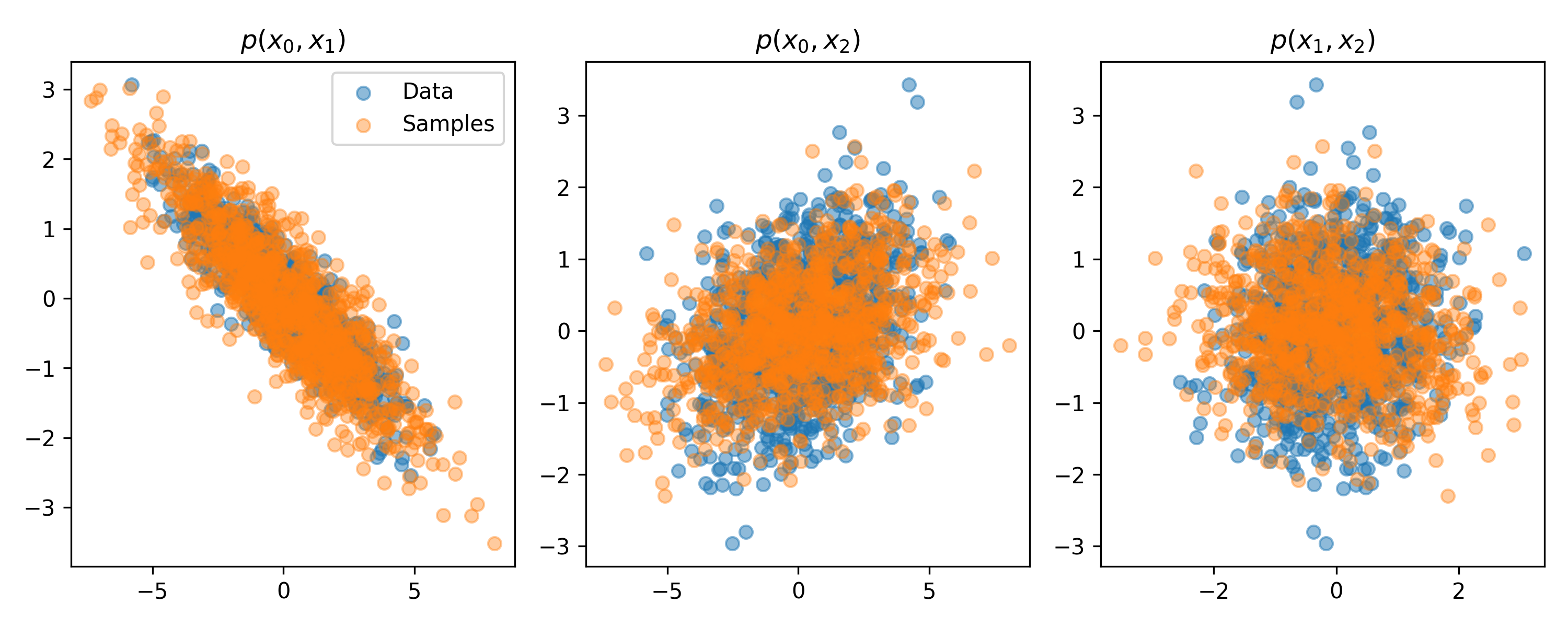}
    \caption{Samples vs. data for a trained Gaussian PGM.}
    \label{fig:gauss-pgm}
\end{figure}

A toy example of using Gaussian PGMs as a generative model is shown in Figure~\ref{fig:gauss-pgm}. In this example, we show slices of a three-dimensional Gaussian (since any Gaussian graphical model can be represented as a single Gaussian), represented using three one-dimensional Gaussian nodes, trained using the standard CD-learning rules [cf.~Eq.~\eqref{eq:train-ebm-cd}]. Here, we use Gibbs sampling to draw samples from the Gaussian PGM.

\subsection{Gaussian mixture model}

As a simple but highly relevant example, we construct a Gaussian mixture model (GMM) based on a factor graph using the building blocks outlined above. GMMs are a powerful class of models~\cite{dandi2024universality, shi2024universal} which rely on two key components: a mixing distribution, in this case a categorical distribution, and a component distribution, which is a Gaussian of the form of Eq.~\eqref{eq:gauss_general}. Here, we use a hard one-hot selector $z \in \{e_1, \ldots, e_K\}$ rather than a finite-$\lambda$ continuous relaxation; this mixture model therefore does not require any estimation oscillators.

Given a dataset $X = \{x^{(1)}, \dots , x^{(N)}\}$, where each $x^{(i)}$ is a $D$-dimensional vector, our goal is to model the probability distribution that generated this data using a mixture of Gaussian distributions with $K$ components
\begin{align}
    \pi_\theta({x})  =  \sum_{k=1}^K {\rm Cat}_{\eta}(z = k) \mathcal{N}(x|{\mu}^{(k)}, {\Sigma}^{(k)}),
\end{align}
with logits $w = (w_1, \dots, w_K)$ and mixture weights $\eta = \text{softmax}(w)$, so that $\sum_k \eta_k = 1$, means $\mu = ({\mu}^{(1)}, \dots, {\mu}^{(K)})$, and covariances $\Sigma = (\Sigma^{(1)}, \dots, \Sigma^{(K)})$. The (marginal) log-likelihood of the data given the model parameters $\theta = (w, \mu, \Sigma)$ is
\begin{equation}
    \mathcal{L}(\theta) = \sum_{i=1}^N \log \pi_\theta({x^{(i)}}).
\end{equation}

\begin{figure}[htbp]
    \centering
    \includegraphics[width=0.9\linewidth]{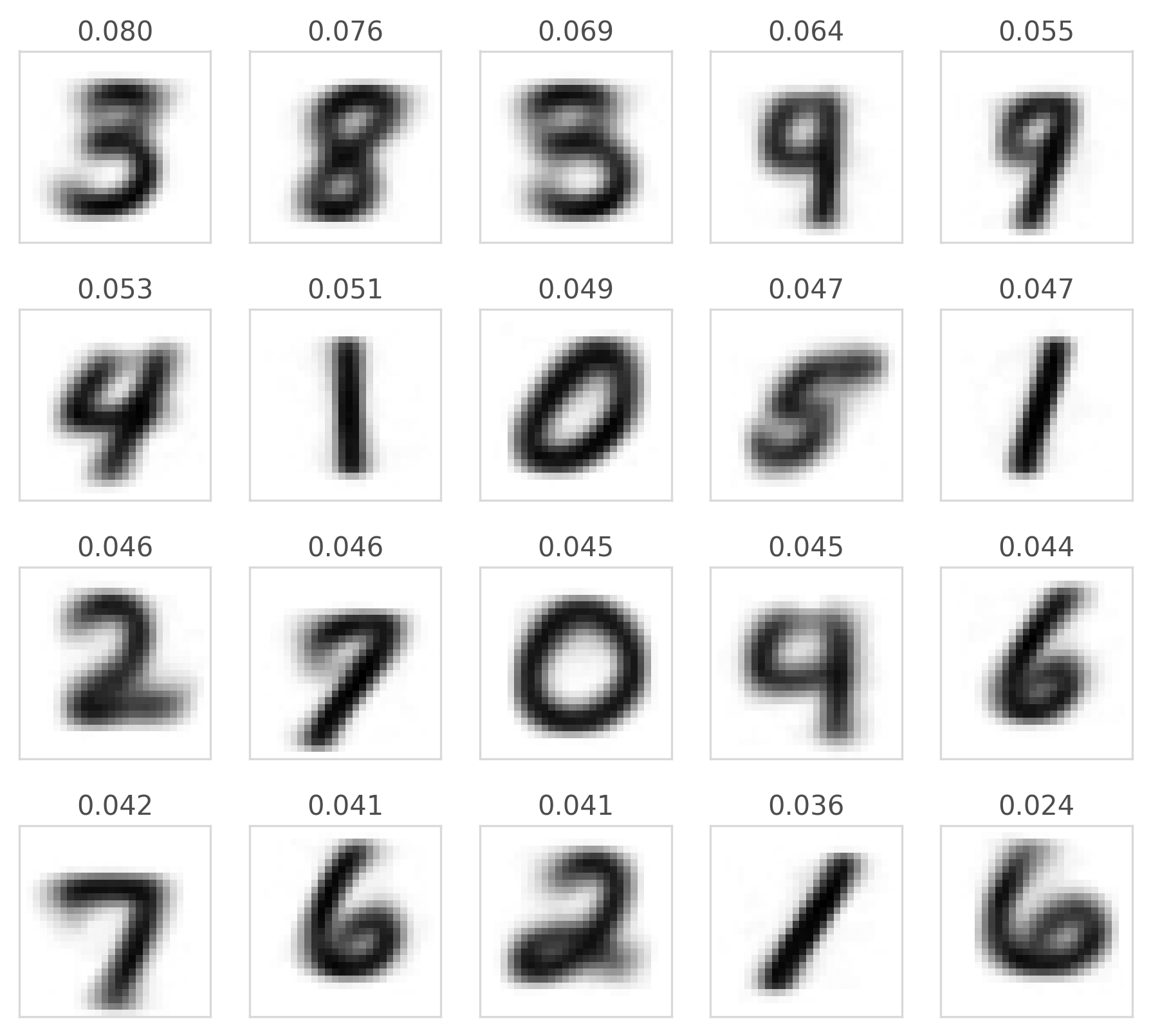}
    \caption{Components of a (purely digital) Gaussian Mixture Model trained on MNIST. The title of each figure represents the associated mixing probability of each component.}
    \label{fig:gmm}
\end{figure}

We can use gradient ascent to maximize the log-likelihood. The GMM can be represented as a factor graph with $K$ Gaussian EBM nodes defined through Gaussian potentials
\begin{equation}
    U_G(x; \mu_k, \Sigma_k) = \frac{1}{2}(x - \mu_k)^\top \Sigma_k^{-1}(x - \mu_k) + \frac{1}{2}\log\det(2\pi\Sigma_k),
\end{equation}
representing the clusters, and one categorical EBM node representing the mixture component selection, for which we use the softmax potential [Eq.~\eqref{eq:softmax}]. Finally, in the hard one-hot limit, these components are connected via a factor energy:
\begin{multline}
    U_\theta(x, z) = \lambda_1 \sum_{i=1}^K z_i^2(z_i - 1)^2 - \sum_{i=1}^K w_i z_i \\
    + \lambda_2 \left(\sum_{i=1}^K z_i - 1\right)^2 + \sum_{k=1}^K z_k U_G(x; \mu^{(k)}, \Sigma^{(k)}).
\end{multline}
As a motivating illustration of this graphical structure, we show a conventional digital GMM with 20 Gaussians trained using stochastic gradient descent on the negative log-likelihood of the full MNIST dataset in Figure~\ref{fig:gmm}. The GMM numerics were computed using distreqx~\cite{lockwood2024distreqx}.

\subsection{Hidden Markov model}

Hidden Markov models (HMMs) are a specific type of state space model characterized by discrete hidden states and observations that are conditionally independent given these states. HMMs were a backbone of pre-neural network machine learning~\cite{rabiner1989tutorial, eddy2004hidden, krogh1994hidden, mamon2007hidden} and have seen more recent usage interfacing with neural networks~\cite{he2018unsupervised, liu2019powering, ghosh2021normalizing, azeraf2021introducing, gangloff2021general, gangloff2023deep}. In general, a state space model consists of latent state variables, observation variables, a transition model, and an observation model. HMMs have a finite set of hidden states, while the observations may be discrete or continuous. In contrast to general state space models, the observation model of the HMM $\pi(x^{(t)} | z^{(t)})$ is not conditioned on previous observations $x^{(t-1)}$. The state variables $z^{(t)}$ represent the hidden state of the system at time $t$, while the observation variables $x^{(t)}$ are the observed data at time $t$. The transition model $\pi(z^{(t)}|z^{(t-1)})$ defines the probabilistic rules for transitioning from one state to another, and the observation model $\pi(x^{(t)}|z^{(t)})$ defines the probabilistic relationship between the state and the observation. The probability of the initial hidden state is defined by a prior $\pi(z^{(1)})$. Building an HMM with EBMs involves defining energy functions for the transition and observation model, and the prior. The latent state variables $z^{(t)}$ and the observation variables $x^{(t)}$ are modeled using energy functions that capture the dynamics and the relationship between states and observations.

The transition model in an HMM framework defines the probability of transitioning from one state to another. In the context of EBMs, we define an energy function $E_{\theta}^{\text{trans}}(z^{(t)}, z^{(t-1)})$ for the transition between states $z^{(t-1)}$ and $z^{(t)}$. The transition probability can then be expressed as:
\begin{equation}
\label{eq:joint-hmm}
   \pi_{\theta}(z^{(t)} | z^{(t-1)}) = \frac{e^{-E_{\theta}^{\text{trans}}(z^{(t)}, z^{(t-1)})}}{Z_{\text{trans}}(z^{(t-1)})} 
\end{equation}
where $Z_{\text{trans}}(z^{(t-1)}) = \sum_{z^{(t)}} e^{-E_{\theta}^{\text{trans}}(z^{(t)}, z^{(t-1)})}$ is the partition function and $\theta$ are the parameters of the transition model.

The observation model defines the probability of observing $x^{(t)}$ given the state $z^{(t)}$. Using EBMs, we define an energy function $E_{\phi}^{\text{obs}}(x^{(t)}, z^{(t)})$ that captures the relationship between the observations and the state. The observation probability is given by:
\[
\pi_{\phi}(x^{(t)} | z^{(t)}) = \frac{e^{-E_{\phi}^{\text{obs}}(x^{(t)}, z^{(t)})}}{Z_{\text{obs}}(z^{(t)})}
\]
where $Z_{\text{obs}}(z^{(t)}) = \sum_{x^{(t)}} e^{-E_{\phi}^{\text{obs}}(x^{(t)}, z^{(t)})}$ is the partition function and $\phi$ are the parameters of the observation model. For continuous emission models, sums over $x$ are replaced by integrals.

The joint probability of the state sequence $z^{(1:T)}$ and the observation sequence $x^{(1:T)}$ in a state space model using EBMs can be expressed as:
\begin{multline}
\label{hmm:full-joint}
    \pi(z^{(1:T)}, x^{(1:T)}) = \\ \pi_{\alpha}(z^{(1)}) \prod_{t=2}^{T} \pi_{\theta}(z^{(t)} | z^{(t-1)}) \prod_{t=1}^{T} \pi_{\phi}(x^{(t)} | z^{(t)})
\end{multline}

where the initial state prior is:
\begin{equation}
\label{eq:hmm-prior}
    \pi_{\alpha}(z^{(1)}) = \frac{e^{-E_{\alpha}^{\text{init}}(z^{(1)})}}{Z_{\text{init}}}.
\end{equation}

As for any other EBM model, the optimization objective for the HMM is the likelihood of the observed data: $\log p(x^{(1:T)}) = \log \sum_{z^{(1:T)}} \pi(x^{(1:T)}, z^{(1:T)})$.

Because of the conditional probabilities of the transition and observation model, the gradient is different from the hidden and fully visible PGMs that have been discussed previously in this section. The gradient with respect to the parameters $\theta$ of the transition model reads:
\begin{align}
& \sum_{x^{(1:T)} \in \mathcal{B}} \nabla_{\theta} \log \pi(x^{(1:T)}) = \notag \\
&\sum_{x^{(1:T)} \in \mathcal{B}}
\mathbbm{E}_{z^{(1:T)} \sim \pi(\cdot \vert x^{(1:T)})}
\Biggl[
\sum_{t=2}^{T} \Bigl(
-\nabla_{\theta} E_{\theta}(z^{(t)}, z^{(t-1)}) \notag \\
&
+ \mathbbm{E}_{z'^{(t)} \sim \pi_{\theta}(\cdot \vert z^{(t-1)})}
\left[
\nabla_{\theta} E_{\theta}(z'^{(t)}, z^{(t-1)})
\right]
\Bigr)
\Biggr].
\end{align}

Hence, for a batch of data trajectories $\mathcal{B}$, the hidden trajectories are sampled $z^{(1:T)}$ and the positive phase is evaluated with these trajectories. For the negative phase we also sample $z'^{(t)} \sim \pi(\cdot \vert z^{(t-1)})$, where we use the shorthand $z'^{(1:T)} \sim \pi( \cdot \vert z^{(1:T)})$ that implies that the conditional $z^{(1:T)}$ comes from the previously sampled trajectory.

The gradient for $\phi$ reads:
\begin{align}
& \sum_{x^{(1:T)} \in \mathcal{B}} \nabla_{\phi} \log \pi(x^{(1:T)}) = \notag \\
&\sum_{x^{(1:T)} \in \mathcal{B}}
\mathbbm{E}_{z^{(1:T)} \sim \pi(\cdot \vert x^{(1:T)})}
\Biggl[
\sum_{t=1}^{T} \Bigl(
-\nabla_{\phi} E_{\phi}(x^{(t)}, z^{(t)}) \notag \\
&
+ \mathbbm{E}_{x'^{(t)} \sim \pi_{\phi}(\cdot \vert z^{(t)})}
\left[
\nabla_{\phi} E_{\phi}(x'^{(t)}, z^{(t)})
\right]
\Bigr)
\Biggr].
\end{align}

\begin{figure}[htbp]
    \centering
    \includegraphics[width=0.9\linewidth]{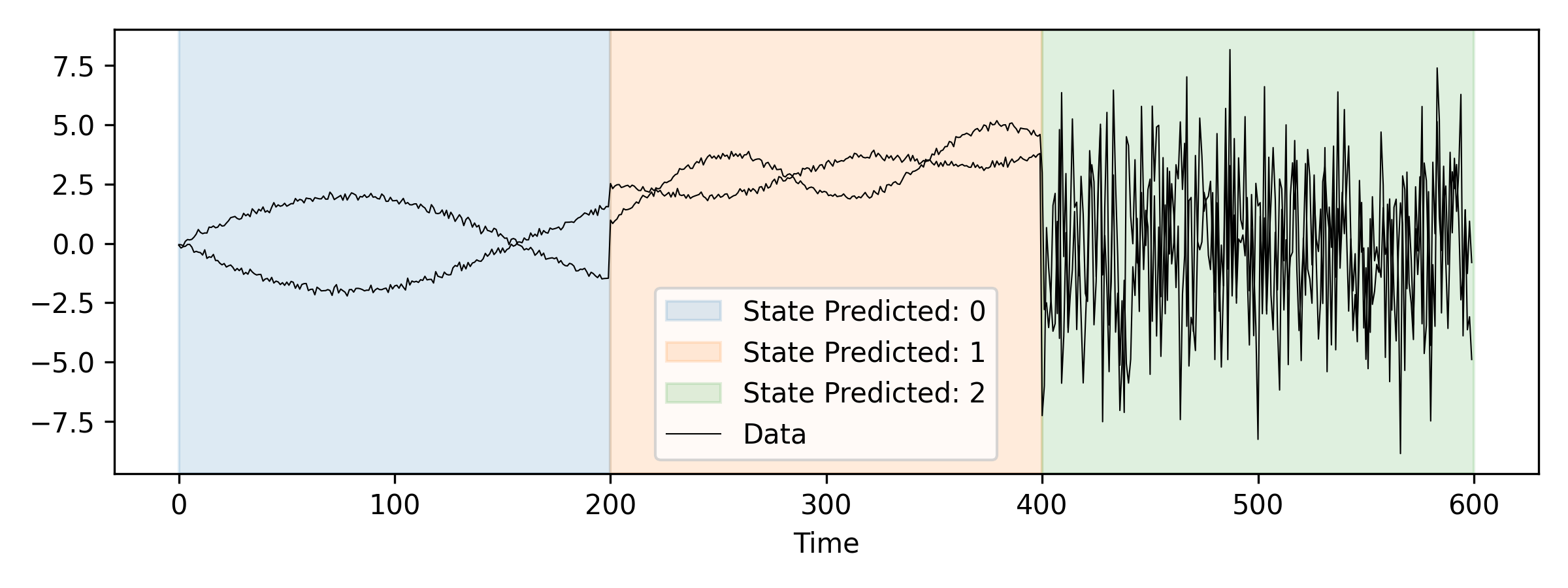}
    \caption{Training a 2D HMM on time series data. Highlighted regions indicate predicted states.}
    \label{fig:hmm-pred}
\end{figure}

The full derivation for these gradient expressions is available in Appendix~\ref{ap:hmm}. As an illustrative example, we show a fully digital Gaussian HMM where $\pi_\theta(x^{(t)} | z^{(t)}) = \mathcal{N}(x^{(t)} \mid \mu_{z^{(t)}}, \Sigma_{z^{(t)}})$ on a simple synthetic time-series dataset. We fit the model using gradient descent and plot the predicted states in Figure~\ref{fig:hmm-pred}. This example was created using dynamax~\cite{dynamax}. This example highlights the potential of PGMs for time-series data.

\subsection{Ising model}

As a further demonstration of the power of continuous EBMs as building blocks for PGMs, we train a continuous Ising model. The Ising model is a universal computing class which has been studied extensively by the physics community~\cite{cipra1987introduction, brush1967history} and is of great interest in machine learning and optimization as well~\cite{patel2020Ising, mohseni2022Ising, ackley1985learning, nikhar2024all}. As previously discussed, Ising machines (hardware implementations of Ising models) are a prominent example of analog hardware. There are many extensions and variations that increase the trainability~\cite{bresler2015efficiently, lokhov2018optimal, laydevant2024training, jelinvcivc2025efficient} and expressivity of Ising models, often relying on using hidden/latent variables~\cite{salakhutdinov2009deep, dunn2013learning, nussbaum2019ising}. 

\begin{figure}[htbp]
    \centering
    \includegraphics[width=0.7\linewidth]{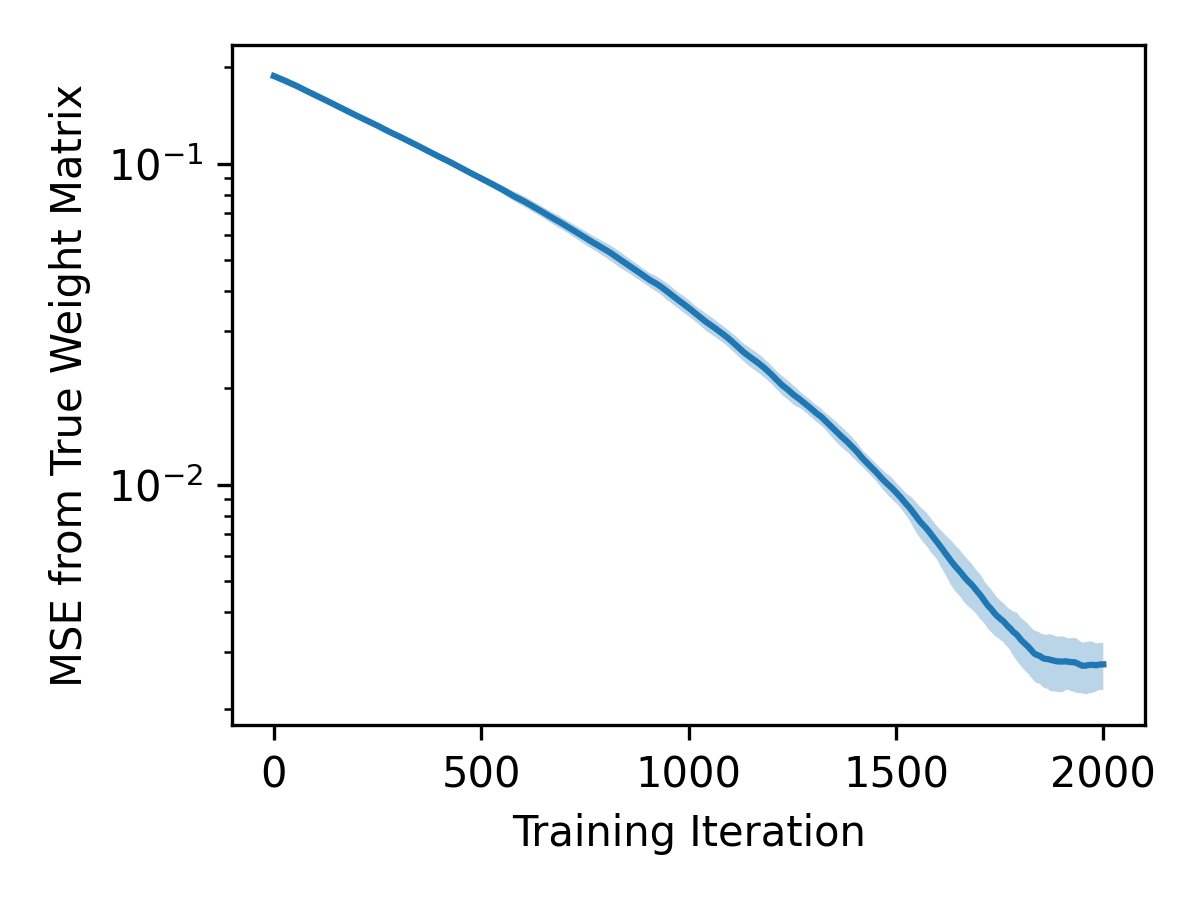}
    \caption{Training a continuous Ising machine. Mean-squared error between the parameter matrix and  the optimal set of parameters $\theta^*$ as a function of the training iterations.}
    \label{fig:Ising-train}
\end{figure}

Although physical implementations, such as superconductors, lack all to all connectivity, this is common in analog hardware platforms and existing work seeks to address the costs and benefits of these sparse Ising models~\cite{niazi2024training, nikhar2024all, sajeeb2025scalable, jelinvcivc2025efficient}. We can relax the ubiquitous discrete model to be continuous, i.e., we can replace the binary variables in Eq.~\eqref{eq:ising-dist}. In this case, the model has an energy function $E_\theta(s) = \sum_i \left ( s_i^2  (s_i^{\phantom{2}} - 1)^2 - hs_i^{\phantom{2}} \right) + \sum_{\langle i,j \rangle} J_{ij}s_is_j$, with $s_i$ being a continuous value. This is analogous to an Ising model but with continuous double-wells at each state. We train a fully visible version of this continuous Ising model on a simple bars and stripes dataset. Here, each node of the Ising model represents one corner of the four squares of the bar/strip grid. We use the standard CD-based training methods [Eq.~\eqref{eq:train-ebm-cd} and Eq.~\eqref{eq:train-ebm-cd2}] with our continuously relaxed Ising machines and show the convergence of the weight matrix to the ground truth in Figure~\ref{fig:Ising-train}. Although continuous Ising models have been studied previously to some extent~\cite{nishikawa1976continuous, van1978phase, bayong1999effect}, there is substantial room for further investigation.

\subsection{Thermoformer: Thermodynamic Transformer}\label{sec:thermoformer}

With matrix-vector products (and, by extension, matrix multiplications, as they consist of multiple matrix-vector products), vector additions, softmax, and a non-linear activation, we have a suite of core machine learning components. A common activation used in modern models is the swish activation~\cite{ramachandran2017searching}, $\text{swish}(x) = x\sigma_{\rm{ML}}(x)$, which can trivially be assembled from our previously outlined building blocks (multiplication and sigmoid)~\cite{swish_gadget_USPTO}. These are sufficient for examples such as MLPs or CNNs. However, if we wish to scale to state-of-the-art architectures, a few additional tools are required. The primary focus here is on one of the most common and widely used architectures in modern machine learning: the transformer-based~\cite{vaswani2017attention} decoder model~\cite{radford2019language}. 

The main component of the transformer architecture that we have not yet specified is the layer norm~\cite{ba2016layernormalization, xiong2020layer}. Layer norm computes $f(x) = \frac{x - \mathbbm{E}[x]}{\sqrt{\text{Var}[x] + \varepsilon}}$. This computation relies on the following: the components of addition (which we have specified), computing the mean (which we have also specified), as well as computing the variance and computing the division, which we specify presently. 

To compute the variance, one could construct the potential to compute the variance of $z$ oscillators (where $z \in \mathbb{R}^N$ and $y \in \mathbb{R}$) via 
\begin{equation}
\label{eq:var-gadget}
 U_\theta(x, y, z) = \frac{1}{2}y^2 + \frac{1}{2} \sum_i (x_i - (z_i - \mu))^2 - \frac{1}{N - 1}y\sum_i x_i^2,
\end{equation}
where $x_i$ are intermediate, auxiliary oscillators. In the deterministic limit, the stationary point of this potential is at $y = \text{Var}[z]$. Then, using the potential $U(y, z) = y^3(z + \varepsilon) - 3y$, which (assuming a 0 initialization, and $x > 0$, and $z = \text{Var}[x]$) has its stationary point at $y = \frac{1}{\sqrt{\text{Var}[x] + \varepsilon}}$. This layer norm potential is merely an example and it requires additional stable nonlinear primitives, and is not meant to be an equilibrium realization of the function as this particular example is unbounded. There is an abundance of architectural design and implementation choices that could be used instead. For example, thermodynamic computers may be more amenable to transformers with other local nonlinearities~\cite{zhu2025transformers, chen2025strongernormalizationfreetransformers, leroux2025analog}, since we have already seen how to do sigmoid, instead of layer norm. 

\begin{figure}[htbp]
    \centering
    \includegraphics[width=0.45\linewidth]{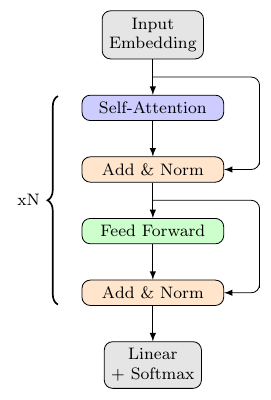}
    \caption{Diagram of transformer-decoder model.}
    \label{fig:decoder-diagram}
\end{figure}

With these components in hand, we can now see how to assemble a transformer in thermodynamic hardware (what we call a ``thermoformer"). Embed the tokens with position embeddings digitally, clamp oscillators to these values, go through the attention mechanism (where $\text{attention} = \text{softmax}\left (\frac{QK^T}{\sqrt{d}}\right )V$ where $Q, K, V$ are the results of inputs multiplied by different weight matrices, in the following we assume all to all connectivity for the softmax, but that is not strictly required and with extra degrees of freedom the same output can be obtained~\cite{thermo_transformer_USPTO} with an increase in the time and energy), add a residual connection~\cite{he2016deep} and norm, and stack these layers to a certain depth. A visualization of the decoder module is shown in Figure~\ref{fig:decoder-diagram}. There are an endless number of transformer variants~\cite{lin2022survey, khan2023survey, min2022transformer}, which could also be implemented if one constructed approximate potentials, but here we focus on the original. Naturally, this approach also scales to transformer variants such as mixture of experts based models~\cite{shazeer2017outrageously,fedus2022switch,mixture_of_experts_USPTO, selection_of_experts_USPTO}.

\begin{figure}[htbp]
    \centering
    \includegraphics[width=0.99\linewidth]{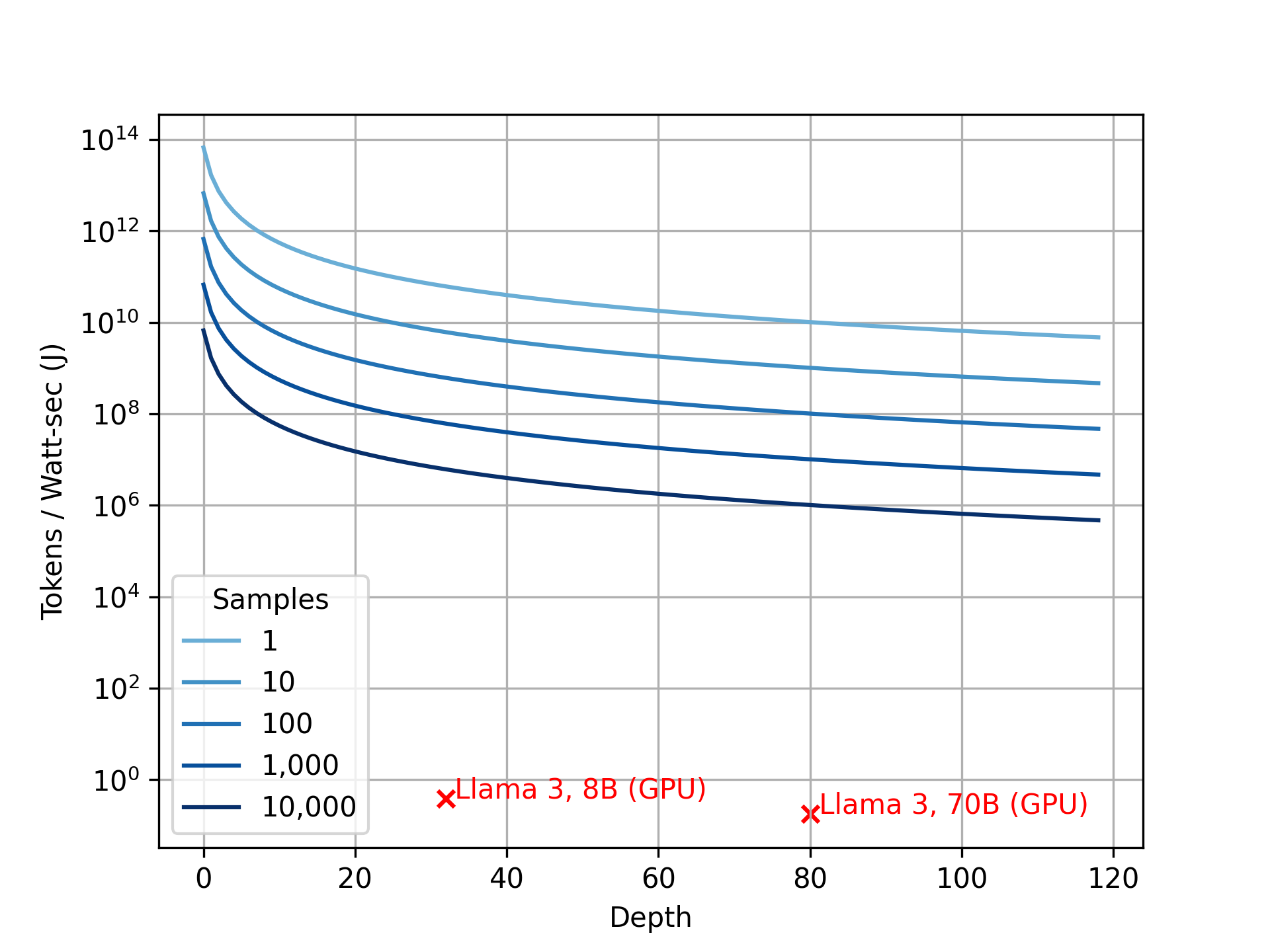}
    \caption{Tokens per Joule for varying decoder depths and varying numbers of samples used. Llama models on H100 GPUs included for reference. Projected chip results exclude cryogenic cooling, control electronics, and calibration overhead.}
    \label{fig:thermo-watt}
\end{figure}

With these potentials constructed, we can now approximate the time and energy values of interest for a superconducting circuit implementation. Using the same approach as before, we can scale these approximations due to the sequential nature of the DAG. Specifically, since each operation must first thermalize, then have its expectation computed before the next operation can thermalize, the total time can be obtained by summing the thermalization times of each block. Similarly, the total energy can be computed by summing the energy contributions from each step. With values that are not unreasonable in current day superconducting hardware (100~fF, 100~pH, 100~$\Omega$, operating at 50~mK, with $\lambda$ parameters of 1), the resulting chip projections suggest potentially favorable tokens per joule under the assumed superconducting device parameters. Decoder models of varying depth are displayed in Figure~\ref{fig:thermo-watt}, as well as some of the prominent semi-open-weight models, Llama 3~\cite{dubey2024llama}. Note that, regardless of whether the samples are in parallel or sequential they require additional (linearly scaling) energy. 

\begin{figure}[htbp]
    \centering
    \includegraphics[width=0.99\linewidth]{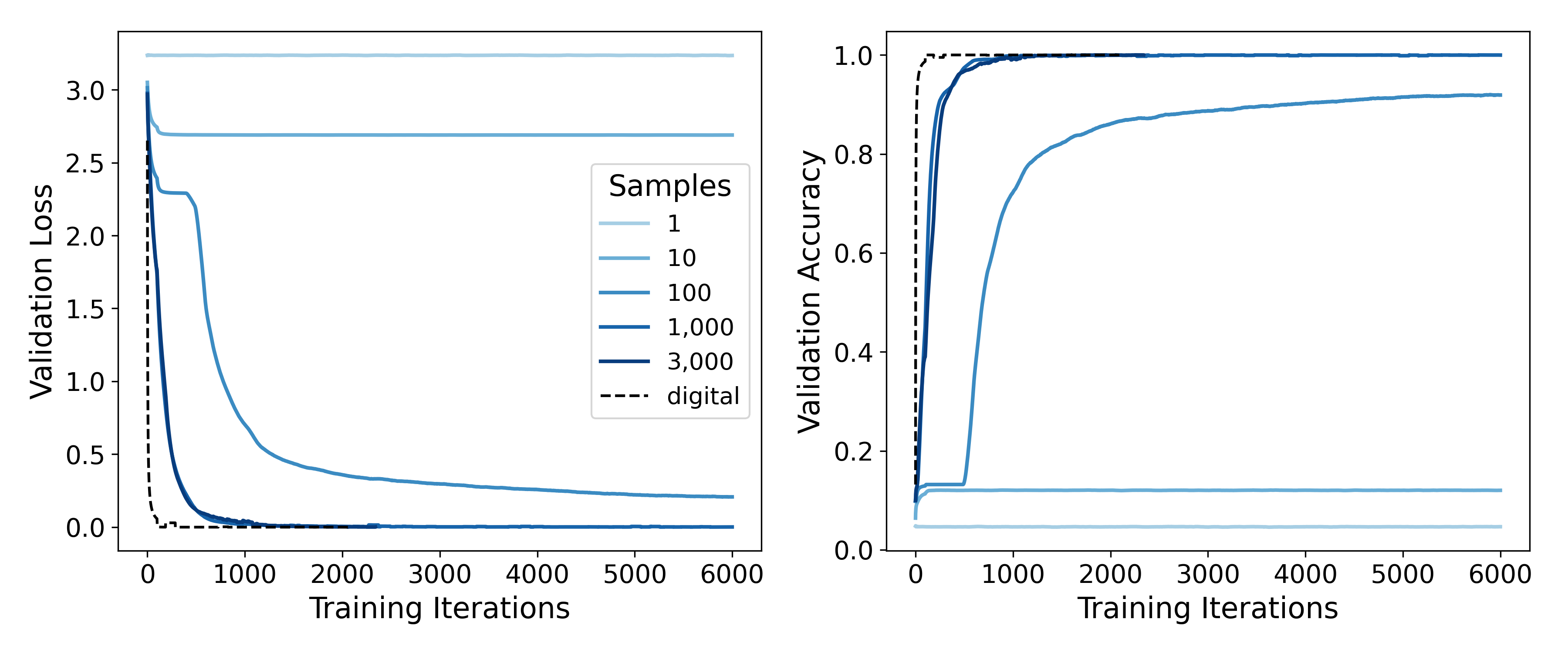}
    \caption{Training of a thermoformer with different numbers of samples used to compute the mean of each operation, compared to a digital decoder.}
    \label{fig:thermo-perf}
\end{figure}

However, this is an incomplete picture. Because our thermoformer is conducting sample-based approximations to potentials that approximate the deterministic operations, the performance is not the same as the deterministic digital case. We show the performance of thermoformers with varying numbers of samples used to approximate the means compared to a digital decoder, tested on a toy dataset of sequential numbers. The results are shown in Figure~\ref{fig:thermo-perf}. As we can see, within a reasonable number of samples, on this toy task, the simulated model approaches the digital baseline as the number of samples increases. This approach uses the gradient estimation rules outlined in Sec.~\ref{sec:grad-comp}. Additionally, the thermodynamic operations that we used have parameters that make their approximations quite loose (that is to say, even when having converged to the true expected value of the distributions, that value is not necessarily all that similar to the true deterministic value). For example, the $\lambda$ values of the softmax computation [Eq.~\eqref{eq:softmax}], are small ($ = 1.0$), which results in values that have some qualitative similarities to the output distribution (e.g., the biggest logit will be the biggest probability), but are not necessarily quantitatively similar (e.g., the outputs may not be strictly positive, or sum to one). How the required number of samples scales with model depth, sequence length, and task difficulty is an open question that we do not address here. Even in this approximate regime, we are able to achieve good performance on this small problem. 

\section{Implementation of building blocks in a superconducting substrate}
\label{sec:superconducting}

To demonstrate the power of the theory described above, we designed and fabricated a superconducting chip that implements the core double-well building block, a \tnode{}~\cite{sc_neuron_USPTO}. This is the first step towards building a fully scaled up energy-based thermodynamic computer. In addition, we present an achievable set of next steps and future experiments to help the broader research community advance this technology, noting that our implementation of this design represents but the first few steps in this new direction, and that much innovative research and development is still required. The experiment described in this section characterizes the simplest elemental block of the framework, a single tunable double-well potential. This fundamental thermodynamic building block can be seen as a continuous version of the discrete probabilistic-bit (and in fact is currently limited to binary readout). By taking the two wells as the two binary positions, a double well \tnode{} can be converted into a probabilistic-bit~\cite{camsari2017implementing, freitas2021stochastic, yang2025250, rhee2023probabilistic, jelinvcivc2025efficient}.

Superconducting circuits offer a fundamental nonlinear element, the Josephson junction~\cite{josephson1962possible, josephson1974discovery}. This non-dissipative element exhibits a sinusoidal relationship between the voltage and current across its two ports. When used in conjunction with the quadratic term from an inductance, this allows us to engineer a system with a double-well potential. The Josephson junction also allows us to engineer nonlinear coupling between nodes. Furthermore, on-chip dissipation in superconducting circuits is naturally very low and allows information processing much closer to the Landauer limit~\cite{saira2020nonequilibrium}. Additionally, the associated temperature and energy regimes allow us to harness ambient thermal fluctuations, and as such, we do not require the injection of noise algorithmically. Networks with thousands of similar building blocks coupled together have already been implemented for the purpose of quantum annealing~\cite{king2023quantum}.

The \tnode{} is a tunable nonlinear system. It is akin to a superconducting flux qubit or fluxmon~\cite{quintana2017superconducting, harris2010experimental, novikov2018exploring, khezri2021anneal}, which are normally used for quantum computing, but it is engineered to operate in the thermodynamic domain, where its dynamics are thermally activated and its time evolution can be modeled by the Langevin equation [Eq.~\eqref{eq:underdamped_langevin}]~\cite{buttiker1983thermal, han1992effect, saira2020nonequilibrium, pratt2025extracting}. We focus on its double-well regime and demonstrate the ability to tune the equilibration times through the control of the potential and temperature.

\subsection{Device theory and design: From thermodynamic neurons to thermodynamic chips}
\label{sec:design}

The device is composed of three \tnode{}s. More precisely, the device is a superconducting aluminum-on-silicon chip hosting three uncoupled \tnode{}s, each having two dedicated control lines. Each \tnode{} is inductively coupled to a coplanar-waveguide (CPW) $\lambda/4$ readout resonator, which is in turn coupled capacitively to a shared CPW transmission line. The resonance frequencies of the three readout resonators are designed to be between 11.7 and \qty{11.9}{\GHz} spaced \qty{100}{\MHz} apart for frequency addressability. The chip contains two \tnode{} design variants: one coplanar and one lumped element. In this article, we focus on the coplanar variant.

\begin{figure}[t]
    \centering
    \includegraphics[width=0.7\linewidth]{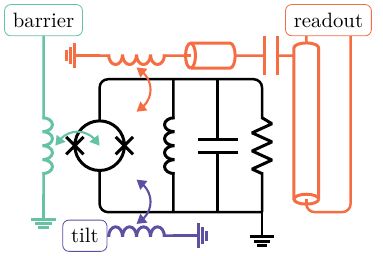} 
    \caption{Circuit diagram of the \tnode{} with a nonlinear tunable double-well potential made with a dc SQUID, an inductor, and a capacitor in parallel. The dissipation and noise of the system is modeled with a shunt resistance. The inductively coupled barrier and tilt lines allow control over the Josephson energy and linear bias. Mutual inductances are indicated by an arc between an inductor and the relevant loop. The system is measured by probing the transmission line that is coupled to a readout resonator whose resonance is dependent on the state of the \tnode{} through the dispersive shift.}
    \label{fig:circuit-diagram}
\end{figure}

The \tnode{} comprises a CPW section similar to that of a $\lambda/4$ resonator but shunted to ground at the open end by a dc-SQUID loop formed by a pair of Josephson junctions. The distributed inductance and capacitance of the CPW line together with the dc-SQUID give rise to a system that can be modeled with an equivalent lumped capacitance $C$, inductance $L$, and Josephson junction critical current $I_c$, connected in parallel, as shown in the circuit diagram of Figure \ref{fig:circuit-diagram}. The classical Hamiltonian of the \tnode{} circuit is given by
\begin{align}
    H = \frac{1}{2C}q^2 + \underbrace{\dfrac{1}{2L} (\phi-\phi_{\text{tilt}})^2 - E_J({\phi}_{\text{bar}}) \cos\left(2\pi \frac{\phi}{\Phi_0}\right)}_{U(\phi)},
    \label{eq:hamiltonian}
\end{align}
where the charge $q$ and flux $\phi$ are conjugate variables. Physically, the flux $\phi$ corresponds to the voltage across the open end of the $\lambda/4$ section, which is therefore also the voltage across the dc-SQUID. The charge $q$ is accumulated between the open end and the ground plane. $\Phi_0 \equiv h/2e$ is the magnetic flux quantum, \mbox{$E_J({\phi}_{\text{bar}}) \equiv E_{J_\text{max}} \cos({\pi \phi_{\text{bar}}/\Phi_0})$} is the effective Josephson energy, with $E_{J_\text{max}} = \Phi_0 I_c/2\pi$, and ${\phi}_{\text{bar}}$ and ${\phi}_{\text{tilt}}$ are control fluxes. These control fluxes can be tuned via two independent flux lines and affect the cosine and quadratic terms, thereby determining the  shape of the potential $U(\phi)$. One line is coupled inductively to the dc-SQUID loop and controls the effective Josephson energy, via control flux ${\phi}_{\text{bar}}$. The cosine term can form a barrier within the quadratic potential thus creating a double-well system when $E_J$ is negative, depending on the values of $E_{J_\text{max}}$ and $L$. The second flux line is coupled inductively to the current mode of the CPW segment and shifts the position of the cosine relative to the parabola by $\phi_{\text{tilt}}$. In the double-well regime, this results in one well being tilted lower than the other. In practice, voltages applied to the control lines are not directly proportional to the control fluxes ${\phi}_{\text{bar}}$ and ${\phi}_{\text{tilt}}$ due to crosstalk. This is covered in \cref{sec:methods}. The Hamiltonian in \cref{eq:hamiltonian} represents a simplified model that neglects the junctions' asymmetry and intrinsic capacitances and inductances. For the analysis of our experimental data, we used a more complex model that provides a better fit to reality, see \cref{ap:hamiltonian-fit}. 


At low temperature, loss, and noise, the circuit is expected to behave according to quantum mechanics. In fact, the \tnode{} circuit is identical to the fluxmon of Ref.~\cite{quintana2017superconducting} where it is used as a flux qubit for quantum annealing applications. In the quantum regime, the classical Hamiltonian of \cref{eq:hamiltonian} may be quantized by converting variables $q$ and $\phi$ to quantum operators $\hat{q}$ and $\hat{\phi}$, according to the commutation relation between $\hat{q}$ and $\hat{\phi}$. Thus, there exist two regimes to describe the \tnode{}: the quantum and thermal regimes. As noise increases as a function of the temperature of the system, the transition between the two regimes is generally characterized by a temperature $T_{\text{cross}}$. When the \tnode{} potential is biased to a double-well shape, this temperature denotes the point at which crossings from one well to the other stop being dominated by macroscopic resonant quantum tunneling (MRT), and instead become caused by thermal activation. It is expressed by~\cite{grabert1984crossover, hanggi1985quantum, devoret1985measurements, li2002quantitative, massarotti2012escape, affleck1981quantum},

\begin{equation}
    T_{\text{cross}} = \frac{\hbar \omega_{\text{b}}}{2\pi k_\text{B}},
    \label{eq:Tcross}
\end{equation}
where $C\omega_{\text{b}}^2 = -U''(0)$ is the curvature of the potential at the barrier peak $\phi_\text{b}$ and we assume a symmetric well, meaning $\phi_{\text{tilt}} = 0$.

In the thermal regime, we can describe the \tnode{} via its Langevin equation of motion [Eq.~\eqref{eq:underdamped_langevin}]. In order to account for thermal fluctuations, we model loss and noise by adding a resistor $R$ in parallel to the rest of the circuit~\cite{martinis1987experimental}, see Figure~\ref{fig:circuit-diagram}. This parallel resistance represents a normal current channel (non-superconducting), and is therefore dissipative. Since a high resistance will lead to a smaller current through the resistor, a high $R$ leads to lower loss. The noise caused by this resistance is Johnson-Nyquist noise, represented with independent Wiener processes $dW_t^{(i)}$~\cite{buttiker1983thermal, han1992effect, pratt2025extracting}.

Although the circuit is made out of superconducting metal, at non-dc frequencies the resistance is still nonzero. In addition, quasiparticles, interactions with lossy dielectrics and coupling to the environment all contribute to losses. In particular, as temperature rises, the increasing quasiparticle density leads to higher losses~\cite{anferov2024improved}. Thus it is important to keep in mind that $R$ is temperature-dependent, which affects both the resistance (loss) and noise.

Taking the flux through the inductor to be the degree of freedom of this system, the Langevin equation of motion (with $p\equiv q$) is
\begin{align}
\begin{split}
d\phi &= \frac{p}{C}dt \\
dp &= -\biggl(\frac{\partial U(\phi)}{\partial \phi} + \frac{1}{RC} \, p\biggr) \; dt + \sqrt{\frac{2}{R\beta}}\;dW_t.
\end{split}
\label{eq:EoM}
\end{align}
In analogy to a mechanical system, we can view the system as a flux ``particle'' of mass $C$ with position $\phi$ and momentum $q$ moving in a potential landscape $U(\phi)$ that is parametrized by ${\phi}_{\text{bar}}$ and $ {\phi}_{\text{tilt}}$. The inverse resistance $1/R$ corresponds to friction slowing down the particle.

As mentioned above, the potential $U(\phi)$ has two main configurations: single- or double-well. The single well case occurs when either the cosine and quadratic terms have their minima aligned, for which the well will look somewhat quartic, or when the cosine term is much smaller than the quadratic term, in which case the well is near harmonic. For the double-well, the cosine needs to be flipped such that the maximum is aligned with the quadratic potential minimum. If, additionally, $E_J > E_L$, this will create a barrier between two wells bounded on the left and right by the harmonic potential. Just like a simple SQUID or Josephson junction, the \tnode{} can be described by a plasma frequency $\omega_\text{p}$. This is the frequency of small oscillations at the bottom of a well, whether in the single or double-well configuration. The plasma frequency approximately corresponds to the frequency between the two lowest lying quantum eigenstates, and is the frequency we refer to as the \tnode{} frequency throughout the text

In this work, we are interested in the thermally activated dynamics, in which noise acts as the only driving force on the system. In particular, we want to measure the rate $\Gamma$ at which a flux particle in one of the wells ``escapes'' to the other well by crossing over the barrier (classically or thermally). Given a \tnode{} with population initially starting all in one well, we expect that the population will decrease exponentially in time to some equilibrium value. The defining timescale for this behavior is the thermalization time $\tau_\text{therm}$, which is simply the inverse of the escape rate out of one well. The escape rate for the double-well system can be described by an Arrhenius-type law~\cite{kramers1940brownian, buttiker1983thermal, han1989thermal, han1992effect}
\begin{equation}
    \Gamma = a_t \frac{\omega_\text{p}}{2\pi} \exp \bigg(- \frac{\Delta U}{E_{\text{esc}}} \bigg),
    \label{eq:arrhenius}
\end{equation}
where $\omega_\text{p}$ is the plasma frequency, $a_t$ is a factor dependent on the damping coefficient $\eta = 1/RC$, $\Delta U$ is the height of the potential barrier, $T$ is the sample temperature, and $E_{\text{esc}}$ is the escape energy. For thermally activated dynamics $E_{\text{esc}}=k_\text{B}T$. There are three relevant damping regimes for $a_t$, namely (1) heavy damping, (2) small damping and (3) extreme underdamping~\cite{buttiker1983thermal}. For these regimes, we have $a_t=|\omega_\text{b}|/\eta$, $a_t = 1$, and $a_t \propto {\eta \sqrt{C\Delta U}}/{k_\text{B}T}$, respectively. Deriving \cref{eq:arrhenius} generally involves assuming $k_\text{B}T \ll \Delta U$, but the expression has been shown to be effective even when $k_\text{B}T \sim \Delta U$~\cite{han1992effect}. Note that below the crossover temperature, barrier crossings are caused by MRT, and we therefore expect that $E_{\text{esc}}$ will not vary as a function of temperature. 

As the system dynamics rely on superconductivity, we are restricted to work below the critical temperature of aluminum ($T_{\text{crit}} = \qty{1.2}{\kelvin}$). Furthermore, it should be noted that the population of quasiparticles in aluminum increases significantly beyond \qty{160}{\milli\kelvin}~\cite{anferov2024improved}, increasing losses by lowering the parallel resistance $R$ and affecting the \tnode{} dynamics.

\subsection{Experimental methods}
\label{sec:methods}

The main experiment presented in this article is the measurement of the escape energy $E_{\text{esc}}$ of the \tnode{} as a function of temperature. For this purpose, we need to
\begin{enumerate}
    \item determine the various circuit parameters, i.e., $C$, $L$, and $I_c$,
    \item be able to initialize the system in a known state and measure whether the flux particle is in the left or right well, and
    \item control the potential shape and temperature in order to measure relaxation curves, that is, to measure the population of the left and right wells as a function of time and barrier height.
\end{enumerate}

In order to properly characterize the system, and in particular, to fit the parameters in the Hamiltonian of \cref{eq:hamiltonian}, we must know the relation between the voltages that we apply from our room temperature electronics and the effective flux being coupled into the SQUID barrier loop $\phi_\text{bar}$ and the main tilt loop $\phi_\text{tilt}$. The approach we take to this calibration is detailed in Appendix~\ref{ap:crosstalk}.

\begin{figure*}[ht]
    \centering
    \includegraphics[width=0.7\linewidth]{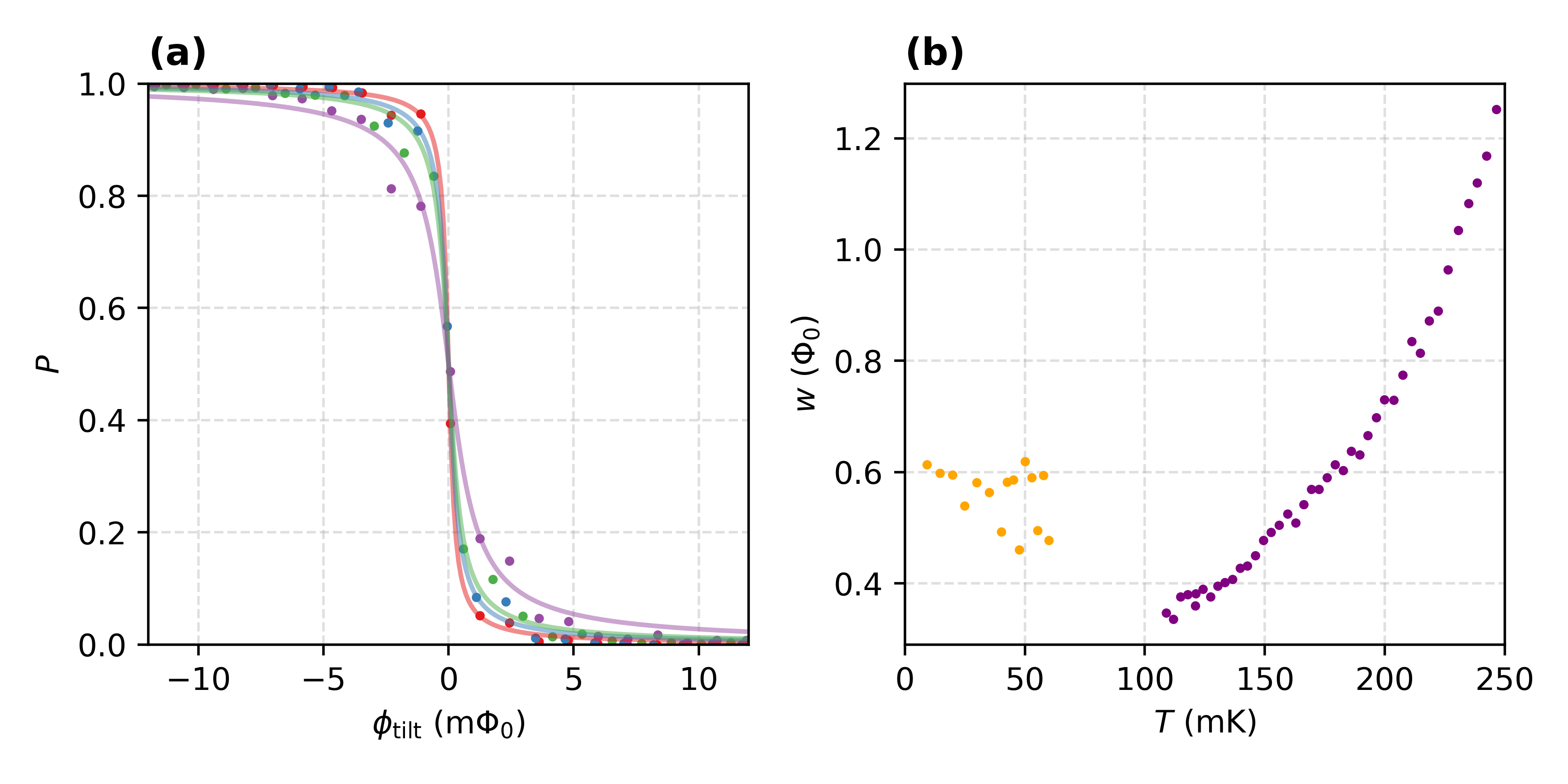}
    \caption{Sigmoidal relationship between the expectation value of position and a linear force for the double-well per \cref{eq:empirical-sigmoid}. (a) Example tilt calibration.
    (b) Temperature-dependence of the width parameter $w$. The \textcolor{readoutColor}{orange} markers designate data taken below \qty{60}{\milli\kelvin} and the \textcolor{tiltColor}{violet} markers designate data taken above \qty{100}{\milli\kelvin}.}
    \label{fig:sigmoid}
\end{figure*}

\subsubsection{Initialization and readout}

Once the flux axes have been calibrated, we can obtain a rough idea of what parameters to use in order to initialize and read out the state. The procedure consists in first lowering the barrier with a $\phi_\text{bar}$ pulse such that the \tnode{} population will equilibrate to a single well. We then tilt the well with a $\phi_\text{tilt}$ pulse to the left or right before raising the barrier with a final $\phi_\text{bar}$ pulse. Assuming that equilibration time is sufficient and that the barrier is not raised ``too fast'', this sequence deterministically confines the \tnode{} population within the left or right well.

Because we observe an anti-crossing with the resonator, we know that for control fluxes within the relaxation operating regime (see the triangle in Figure~\ref{fig:exp-diagram}) the \tnode{} has a frequency lower than that of the resonator. From simulations, we can estimate the rough location where we expect the \tnode{} to be in the single-well regime, and therefore where we may initialize the population. Near the middle of the triangle is a good point to start. Once a basic readout has been achieved, we can sweep over the various parameters involved to optimize it.

Readout is performed dispersively~\cite{quintana2017superconducting}, allowing for fast discrete measurements of the flux degree of freedom of the \tnode{} in the left ($L$) or right ($R$) well. It is achieved by inductively coupling the \tnode{} to a CPW readout resonator that in turn is capacitively coupled to a \qty{50}{\ohm} transmission line. Occupancy of either well induces a dispersive shift to the readout resonator frequency. These two dispersive shifts depend on the control biases and differ when an asymmetry is introduced through the tilt control; for more details see Appendix~\ref{ap:relax}. This tilt asymmetry is crucial, since at zero tilt, the two well states have exactly the same frequency, thereby causing an identical frequency shift to the resonator. By probing the transmission line at an appropriate frequency, the flux can be inferred from the state-dependent response when this frequency difference is resolvable.

Note that the measurement procedure interrupts the flux trajectory, and is therefore ``destructive,'' as it involves adjusting the \tnode{} potential to an asymmetric double-well with a sufficient dispersive shift difference. More details on readout can be found in \cref{ap:readout}. The result of each individual measurement is a binary variable, 1 or 0, corresponding to the left or right well. To obtain population statistics, we repeat the experimental sequence 1000 times and average the result, which we denote $P$.

Once the correct flux amplitudes, pulse timings, and measurement frequencies have been determined, we may characterize the initialization and readout performance with a measurement of the ``s-curve''. For this experiment, we perform normal state initialization and readout, but we sweep over the initial tilt pulse amplitude. For a double-well system, the expected value of the flux particle population with respect to $\phi_\text{tilt}$ is sigmoidal. At large initialization tilts, we expect to measure the full population $P$ in the left ($P=1$) or right ($P=0$) well. Near zero tilt, the population should be approximately $P=0.5$. We use the following empirical relationship to characterize this
\begin{equation}
    P(\phi_\text{tilt}) = \frac{1}{2} - \frac{1}{\pi}\arctan \Big(\frac{\pi}{2}\frac{\phi_\text{tilt}}{w}\Big)
    \label{eq:empirical-sigmoid}
\end{equation}
where $w$ is the s-curve sigmoid width parameter. This width largely depends on two effects: how quickly the barrier is raised and the thermalization temperature. In principle, raising the barrier slowly with even a small amount of tilt would confine the full population to the corresponding well, leading to a narrow width. However, for very slow barrier ramps, thermal noise will lead the population to continuously re-equilibrate, leading to a larger width. Figure~\ref{fig:sigmoid} presents the result of the s-curve characterization. In Figure~\ref{fig:sigmoid}~(a), we show a few selected s-curves measured at various temperatures, but with constant barrier ramp time. For all temperature points, the state initialization and readout is perfect for tilts $\phi_\text{tilt} > 10 \text{m}\Phi_0$. In Figure~\ref{fig:sigmoid}~(b), we directly plot the fitted value of the width, which increases with temperature.

In our experiments, the $\phi_\text{bar}$ and $\phi_\text{tilt}$ pulses used to initialize the population last between 3 and \qty{5}{\us}, and we use a linear ramp of \qty{26.67}{\ns} to raise the barrier for readout.

\begin{figure*}[hbt]
    \centering
    \includegraphics[width=\linewidth]{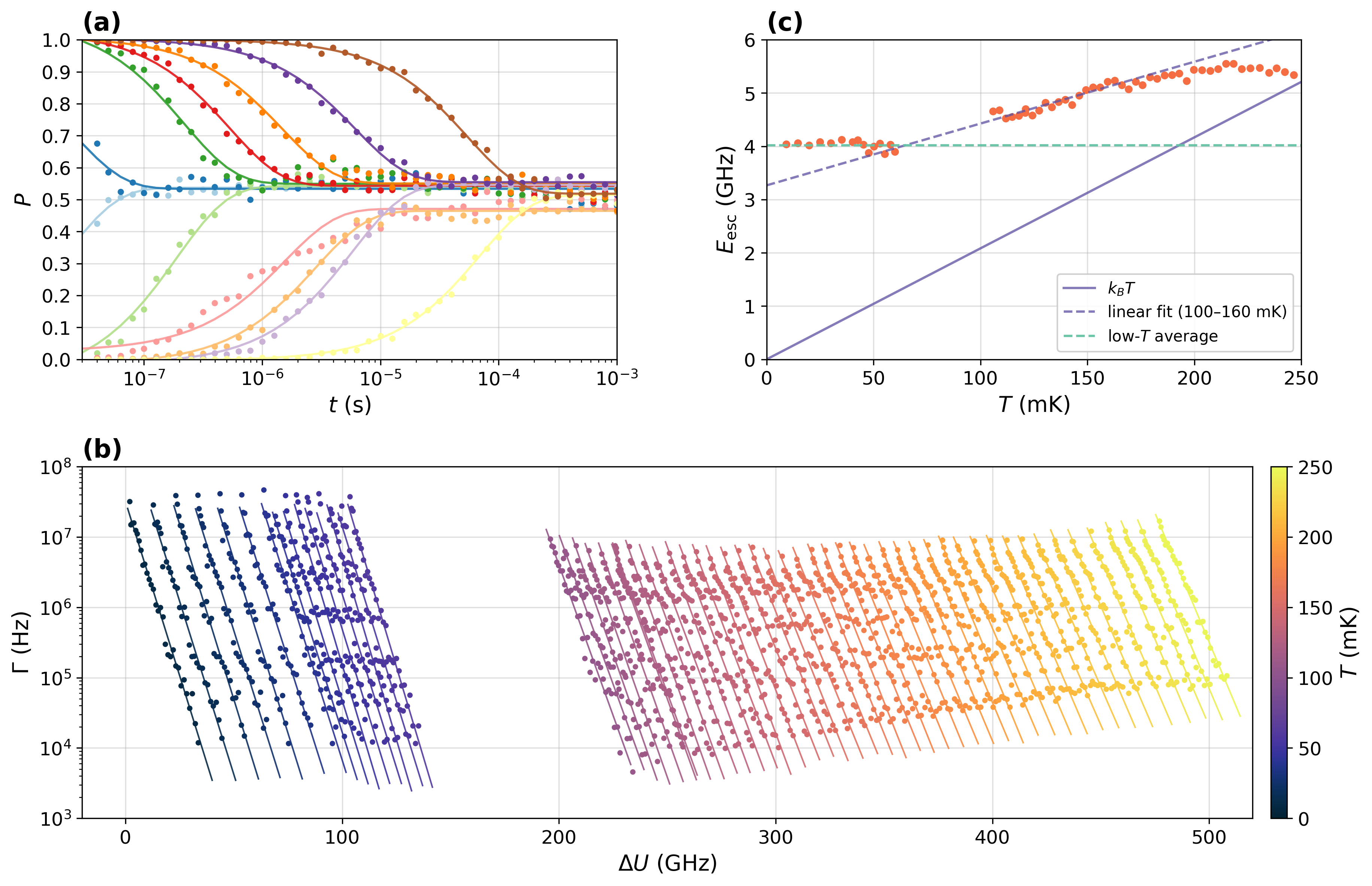} 
    \caption{Results of the relaxation experiments. (a) Trajectories of the left well population $P$ versus time for a selection of barrier heights and temperatures. The lines are fits to the exponential model described in \cref{sec:relaxfit}; as expected, the data are well described by an exponential. (b) Relaxation rate $\Gamma$ versus barrier height $\Delta U$ for all measured temperatures. The straight lines are linear fits in logarithmic scale and are plotted for $\Delta U$ from 1 to \qty{40}{\GHz}. For clarity, each curve is offset horizontally according to the temperature difference between them, with a spacing of \qty{2}{\GHz\per\milli\kelvin} relative to the first curve at \qty{9}{\milli\kelvin}. (c) Escape energies $E_{\text{esc}}$ versus temperature. The energies are extracted from the inverse negative slope of a linear fit to $\ln(\Gamma/\omega_\mathrm{p})$ based on \cref{eq:arrhenius}. The data is shown in orange. The violet solid line indicates $E_{\text{esc}}=k_\text{B}T$, and the associated dashed line is a fit to the data between 100 and \qty{160}{\milli\kelvin}, ignoring higher temperatures because of potential quasiparticle effects. The mint dashed line shows the average $E_\text{esc}$ in the low-temperature regime where the data is flat.}
    \label{fig:relaxation}
\end{figure*}

\subsubsection{Calculation of escape energy}
\label{sec:relaxfit}
Since each of the relaxation curves that we measure is an exponential decay, it is possible to fit them to a simple exponential model with three parameters:
\begin{align}
    P(t) = A\exp(-t/\tau) + B
\end{align}
where $\tau = 1/\Gamma$ is the relaxation time constant, and $A$ and $B$ are used as fit parameters and are near 0.5.

In order to calculate the effective escape energy $E_\text{esc}$, we must fit the relaxation rates obtained in the experiment according to the model of \cref{eq:arrhenius}. We rely on our knowledge of the system to estimate the plasma frequency $\omega_\text{p}$ and the barrier height $\Delta U$ as a function of $\phi_\text{bar}$. We may compute the natural logarithm of $\Gamma/\tfrac{\omega_\text{p}}{2\pi}$ and fit it linearly against $\Delta U$. Then, the inverse negative slope of this line is the activation energy, and the intercept is the natural logarithm of $a_t$:
\begin{equation}
    \ln\left(\Gamma/\frac{\omega_\text{p}}{2\pi}\right) = \ln(a_t)  - \frac{\Delta U}{E_{\text{esc}}}.
\end{equation}

\subsection{Experimental results}

Figure~\ref{fig:relaxation} presents the main results of this section. In Figure~\ref{fig:relaxation}~(a), we plot population $P$ data from the relaxation experiment for a few select barrier settings and temperatures. With a perfect initialization in the left or right well, at $t=0$ the \tnode{} is always found in the well it was prepared in. Then depending on the barrier height and thermal energy, the average population decays exponentially until it stabilizes at approximately $P=0.5$. We fit each relaxation curve and extract the time constant $\tau$. Over all measured barriers and temperatures, we notice that the fit quality as reported via the residuals sum of squares (RSS) is better (i.e., lower RSS) for larger barriers and lower temperatures.

We obtain $\tau$ for both left and right initializations. We then average them and inverse the result to obtain the relaxation rate $\Gamma$, which we plot in Figure~\ref{fig:relaxation}~(b) as a function of estimated barrier height $\Delta U$ for each temperature. When plotted with a logarithmic scale and for constant temperature, the rates trace a straight line, decreasing exponentially as a function of $\Delta U$. This allows us to fit each $\Gamma$ curve according to the exponential model of \cref{eq:arrhenius}, as described in \cref{sec:relaxfit}. Then, the negative inverse slope gives the escape energy $E_\text{esc}$, which, when well within the thermal regime, should be equal to $k_\text{B} T$.

We plot the final result, $E_\text{esc}$ as a function of $T$, in Figure~\ref{fig:relaxation}. Due to the control issue potentially caused by contamination (see \cref{ap:contamination}), we do not have data points between 60 and \qty{100}{\milli\kelvin}. At low temperature, the escape energy appears to be constant. Above \qty{100}{\milli\kelvin}, $E_\text{esc}$ increases linearly until approximately 165 to \qty{200}{\milli\kelvin}, at which point the escape energy stops increasing and instead appears to taper off. 

The first feature we discuss is the escape energy at low temperature. This energy corresponds to the quantum tunneling through the barrier, and is therefore not temperature-dependent; this is what we observe. However, the value of the energy in the quantum regime should correspond to $k_\text{B}T_\text{cross}$, and can therefore be estimated from our knowledge of the potential $U(\phi)$. Given the circuit parameters determined in \cref{ap:hamiltonian-fit}, the frequency corresponding to the inverse curvature at the barrier top $\omega_b$ varies between 6 and \qty{10}{\GHz} for the range of $\phi_\text{bar}$ used. According to \cref{eq:Tcross}, this leads to $T_\text{cross}$ between 46 and \qty{76}{\milli\kelvin}. If we average the points below \qty{60}{\milli\kelvin}, we obtain an escape energy of $E_\text{esc}/h = \qty{4.02}{\GHz}$, which corresponds to a temperature of \qty{193}{\milli\kelvin}. Thus, the escape energy we measure is larger than expected by a factor of almost 3. Here, we emphasize that both the ratio between the measured $E_\text{esc}$ and $\omega_b$ depend quite strongly on the Hamiltonian parameters of the \tnode{}, that is, $C$, $L$, and $I_c$. Given the very large uncertainty on these parameters, it is not unlikely that wrong values could be causing this difference.

Above the crossover temperature, the escape energy should start increasing and eventually should be equal to $k_B T$. While we cannot observe the section where this transition occurs, due to the control issue of \cref{ap:contamination}, we can see that the escape energy does increase linearly at a rate of approximately \qty{12}{\GHz\per\kelvin}. The rate of increase, however, is smaller than the expected $k_B/h = \qty{20.8}{\GHz\per\kelvin}$. Additionally, $E_\text{esc}$ gradually stops increasing above \qty{165}{\milli\kelvin}. There are multiple potential reasons for this. One obvious candidate is the quasiparticle population in the superconducting metal. Indeed, for aluminum, the equilibrium quasiparticle population starts to increase above \qty{160}{\milli\kelvin}, causing a large increase in loss. This effect has been measured in superconducting qubits and resonators. The model for the escape energy increasing as $k_B T$ is for constant loss (i.e., friction, $1/R$). Here, as $T$ increases, $1/R$ also increases. In addition, as mentioned in \cref{sec:design}, there are multiple functional forms for $a_t$ in \cref{eq:arrhenius} that depend on damping. It is possible that the increased loss changes the value of $a_t$ as temperature increases, something we have not accounted for.

A related possibility is that our ``crossover'' region may be particularly large, and that therefore, we never fully enter the thermal regime before quasiparticle loss takes over. Indeed, because we measure relaxation curves that include very small barriers, the range of $\omega_b$ and $\omega_p$ covered by each $\phi_\text{bar}$ sweep is quite large, increasing by a factor 2 to 3 over the sweep. Thus, there is not a single $\omega_b$ or $\omega_p$ that conclusively characterize the transition, but a range of values, thereby leading to a range of $T_\text{cross}$.

\section{Discussion}

There are a number of converging lines of argument that independently motivate the idea of thermodynamic computing. The first line of argument follows from the way in which scale-dependent effects will impede current hardware miniaturization efforts. What's known as Moore’s law is an informal law, encoding the empirical observation that compute density has tended to roughly double every two years, as measured by the number of transistors on a microchip. On current estimates, by the end of this decade, circuits will become so small and dense that scale-dependent quantum effects will begin to disrupt the determinism of binary logic upon which current forms of digital computing depend~\cite{shalf2020future}. Along the current path, development of digital computing hardware will therefore run into some physical hard limits. We argue that thermodynamic computing provides an alternative to the current paradigm that can avoid this dead end. Additionally, thermodynamic computing promises both a more efficient path to the development of computing technologies (and of artificial intelligence in particular). Recent growth in the demand for compute resources has led to an associated growth in large-scale projects aiming to vastly increase energy production. Given the orders-of-magnitude efficiency increases that it enables, we believe that thermodynamic computing opens paths to scaling up artificial intelligence without imposing unreasonable demands on modern power grids, thus promising to reduce environmental and financial costs.

\subsection{Superconducting outlook}

Multiple technical hurdles have to be overcome to achieve a large-scale thermodynamic computer based on superconducting circuits. 
Although devices with thousands of similar nodes have been demonstrated~\cite{king2023quantum}, scalability remains a key challenge, in particular with regard to control and calibration. Primarily driven by the field of superconducting quantum computing, efforts are underway to make the platform more scalable through CMOS-compatible fabrication~\cite{vandamme2024advanced}, 3D integration~\cite{rosenberg20173d, yost2020solid, mallek2021fabrication, vahidpour2017superconducting} and multiplexed control~\cite{acharya2023multiplexed}. Additionally, there are pathways to miniaturize the circuit components. For example, the capacitive elements could be miniaturized by introducing higher-permittivity dielectric materials and inductors can be sized down by utilizing high kinetic inductance materials~\cite{gupta2024low}. A thermodynamic computer, contrary to a superconducting quantum computer, does not rely on highly quantum coherent operations and is therefore less prone to dissipative dielectrics.

Another challenge lies in the readout method, as the presented method interrupts the \tnode{} trajectory and is limited to solely measuring the discrete left or right position. Future work could improve readout by implementing continuous measurements, enabling broader applications like monitoring the state continuously to estimate work statistics~\cite{wimsatt2021harnessing}. More subtle approaches have already been presented in other works, including a magnetometer based on a resonator terminated by an rf-SQUID~\cite{novikov2018exploring}. Another digital approach to measuring  stationary fields confined in cavities has been proposed~\cite{strandberg2024digital} but this technique is experimentally challenging and less scalable.

For any actual thermodynamic calculation, multiple \tnode{s} will need to be coupled together in order to form a large potential landscape with many degrees of freedom. Such coupling can be made with tunable inductive couplers, as shown in D-Wave devices or flux qubits~\cite{quintana2017superconducting}. Additionally, such couplers could be engineered to generate a 3-body interaction.

While noise and thermal fluctuations can be controlled via the temperature of the device, it could be useful to also directly control the effective friction or resistance of the system. There are a few potential avenues for doing so. One is simply to control resistance directly via the temperature, which changes the quasiparticle population. This may not be the most useful technique since it forces the resistance to change in lockstep with the noise, but does not require any experimental additions. Another way would be to inject quasiparticles from room temperature, either with a strong microwave pulse or an infrared source. Finally, a more complex method would be to directly fabricate a normal metal resistance on chip. These methods could also be used together for a more complete control over the friction of individual \tnode{s}.

While the device's on-chip power is low, the total energy consumption is considerable and near-exclusively determined by cooling infrastructure. However, there are pathways to reduce the required cooling power. These include moving to a new material platform with a higher critical temperature such as niobium ($T_{\text{crit}}=\qty{9.3}{\kelvin}$). Moreover, the thermal load can be reduced by integrating the control electronics into the cryogenic setup, reducing the thermal load.

Another challenge is that estimation oscillators are likely a difficult operation to perform on superconducting hardware (due to the magnitude of change in capacitance required), but there are some proposals that one could consider developing further~\cite{paik2005cooper}.

\subsection{Other applications and future directions}

As outlined in Sec.~\ref{sec:examples}, there are a number of other potential applications that can be explored. Each application or hardware modality may have a specific set of requirements or require adaptations, but the framework that we have proposed is general and flexible enough to integrate many workflows. For example, recent work has shown the power of quartic potentials similar to those presented here, which undergo (overdamped) Langevin dynamics and which can even be trained directly as diffusion models~\cite{whitelam2025generative}. 

We have demonstrated that our energy-based thermodynamic computing framework can successfully implement a broad range of different probabilistic machine learning methods. Thermodynamic computers need not be used in isolation and could substantially improve the efficiency of both deterministic and quantum computing stacks. One potential first such application of superconducting thermodynamic computers is as co-processors with superconducting quantum computing chips, for example, to help with error correction (which digital graphical models are already used for~\cite{liu2019neural, old2023generalized}). 

Having already outlined further steps to advance superconducting hardware, we note in closing that the ideas of this framework could be implemented in a variety of modalities and are not intrinsically tied to superconductors. We have explored similar directions in recent CMOS work~\cite{jelinvcivc2025efficient, Freitas2026NEATRN}.

\section{Conclusion}

In this paper, we presented a framework for energy-based thermodynamic computing based on probabilistic graphical models and superconducting circuits driven by thermal fluctuations. We have described the relevant building blocks and carried out idealized analyses of equilibration times for relevant operations of the device. These results suggest a variety of avenues for further explorations. Our work contributes to the exciting development of physics-based computing hardware at scale for meaningful machine learning workflows, which dovetails with neuromorphic approaches in neuroscience and machine learning. We hope that this work helps to inspire the research and development of large-scale energy-based thermodynamic computers.
\vspace{1em}

\section{Acknowledgments}


The authors thank Jason Shi for their contributions to thermoformer numerics, Ian MacCormack for their contributions to the Fokker-Planck analytics, Geremia Massarelli and Jeremy Rothschild for their work on timescale separation, 
Maxwell Ramstead for their writing help, 
and Thomas Hubregtsen for early numerics infrastructure. 

Experiments were performed at Espace~Quantique~1 at the DistriQ~Quantum~Innovation~Zone in Sherbrooke, Canada. Fabrication was performed at the Interdisciplinary Institute for Technological Innovation (3IT) and l\textquotesingle Infrastructure matériaux et dispositifs quantiques (IMDQ) facilities of l\textquotesingle Universit\'e~de~Sherbrooke. We thank the cleanroom staff as well as the fabrication team of Nord~Quantique for their kind assistance with developing the fabrication process.

\noindent

\bibliography{newreferences}

@article{masanet2020recalibrating,
  title={Recalibrating global data center energy-use estimates},
  author={Masanet, Eric and Shehabi, Arman and Lei, Nuoa and Smith, Sarah and Koomey, Jonathan},
  journal={Science},
  volume={367},
  number={6481},
  pages={984--986},
  year={2020},
  publisher={American Association for the Advancement of Science}
}

@inproceedings{reuther2022ai,
  title={AI and ML accelerator survey and trends},
  author={Reuther, Albert and Michaleas, Peter and Jones, Michael and Gadepally, Vijay and Samsi, Siddharth and Kepner, Jeremy},
  booktitle={2022 IEEE High Performance Extreme Computing Conference (HPEC)},
  pages={1--10},
  year={2022},
  organization={IEEE}
}

@article{de2023growing,
  title={The growing energy footprint of artificial intelligence},
  author={de Vries, Alex},
  journal={Joule},
  volume={7},
  number={10},
  pages={2191--2194},
  year={2023},
  publisher={Elsevier}
}

@inproceedings{luccioni2024power,
  title={Power hungry processing: Watts driving the cost of AI deployment?},
  author={Luccioni, Sasha and Jernite, Yacine and Strubell, Emma},
  booktitle={The 2024 ACM Conference on Fairness, Accountability, and Transparency},
  pages={85--99},
  year={2024}
}

@article{wolpert2024stochastic,
  title={Is stochastic thermodynamics the key to understanding the energy costs of computation?},
  author={Wolpert, David H and Korbel, Jan and Lynn, Christopher W and Tasnim, Farita and Grochow, Joshua A and Karde{\c{s}}, G{\"u}lce and Aimone, James B and Balasubramanian, Vijay and De Giuli, Eric and Doty, David and others},
  journal={Proceedings of the National Academy of Sciences},
  volume={121},
  number={45},
  pages={e2321112121},
  year={2024},
  publisher={National Academy of Sciences}
}

@article{wolpert2019stochastic,
  title={The stochastic thermodynamics of computation},
  author={Wolpert, David H},
  journal={Journal of Physics A: Mathematical and Theoretical},
  volume={52},
  number={19},
  pages={193001},
  year={2019},
  publisher={IOP Publishing}
}

@article{niazi2024training,
  title={Training deep Boltzmann networks with sparse Ising machines},
  author={Niazi, Shaila and Chowdhury, Shuvro and Aadit, Navid Anjum and Mohseni, Masoud and Qin, Yao and Camsari, Kerem Y},
  journal={Nature Electronics},
  pages={1--10},
  year={2024},
  publisher={Nature Publishing Group UK London}
}

@article{camsari2017stochastic,
  title={Stochastic p-bits for invertible logic},
  author={Camsari, Kerem Yunus and Faria, Rafatul and Sutton, Brian M and Datta, Supriyo},
  journal={Physical Review X},
  volume={7},
  number={3},
  pages={031014},
  year={2017},
  publisher={APS}
}

@article{mohseni2024build,
  title={How to Build a Quantum Supercomputer: Scaling Challenges and Opportunities},
  author={Mohseni, Masoud and Scherer, Artur and Johnson, K Grace and Wertheim, Oded and Otten, Matthew and Aadit, Navid Anjum and Bresniker, Kirk M and Camsari, Kerem Y and Chapman, Barbara and Chatterjee, Soumitra and others},
  journal={arXiv preprint arXiv:2411.10406},
  year={2024}
}

@article{conte2019thermodynamic,
  title={Thermodynamic computing},
  author={Conte, Tom and DeBenedictis, Erik and Ganesh, Natesh and Hylton, Todd and Strachan, John Paul and Williams, R Stanley and Alemi, Alexander and Altenberg, Lee and Crooks, Gavin and Crutchfield, James and others},
  journal={arXiv preprint arXiv:1911.01968},
  year={2019}
}

@article{huembeli2022physics,
  title={The physics of energy-based models},
  author={Huembeli, Patrick and Arrazola, Juan Miguel and Killoran, Nathan and Mohseni, Masoud and Wittek, Peter},
  journal={Quantum Machine Intelligence},
  volume={4},
  number={1},
  pages={1},
  year={2022},
  publisher={Springer}
}

@article{lecun2006tutorial,
  title={A tutorial on energy-based learning},
  author={LeCun, Yann and Chopra, Sumit and Hadsell, Raia and Ranzato, M and Huang, Fujie and others},
  journal={Predicting structured data},
  volume={1},
  number={0},
  year={2006}
}

@article{du2019implicit,
  title={Implicit generation and modeling with energy based models},
  author={Du, Yilun and Mordatch, Igor},
  journal={Advances in Neural Information Processing Systems},
  volume={32},
  year={2019}
}

@article{song2021train,
  title={How to train your energy-based models},
  author={Song, Yang and Kingma, Diederik P},
  journal={arXiv preprint arXiv:2101.03288},
  year={2021}
}

@article{aadit2022massively,
  title={Massively parallel probabilistic computing with sparse Ising machines},
  author={Aadit, Navid Anjum and Grimaldi, Andrea and Carpentieri, Mario and Theogarajan, Luke and Martinis, John M and Finocchio, Giovanni and Camsari, Kerem Y},
  journal={Nature Electronics},
  volume={5},
  number={7},
  pages={460--468},
  year={2022},
  publisher={Nature Publishing Group}
}

@article{singh2024cmos,
  title={CMOS plus stochastic nanomagnets enabling heterogeneous computers for probabilistic inference and learning},
  author={Singh, Nihal Sanjay and Kobayashi, Keito and Cao, Qixuan and Selcuk, Kemal and Hu, Tianrui and Niazi, Shaila and Aadit, Navid Anjum and Kanai, Shun and Ohno, Hideo and Fukami, Shunsuke and others},
  journal={Nature Communications},
  volume={15},
  number={1},
  pages={2685},
  year={2024},
  publisher={Nature Publishing Group UK London}
}

@article{mohseni2022ising,
  title={Ising machines as hardware solvers of combinatorial optimization problems},
  author={Mohseni, Naeimeh and McMahon, Peter L and Byrnes, Tim},
  journal={Nature Reviews Physics},
  volume={4},
  number={6},
  pages={363--379},
  year={2022},
  publisher={Nature Publishing Group UK London}
}

@article{inagaki2016coherent,
  title={A coherent Ising machine for 2000-node optimization problems},
  author={Inagaki, Takahiro and Haribara, Yoshitaka and Igarashi, Koji and Sonobe, Tomohiro and Tamate, Shuhei and Honjo, Toshimori and Marandi, Alireza and McMahon, Peter L and Umeki, Takeshi and Enbutsu, Koji and others},
  journal={Science},
  volume={354},
  number={6312},
  pages={603--606},
  year={2016},
  publisher={American Association for the Advancement of Science}
}

@article{moy20221,
  title={A 1,968-node coupled ring oscillator circuit for combinatorial optimization problem solving},
  author={Moy, William and Ahmed, Ibrahim and Chiu, Po-wei and Moy, John and Sapatnekar, Sachin S and Kim, Chris H},
  journal={Nature Electronics},
  volume={5},
  number={5},
  pages={310--317},
  year={2022},
  publisher={Nature Publishing Group UK London}
}

@article{laydevant2024training,
  title={Training an ising machine with equilibrium propagation},
  author={Laydevant, J{\'e}r{\'e}mie and Markovi{\'c}, Danijela and Grollier, Julie},
  journal={Nature Communications},
  volume={15},
  number={1},
  pages={3671},
  year={2024},
  publisher={Nature Publishing Group UK London}
}

@article{chou2019analog,
  title={Analog coupled oscillator based weighted Ising machine},
  author={Chou, Jeffrey and Bramhavar, Suraj and Ghosh, Siddhartha and Herzog, William},
  journal={Scientific reports},
  volume={9},
  number={1},
  pages={14786},
  year={2019},
  publisher={Nature Publishing Group UK London}
}

@article{king2022coherent,
  title={Coherent quantum annealing in a programmable 2,000 qubit Ising chain},
  author={King, Andrew D and Suzuki, Sei and Raymond, Jack and Zucca, Alex and Lanting, Trevor and Altomare, Fabio and Berkley, Andrew J and Ejtemaee, Sara and Hoskinson, Emile and Huang, Shuiyuan and others},
  journal={Nature Physics},
  volume={18},
  number={11},
  pages={1324--1328},
  year={2022},
  publisher={Nature Publishing Group UK London}
}

@article{king2023quantum,
  title={Quantum critical dynamics in a 5,000-qubit programmable spin glass},
  author={King, Andrew D and Raymond, Jack and Lanting, Trevor and Harris, Richard and Zucca, Alex and Altomare, Fabio and Berkley, Andrew J and Boothby, Kelly and Ejtemaee, Sara and Enderud, Colin and others},
  journal={Nature},
  volume={617},
  number={7959},
  pages={61--66},
  year={2023},
  publisher={Nature Publishing Group UK London}
}

@article{johnson2011quantum,
  title={Quantum annealing with manufactured spins},
  author={Johnson, Mark W and Amin, Mohammad HS and Gildert, Suzanne and Lanting, Trevor and Hamze, Firas and Dickson, Neil and Harris, Richard and Berkley, Andrew J and Johansson, Jan and Bunyk, Paul and others},
  journal={Nature},
  volume={473},
  number={7346},
  pages={194--198},
  year={2011},
  publisher={Nature Publishing Group UK London}
}

@article{shainline2017superconducting,
  title={Superconducting optoelectronic circuits for neuromorphic computing},
  author={Shainline, Jeffrey M and Buckley, Sonia M and Mirin, Richard P and Nam, Sae Woo},
  journal={Physical Review Applied},
  volume={7},
  number={3},
  pages={034013},
  year={2017},
  publisher={APS}
}

@inproceedings{coles2023thermodynamic,
  title={Thermodynamic AI and the fluctuation frontier},
  author={Coles, Patrick J and Szczepanski, Collin and Melanson, Denis and Donatella, Kaelan and Martinez, Antonio J and Sbahi, Faris},
  booktitle={2023 IEEE International Conference on Rebooting Computing (ICRC)},
  pages={1--10},
  year={2023},
  organization={IEEE}
}

@article{aifer2024thermodynamic,
  title={Thermodynamic linear algebra},
  author={Aifer, Maxwell and Donatella, Kaelan and Gordon, Max Hunter and Duffield, Samuel and Ahle, Thomas and Simpson, Daniel and Crooks, Gavin and Coles, Patrick J},
  journal={npj Unconventional Computing},
  volume={1},
  number={1},
  pages={13},
  year={2024},
  publisher={Nature Publishing Group UK London}
}

@misc{langevin1908sur,
	title = {Sur la théorie du mouvement brownien},
	author = {Langevin, Paul},
	month = mar,
	year = {1908},
}

@phdthesis{crooks1999excursions,
	address = {Berkeley, California, United States of America},
	title = {Excursions in {Statistical} {Dynamics}},
	school = {University of California at Berkeley},
	author = {Crooks, Gavin E},
	year = {1999},
}

@article{saira2020nonequilibrium,
	title = {Nonequilibrium thermodynamics of erasure with superconducting flux logic},
	volume = {2},
	url = {https://link.aps.org/doi/10.1103/PhysRevResearch.2.013249},
	doi = {10.1103/PhysRevResearch.2.013249},
	number = {1},
	urldate = {2023-01-03},
	journal = {Physical Review Research},
	author = {Saira, Olli-Pentti and Matheny, Matthew H. and Katti, Raj and Fon, Warren and Wimsatt, Gregory and Crutchfield, James P. and Han, Siyuan and Roukes, Michael L.},
	month = mar,
	year = {2020},
	note = {Publisher: American Physical Society},
	pages = {013249},
}

@inproceedings{zhai2016deep,
  title={Deep structured energy based models for anomaly detection},
  author={Zhai, Shuangfei and Cheng, Yu and Lu, Weining and Zhang, Zhongfei},
  booktitle={International conference on machine learning},
  pages={1100--1109},
  year={2016},
  organization={PMLR}
}

@article{ranzato2006efficient,
  title={Efficient learning of sparse representations with an energy-based model},
  author={Ranzato, Marc'Aurelio and Poultney, Christopher and Chopra, Sumit and Cun, Yann},
  journal={Advances in neural information processing systems},
  volume={19},
  year={2006}
}

@article{osadchy2004synergistic,
  title={Synergistic face detection and pose estimation with energy-based models},
  author={Osadchy, Margarita and Miller, Matthew and Cun, Yann},
  journal={Advances in neural information processing systems},
  volume={17},
  year={2004}
}

@inproceedings{liu2022compositional,
  title={Compositional visual generation with composable diffusion models},
  author={Liu, Nan and Li, Shuang and Du, Yilun and Torralba, Antonio and Tenenbaum, Joshua B},
  booktitle={European Conference on Computer Vision},
  pages={423--439},
  year={2022},
  organization={Springer}
}

@inproceedings{carreira2005contrastive,
  title={On contrastive divergence learning},
  author={Carreira-Perpinan, Miguel A and Hinton, Geoffrey},
  booktitle={International workshop on artificial intelligence and statistics},
  pages={33--40},
  year={2005},
  organization={PMLR}
}

@article{hinton2002training,
  title={Training products of experts by minimizing contrastive divergence},
  author={Hinton, Geoffrey E},
  journal={Neural computation},
  volume={14},
  number={8},
  pages={1771--1800},
  year={2002},
  publisher={MIT Press}
}

@article{bengio2009justifying,
  title={Justifying and generalizing contrastive divergence},
  author={Bengio, Yoshua and Delalleau, Olivier},
  journal={Neural computation},
  volume={21},
  number={6},
  pages={1601--1621},
  year={2009},
  publisher={MIT Press}
}

@inproceedings{sutskever2010convergence,
  title={On the convergence properties of contrastive divergence},
  author={Sutskever, Ilya and Tieleman, Tijmen},
  booktitle={Proceedings of the thirteenth international conference on artificial intelligence and statistics},
  pages={789--795},
  year={2010},
  organization={JMLR Workshop and Conference Proceedings}
}

@article{song2020score,
  title={Score-based generative modeling through stochastic differential equations},
  author={Song, Yang and Sohl-Dickstein, Jascha and Kingma, Diederik P and Kumar, Abhishek and Ermon, Stefano and Poole, Ben},
  journal={arXiv preprint arXiv:2011.13456},
  year={2020}
}

@inproceedings{zhang2022langevin,
  title={A langevin-like sampler for discrete distributions},
  author={Zhang, Ruqi and Liu, Xingchao and Liu, Qiang},
  booktitle={International Conference on Machine Learning},
  pages={26375--26396},
  year={2022},
  organization={PMLR}
}

@inproceedings{sun2023discrete,
  title={Discrete langevin samplers via wasserstein gradient flow},
  author={Sun, Haoran and Dai, Hanjun and Dai, Bo and Zhou, Haomin and Schuurmans, Dale},
  booktitle={International Conference on Artificial Intelligence and Statistics},
  pages={6290--6313},
  year={2023},
  organization={PMLR}
}

@inproceedings{cheng2018underdamped,
  title={Underdamped Langevin MCMC: A non-asymptotic analysis},
  author={Cheng, Xiang and Chatterji, Niladri S and Bartlett, Peter L and Jordan, Michael I},
  booktitle={Conference on learning theory},
  pages={300--323},
  year={2018},
  organization={PMLR}
}

@inproceedings{cheng2018convergence,
  title={Convergence of Langevin MCMC in KL-divergence},
  author={Cheng, Xiang and Bartlett, Peter},
  booktitle={Algorithmic Learning Theory},
  pages={186--211},
  year={2018},
  organization={PMLR}
}

@article{roberts1996exponential,
author = {Gareth O. Roberts and Richard L. Tweedie},
title = {{Exponential convergence of Langevin distributions and their discrete approximations}},
volume = {2},
journal = {Bernoulli},
number = {4},
publisher = {Bernoulli Society for Mathematical Statistics and Probability},
pages = {341 -- 363},
year = {1996},
}

@article{foster2021shifted,
  title={The shifted ODE method for underdamped Langevin MCMC},
  author={Foster, James and Lyons, Terry and Oberhauser, Harald},
  journal={arXiv preprint arXiv:2101.03446},
  year={2021}
}

@article{balakrishnan1979fluctuation,
  title={Fluctuation-dissipation theorems from the generalised Langevin equation},
  author={Balakrishnan, Venkataraman},
  journal={Pramana},
  volume={12},
  pages={301--315},
  year={1979},
  publisher={Springer}
}

@book{risken1989fokker, 
    address = {Berlin},
    author = {Risken, Hannes},
    edition = {Second Edition},
    editor = {Haken, Hermann},
    publisher = {Springer},
    title = {The {Fokker-Planck} Equation: Methods of Solution and Applications},
    year = {1989}}

@article{bennett1982thermodynamics,
  title={The thermodynamics of computation—a review},
  author={Bennett, Charles H},
  journal={International Journal of Theoretical Physics},
  volume={21},
  pages={905--940},
  year={1982},
  publisher={Springer}
}

@article{whitelam2024thermodynamic,
  title={Thermodynamic computing out of equilibrium},
  author={Whitelam, Stephen and Casert, Corneel},
  journal={arXiv preprint arXiv:2412.17183},
  year={2024}
}

@misc{koller2009probabilistic,
  title={Probabilistic Graphical Models: Principles and Techniques},
  author={Koller, Daphne},
  year={2009},
  publisher={The MIT Press}
}

@article{loeliger2004introduction,
  title={An introduction to factor graphs},
  author={Loeliger, H-A},
  journal={IEEE Signal Processing Magazine},
  volume={21},
  number={1},
  pages={28--41},
  year={2004},
  publisher={IEEE}
}

@article{cipra1987introduction,
  title={An introduction to the Ising model},
  author={Cipra, Barry A},
  journal={The American Mathematical Monthly},
  volume={94},
  number={10},
  pages={937--959},
  year={1987},
  publisher={Taylor \& Francis}
}

@article{du2024compositional,
  title={Compositional generative modeling: A single model is not all you need},
  author={Du, Yilun and Kaelbling, Leslie},
  journal={arXiv preprint arXiv:2402.01103},
  year={2024}
}

@article{de2015rapidly,
  title={Rapidly mixing gibbs sampling for a class of factor graphs using hierarchy width},
  author={De Sa, Christopher M and Zhang, Ce and Olukotun, Kunle and R{\'e}, Christopher},
  journal={Advances in neural information processing systems},
  volume={28},
  year={2015}
}

@inproceedings{gonzalez2011parallel,
  title={Parallel gibbs sampling: From colored fields to thin junction trees},
  author={Gonzalez, Joseph and Low, Yucheng and Gretton, Arthur and Guestrin, Carlos},
  booktitle={Proceedings of the Fourteenth International Conference on Artificial Intelligence and Statistics},
  pages={324--332},
  year={2011},
  organization={JMLR Workshop and Conference Proceedings}
}

@inproceedings{terenin2020asynchronous,
  title={Asynchronous gibbs sampling},
  author={Terenin, Alexander and Simpson, Daniel and Draper, David},
  booktitle={International Conference on Artificial Intelligence and Statistics},
  pages={144--154},
  year={2020},
  organization={PMLR}
}

@article{daskalakis2018hogwild,
  title={Hogwild!-gibbs can be panaccurate},
  author={Daskalakis, Constantinos and Dikkala, Nishanth and Jayanti, Siddhartha},
  journal={Advances in Neural Information Processing Systems},
  volume={31},
  year={2018}
}

@article{shi2024universal,
  title={On universal inference in gaussian mixture models},
  author={Shi, Hongjian and Drton, Mathias},
  journal={arXiv preprint arXiv:2407.19361},
  year={2024}
}

@article{dandi2024universality,
  title={Universality laws for gaussian mixtures in generalized linear models},
  author={Dandi, Yatin and Stephan, Ludovic and Krzakala, Florent and Loureiro, Bruno and Zdeborov{\'a}, Lenka},
  journal={Advances in Neural Information Processing Systems},
  volume={36},
  year={2024}
}

@article{lahiri2016universal,
  title={A universal tradeoff between power, precision and speed in physical communication},
  author={Lahiri, Subhaneil and Sohl-Dickstein, Jascha and Ganguli, Surya},
  journal={arXiv preprint arXiv:1603.07758},
  year={2016}
}

@article{ramachandran2017searching,
  title={Searching for activation functions},
  author={Ramachandran, Prajit and Zoph, Barret and Le, Quoc V},
  journal={arXiv preprint arXiv:1710.05941},
  year={2017}
}

@article{vaswani2017attention,
  author       = {Ashish Vaswani and
                  Noam Shazeer and
                  Niki Parmar and
                  Jakob Uszkoreit and
                  Llion Jones and
                  Aidan N. Gomez and
                  Lukasz Kaiser and
                  Illia Polosukhin},
  title        = {Attention Is All You Need},
  journal      = {CoRR},
  volume       = {abs/1706.03762},
  year         = {2017},
  url          = {http://arxiv.org/abs/1706.03762},
  eprinttype    = {arXiv},
  eprint       = {1706.03762},
  bibsource    = {dblp computer science bibliography, https://dblp.org}
}

@inproceedings{xiong2020layer,
  title={On layer normalization in the transformer architecture},
  author={Xiong, Ruibin and Yang, Yunchang and He, Di and Zheng, Kai and Zheng, Shuxin and Xing, Chen and Zhang, Huishuai and Lan, Yanyan and Wang, Liwei and Liu, Tieyan},
  booktitle={International Conference on Machine Learning},
  pages={10524--10533},
  year={2020},
  organization={PMLR}
}

@misc{ba2016layernormalization,
      title={Layer Normalization}, 
      author={Jimmy Lei Ba and Jamie Ryan Kiros and Geoffrey E. Hinton},
      year={2016},
      eprint={1607.06450},
      archivePrefix={arXiv},
      primaryClass={stat.ML},
      url={https://arxiv.org/abs/1607.06450}, 
}

@article{lin2022survey,
  title={A survey of transformers},
  author={Lin, Tianyang and Wang, Yuxin and Liu, Xiangyang and Qiu, Xipeng},
  journal={AI open},
  volume={3},
  pages={111--132},
  year={2022},
  publisher={Elsevier}
}

@article{khan2023survey,
  title={A survey of the vision transformers and their CNN-transformer based variants},
  author={Khan, Asifullah and Rauf, Zunaira and Sohail, Anabia and Khan, Abdul Rehman and Asif, Hifsa and Asif, Aqsa and Farooq, Umair},
  journal={Artificial Intelligence Review},
  volume={56},
  number={Suppl 3},
  pages={2917--2970},
  year={2023},
  publisher={Springer}
}

@article{min2022transformer,
  title={Transformer for graphs: An overview from architecture perspective},
  author={Min, Erxue and Chen, Runfa and Bian, Yatao and Xu, Tingyang and Zhao, Kangfei and Huang, Wenbing and Zhao, Peilin and Huang, Junzhou and Ananiadou, Sophia and Rong, Yu},
  journal={arXiv preprint arXiv:2202.08455},
  year={2022}
}

@article{paik2005cooper,
  title={Cooper-pair box as a variable capacitor},
  author={Paik, Hanhee and Strauch, FW and Ramos, RC and Berkley, AJ and Xu, H and Dutta, SK and Johnson, PR and Dragt, AJ and Anderson, JR and Lobb, CJ and others},
  journal={IEEE transactions on applied superconductivity},
  volume={15},
  number={2},
  pages={884--887},
  year={2005},
  publisher={IEEE}
}

@article{dubey2024llama,
  title={The llama 3 herd of models},
  author={Dubey, Abhimanyu and Jauhri, Abhinav and Pandey, Abhinav and Kadian, Abhishek and Al-Dahle, Ahmad and Letman, Aiesha and Mathur, Akhil and Schelten, Alan and Yang, Amy and Fan, Angela and others},
  journal={arXiv preprint arXiv:2407.21783},
  year={2024}
}

@article{liu2019neural,
  title={Neural belief-propagation decoders for quantum error-correcting codes},
  author={Liu, Ye-Hua and Poulin, David},
  journal={Physical review letters},
  volume={122},
  number={20},
  pages={200501},
  year={2019},
  publisher={APS}
}

@article{old2023generalized,
  title={Generalized belief propagation algorithms for decoding of surface codes},
  author={Old, Josias and Rispler, Manuel},
  journal={Quantum},
  volume={7},
  pages={1037},
  year={2023},
  publisher={Verein zur F{\"o}rderung des Open Access Publizierens in den Quantenwissenschaften}
}

@article{agrawal2023sarathi,
  title={Sarathi: Efficient llm inference by piggybacking decodes with chunked prefills},
  author={Agrawal, Amey and Panwar, Ashish and Mohan, Jayashree and Kwatra, Nipun and Gulavani, Bhargav S and Ramjee, Ramachandran},
  journal={arXiv preprint arXiv:2308.16369},
  year={2023}
}

@article{yuan2024llm,
  title={Llm inference unveiled: Survey and roofline model insights},
  author={Yuan, Zhihang and Shang, Yuzhang and Zhou, Yang and Dong, Zhen and Zhou, Zhe and Xue, Chenhao and Wu, Bingzhe and Li, Zhikai and Gu, Qingyi and Lee, Yong Jae and others},
  journal={arXiv preprint arXiv:2402.16363},
  year={2024}
}

@article{bolte2020mathematical,
  title={A mathematical model for automatic differentiation in machine learning},
  author={Bolte, J{\'e}r{\^o}me and Pauwels, Edouard},
  journal={Advances in Neural Information Processing Systems},
  volume={33},
  pages={10809--10819},
  year={2020}
}

@inproceedings{lecun1988theoretical,
  title={A theoretical framework for back-propagation},
  author={LeCun, Yann and Touresky, D and Hinton, G and Sejnowski, T},
  booktitle={Proceedings of the 1988 connectionist models summer school},
  volume={1},
  pages={21--28},
  year={1988}
}

@article{risken1985eigenvalues,
  title={Eigenvalues and eigenfunctions of the Fokker-Planck equation for the extremely underdamped Brownian motion in a double-well potential},
  author={Risken, H and Voigtlaender, K},
  journal={Journal of statistical physics},
  volume={41},
  pages={825--863},
  year={1985},
  publisher={Springer}
}

@article{wimsatt2021harnessing,
  title={Harnessing fluctuations in thermodynamic computing via time-reversal symmetries},
  author={Wimsatt, Gregory and Saira, Olli-Pentti and Boyd, Alexander B and Matheny, Matthew H and Han, Siyuan and Roukes, Michael L and Crutchfield, James P},
  journal={Physical Review Research},
  volume={3},
  number={3},
  pages={033115},
  year={2021},
  publisher={APS}
}

@inproceedings{jarzynski2012equalities,
  title={Equalities and inequalities: Irreversibility and the second law of thermodynamics at the nanoscale},
  author={Jarzynski, Christopher},
  booktitle={Time: Poincar{\'e} Seminar 2010},
  pages={145--172},
  year={2012},
  organization={Springer}
}

@article{vaikuntanathan2011escorted,
  title={Escorted free energy simulations},
  author={Vaikuntanathan, Suriyanarayanan and Jarzynski, Christopher},
  journal={The Journal of chemical physics},
  volume={134},
  number={5},
  year={2011},
  publisher={AIP Publishing}
}

@article{park2003free,
  title={Free energy calculation from steered molecular dynamics simulations using Jarzynski’s equality},
  author={Park, Sanghyun and Khalili-Araghi, Fatemeh and Tajkhorshid, Emad and Schulten, Klaus},
  journal={The Journal of chemical physics},
  volume={119},
  number={6},
  pages={3559--3566},
  year={2003},
  publisher={American Institute of Physics}
}

@article{ray2023gigahertz,
  title={Gigahertz sub-Landauer momentum computing},
  author={Ray, Kyle J and Crutchfield, James P},
  journal={Physical Review Applied},
  volume={19},
  number={1},
  pages={014049},
  year={2023},
  publisher={APS}
}

@article{brush1967history,
  title={History of the Lenz-Ising model},
  author={Brush, Stephen G},
  journal={Reviews of modern physics},
  volume={39},
  number={4},
  pages={883},
  year={1967},
  publisher={APS}
}

@article{patel2020ising,
  title={Ising model optimization problems on a FPGA accelerated restricted Boltzmann machine},
  author={Patel, Saavan and Chen, Lili and Canoza, Philip and Salahuddin, Sayeef},
  journal={arXiv preprint arXiv:2008.04436},
  year={2020}
}

@inproceedings{salakhutdinov2009deep,
  title={Deep boltzmann machines},
  author={Salakhutdinov, Ruslan and Hinton, Geoffrey},
  booktitle={Artificial intelligence and statistics},
  pages={448--455},
  year={2009},
  organization={PMLR}
}

@article{ackley1985learning,
  title={A learning algorithm for Boltzmann machines},
  author={Ackley, David H and Hinton, Geoffrey E and Sejnowski, Terrence J},
  journal={Cognitive science},
  volume={9},
  number={1},
  pages={147--169},
  year={1985},
  publisher={Elsevier}
}

@article{nikhar2024all,
  title={All-to-all reconfigurability with sparse and higher-order Ising machines},
  author={Nikhar, Srijan and Kannan, Sidharth and Aadit, Navid Anjum and Chowdhury, Shuvro and Camsari, Kerem Y},
  journal={Nature Communications},
  volume={15},
  number={1},
  pages={8977},
  year={2024},
  publisher={Nature Publishing Group UK London}
}

@incollection{bishop1998latent,
  title={Latent variable models},
  author={Bishop, Christopher M},
  booktitle={Learning in graphical models},
  pages={371--403},
  year={1998},
  publisher={Springer}
}

@article{golubov2004current,
  title={The current-phase relation in Josephson junctions},
  author={Golubov, Alexandre Avraamovitch and Kupriyanov, M Yu and Il’Ichev, E},
  journal={Reviews of modern physics},
  volume={76},
  number={2},
  pages={411},
  year={2004},
  publisher={APS}
}

@book{sarkka2019applied,
  title={Applied stochastic differential equations},
  author={S{\"a}rkk{\"a}, Simo and Solin, Arno},
  volume={10},
  year={2019},
  publisher={Cambridge University Press}
}

@article{fredkin1982conservative,
  title={Conservative logic},
  author={Fredkin, Edward and Toffoli, Tommaso},
  journal={International Journal of theoretical physics},
  volume={21},
  number={3},
  pages={219--253},
  year={1982},
  publisher={Springer}
}

@article{vaccaro2011information,
  title={Information erasure without an energy cost},
  author={Vaccaro, Joan A and Barnett, Stephen M},
  journal={Proceedings of the Royal Society A: Mathematical, Physical and Engineering Sciences},
  volume={467},
  number={2130},
  pages={1770--1778},
  year={2011},
  publisher={The Royal Society Publishing}
}

@article{drton2017structure,
  title={Structure learning in graphical modeling},
  author={Drton, Mathias and Maathuis, Marloes H},
  journal={Annual Review of Statistics and Its Application},
  volume={4},
  number={1},
  pages={365--393},
  year={2017},
  publisher={Annual Reviews}
}

@inproceedings{he2016deep,
  title={Deep residual learning for image recognition},
  author={He, Kaiming and Zhang, Xiangyu and Ren, Shaoqing and Sun, Jian},
  booktitle={Proceedings of the IEEE conference on computer vision and pattern recognition},
  pages={770--778},
  year={2016}
}

@article{kaiser2021probabilistic,
  title={Probabilistic computing with p-bits},
  author={Kaiser, Jan and Datta, Supriyo},
  journal={Applied Physics Letters},
  volume={119},
  number={15},
  year={2021},
  publisher={AIP Publishing}
}

@article{chowdhury2023full,
  title={A full-stack view of probabilistic computing with p-bits: devices, architectures, and algorithms},
  author={Chowdhury, Shuvro and Grimaldi, Andrea and Aadit, Navid Anjum and Niazi, Shaila and Mohseni, Masoud and Kanai, Shun and Ohno, Hideo and Fukami, Shunsuke and Theogarajan, Luke and Finocchio, Giovanni and others},
  journal={IEEE Journal on Exploratory Solid-State Computational Devices and Circuits},
  volume={9},
  number={1},
  pages={1--11},
  year={2023},
  publisher={IEEE}
}

@article{camsari2019p,
  title={P-bits for probabilistic spin logic},
  author={Camsari, Kerem Y and Sutton, Brian M and Datta, Supriyo},
  journal={Applied Physics Reviews},
  volume={6},
  number={1},
  year={2019},
  publisher={AIP Publishing}
}

@article{fortunato2018noisy,
  author       = {Meire Fortunato and
                  Mohammad Gheshlaghi Azar and
                  Bilal Piot and
                  Jacob Menick and
                  Ian Osband and
                  Alex Graves and
                  Vlad Mnih and
                  R{\'{e}}mi Munos and
                  Demis Hassabis and
                  Olivier Pietquin and
                  Charles Blundell and
                  Shane Legg},
  title        = {Noisy Networks for Exploration},
  journal      = {CoRR},
  volume       = {abs/1706.10295},
  year         = {2017},
  url          = {http://arxiv.org/abs/1706.10295},
  eprinttype    = {arXiv},
  eprint       = {1706.10295},
  bibsource    = {dblp computer science bibliography, https://dblp.org}
}

@article{plappert2017parameter,
  title={Parameter space noise for exploration},
  author={Plappert, Matthias and Houthooft, Rein and Dhariwal, Prafulla and Sidor, Szymon and Chen, Richard Y and Chen, Xi and Asfour, Tamim and Abbeel, Pieter and Andrychowicz, Marcin},
  journal={arXiv preprint arXiv:1706.01905},
  year={2017}
}

@inproceedings{eberhard2023pink,
  title={Pink noise is all you need: Colored noise exploration in deep reinforcement learning},
  author={Eberhard, Onno and Hollenstein, Jakob and Pinneri, Cristina and Martius, Georg},
  booktitle={The Eleventh International Conference on Learning Representations},
  year={2023}
}

@inproceedings{gal2016dropout,
  title={Dropout as a bayesian approximation: Representing model uncertainty in deep learning},
  author={Gal, Yarin and Ghahramani, Zoubin},
  booktitle={international conference on machine learning},
  pages={1050--1059},
  year={2016},
  organization={PMLR}
}

@inproceedings{lockwood2022review,
  title={A review of uncertainty for deep reinforcement learning},
  author={Lockwood, Owen and Si, Mei},
  booktitle={Proceedings of the AAAI Conference on Artificial Intelligence and Interactive Digital Entertainment},
  volume={18},
  pages={155--162},
  year={2022}
}

@article{gal2017concrete,
  title={Concrete dropout},
  author={Gal, Yarin and Hron, Jiri and Kendall, Alex},
  journal={Advances in neural information processing systems},
  volume={30},
  year={2017}
}

@misc{sekimoto2010stochastic,
  title={Stochastic energetics},
  author={Sekimoto, Ken},
  year={2010},
  publisher={Springer}
}

@article{shiraishi2023introduction,
  title={An Introduction to Stochastic Thermodynamics},
  author={Shiraishi, Naoto},
  journal={Fundamental Theories of Physics. Springer, Singapore},
  year={2023},
  publisher={Springer}
}

@article{srivastava2014dropout,
  title={Dropout: a simple way to prevent neural networks from overfitting},
  author={Srivastava, Nitish and Hinton, Geoffrey and Krizhevsky, Alex and Sutskever, Ilya and Salakhutdinov, Ruslan},
  journal={The journal of machine learning research},
  volume={15},
  number={1},
  pages={1929--1958},
  year={2014},
  publisher={JMLR. org}
}

@article{shen2017continuous,
  title={Continuous dropout},
  author={Shen, Xu and Tian, Xinmei and Liu, Tongliang and Xu, Fang and Tao, Dacheng},
  journal={IEEE transactions on neural networks and learning systems},
  volume={29},
  number={9},
  pages={3926--3937},
  year={2017},
  publisher={IEEE}
}

@article{kingma2015variational,
  title={Variational dropout and the local reparameterization trick},
  author={Kingma, Durk P and Salimans, Tim and Welling, Max},
  journal={Advances in neural information processing systems},
  volume={28},
  year={2015}
}

@inproceedings{molchanov2017variational,
  title={Variational dropout sparsifies deep neural networks},
  author={Molchanov, Dmitry and Ashukha, Arsenii and Vetrov, Dmitry},
  booktitle={International conference on machine learning},
  pages={2498--2507},
  year={2017},
  organization={PMLR}
}

@article{arya2022automatic,
  title={Automatic differentiation of programs with discrete randomness},
  author={Arya, Gaurav and Schauer, Moritz and Sch{\"a}fer, Frank and Rackauckas, Christopher},
  journal={Advances in Neural Information Processing Systems},
  volume={35},
  pages={10435--10447},
  year={2022}
}

@article{schafer2021abstractdifferentiation,
  title={Abstractdifferentiation. jl: Backend-agnostic differentiable programming in Julia},
  author={Sch{\"a}fer, Frank and Tarek, Mohamed and White, Lyndon and Rackauckas, Chris},
  journal={arXiv preprint arXiv:2109.12449},
  year={2021}
}

@article{arya2023differentiating,
  title={Differentiating Metropolis-Hastings to optimize intractable densities},
  author={Arya, Gaurav and Seyer, Ruben and Sch{\"a}fer, Frank and Chandra, Kartik and Lew, Alexander K and Huot, Mathieu and Mansinghka, Vikash K and Ragan-Kelley, Jonathan and Rackauckas, Christopher and Schauer, Moritz},
  journal={arXiv preprint arXiv:2306.07961},
  year={2023}
}

@article{baydin2018automatic,
  title={Automatic differentiation in machine learning: a survey},
  author={Baydin, Atilim Gunes and Pearlmutter, Barak A and Radul, Alexey Andreyevich and Siskind, Jeffrey Mark},
  journal={Journal of machine learning research},
  volume={18},
  number={153},
  pages={1--43},
  year={2018}
}

@article{margossian2019review,
  title={A review of automatic differentiation and its efficient implementation},
  author={Margossian, Charles C},
  journal={Wiley interdisciplinary reviews: data mining and knowledge discovery},
  volume={9},
  number={4},
  pages={e1305},
  year={2019},
  publisher={Wiley Online Library}
}

@inproceedings{moses2021reverse,
  title={Reverse-mode automatic differentiation and optimization of GPU kernels via Enzyme},
  author={Moses, William S and Churavy, Valentin and Paehler, Ludger and H{\"u}ckelheim, Jan and Narayanan, Sri Hari Krishna and Schanen, Michel and Doerfert, Johannes},
  booktitle={Proceedings of the international conference for high performance computing, networking, storage and analysis},
  pages={1--16},
  year={2021}
}

@article{schafer2020spectral,
  title={Spectral structure and many-body dynamics of ultracold bosons in a double-well},
  author={Sch{\"a}fer, Frank and Bastarrachea-Magnani, Miguel A and Lode, Axel UJ and de Parny, Laurent de Forges and Buchleitner, Andreas},
  journal={Entropy},
  volume={22},
  number={4},
  pages={382},
  year={2020},
  publisher={MDPI}
}

@article{borah2021measurement,
  title={Measurement-based feedback quantum control with deep reinforcement learning for a double-well nonlinear potential},
  author={Borah, Sangkha and Sarma, Bijita and Kewming, Michael and Milburn, Gerard J and Twamley, Jason},
  journal={Physical review letters},
  volume={127},
  number={19},
  pages={190403},
  year={2021},
  publisher={APS}
}

@article{wu2022nonequilibrium,
  title={Nonequilibrium quantum thermodynamics of a particle trapped in a controllable time-varying potential},
  author={Wu, Qiongyuan and Mancino, Luca and Carlesso, Matteo and Ciampini, Mario A and Magrini, Lorenzo and Kiesel, Nikolai and Paternostro, Mauro},
  journal={PRX Quantum},
  volume={3},
  number={1},
  pages={010322},
  year={2022},
  publisher={APS}
}

@article{bickson2008gaussian,
  title={Gaussian belief propagation: Theory and aplication},
  author={Bickson, Danny},
  journal={arXiv preprint arXiv:0811.2518},
  year={2008}
}

@article{su2015convergence,
  title={On convergence conditions of Gaussian belief propagation},
  author={Su, Qinliang and Wu, Yik-Chung},
  journal={IEEE Transactions on Signal Processing},
  volume={63},
  number={5},
  pages={1144--1155},
  year={2015},
  publisher={IEEE}
}

@article{ortiz2021visual,
  title={A visual introduction to Gaussian belief propagation},
  author={Ortiz, Joseph and Evans, Talfan and Davison, Andrew J},
  journal={arXiv preprint arXiv:2107.02308},
  year={2021}
}

@inproceedings{satorras2021neural,
  title={Neural enhanced belief propagation on factor graphs},
  author={Satorras, Victor Garcia and Welling, Max},
  booktitle={International Conference on Artificial Intelligence and Statistics},
  pages={685--693},
  year={2021},
  organization={PMLR}
}

@phdthesis{ortiz2023gaussian,
  title={Gaussian belief propagation for real-time decentralised inference},
  author={Ortiz, Joseph},
  year={2023},
  school={Imperial College London}
}

@inproceedings{liang2021neural,
  title={Neural enhanced belief propagation for cooperative localization},
  author={Liang, Mingchao and Meyer, Florian},
  booktitle={2021 IEEE Statistical Signal Processing Workshop (SSP)},
  pages={326--330},
  year={2021},
  organization={IEEE}
}

@article{liang2023neural,
  title={Neural enhanced belief propagation for multiobject tracking},
  author={Liang, Mingchao and Meyer, Florian},
  journal={IEEE Transactions on Signal Processing},
  year={2023},
  publisher={IEEE}
}

@article{patwardhan2022distributing,
  title={Distributing collaborative multi-robot planning with Gaussian belief propagation},
  author={Patwardhan, Aalok and Murai, Riku and Davison, Andrew J},
  journal={IEEE Robotics and Automation Letters},
  volume={8},
  number={2},
  pages={552--559},
  year={2022},
  publisher={IEEE}
}

@article{kudithipudi2025neuromorphic,
  title={Neuromorphic computing at scale},
  author={Kudithipudi, Dhireesha and Schuman, Catherine and Vineyard, Craig M and Pandit, Tej and Merkel, Cory and Kubendran, Rajkumar and Aimone, James B and Orchard, Garrick and Mayr, Christian and Benosman, Ryad and others},
  journal={Nature},
  volume={637},
  number={8047},
  pages={801--812},
  year={2025},
  publisher={Nature Publishing Group UK London}
}

@book{robert1999monte,
  title={Monte Carlo statistical methods},
  author={Robert, Christian P and Casella, George and Casella, George},
  volume={2},
  year={1999},
  publisher={Springer}
}

@article{rabiner1989tutorial,
  title={A tutorial on hidden Markov models and selected applications in speech recognition},
  author={Rabiner, Lawrence R},
  journal={Proceedings of the IEEE},
  volume={77},
  number={2},
  pages={257--286},
  year={1989},
  publisher={Ieee}
}

@article{eddy2004hidden,
  title={What is a hidden Markov model?},
  author={Eddy, Sean R},
  journal={Nature biotechnology},
  volume={22},
  number={10},
  pages={1315--1316},
  year={2004},
  publisher={Nature Publishing Group UK London}
}

@article{krogh1994hidden,
  title={Hidden Markov models in computational biology: Applications to protein modeling},
  author={Krogh, Anders and Brown, Michael and Mian, I Saira and Sj{\"o}lander, Kimmen and Haussler, David},
  journal={Journal of molecular biology},
  volume={235},
  number={5},
  pages={1501--1531},
  year={1994},
  publisher={Elsevier}
}

@book{mamon2007hidden,
  title={Hidden Markov models in finance},
  author={Mamon, Rogemar S and Elliott, Robert James},
  volume={4},
  year={2007},
  publisher={Springer}
}

@article{he2018unsupervised,
  title={Unsupervised learning of syntactic structure with invertible neural projections},
  author={He, Junxian and Neubig, Graham and Berg-Kirkpatrick, Taylor},
  journal={arXiv preprint arXiv:1808.09111},
  year={2018}
}

@article{ghosh2021normalizing,
  title={Normalizing flow based hidden Markov models for classification of speech phones with explainability},
  author={Ghosh, Anubhab and Honor{\'e}, Antoine and Liu, Dong and Henter, Gustav Eje and Chatterjee, Saikat},
  journal={arXiv preprint arXiv:2107.00730},
  year={2021}
}

@article{liu2019powering,
  author       = {Dong Liu and
                  Antoine Honor{\'{e}} and
                  Saikat Chatterjee and
                  Lars K. Rasmussen},
  title        = {Powering Hidden Markov Model by Neural Network based Generative Models},
  journal      = {CoRR},
  volume       = {abs/1910.05744},
  year         = {2019},
  url          = {http://arxiv.org/abs/1910.05744},
  eprinttype    = {arXiv},
  eprint       = {1910.05744},
  bibsource    = {dblp computer science bibliography, https://dblp.org}
}

@article{azeraf2021introducing,
  title={Introducing the hidden neural markov chain framework},
  author={Azeraf, Elie and Monfrini, Emmanuel and Vignon, Emmanuel and Pieczynski, Wojciech},
  journal={arXiv preprint arXiv:2102.11038},
  year={2021}
}

@article{kramers1940brownian,
  title={Brownian motion in a field of force and the diffusion model of chemical reactions},
  author={Kramers, Hendrik Anthony},
  journal={physica},
  volume={7},
  number={4},
  pages={284--304},
  year={1940},
  publisher={Elsevier}
}

@article{buttiker1983thermal,
  title={Thermal activation in extremely underdamped Josephson-junction circuits},
  author={B{\"u}ttiker, M and Harris, EP and Landauer, R},
  journal={Physical Review B},
  volume={28},
  number={3},
  pages={1268},
  year={1983},
  publisher={APS}
}

@article{ito2024geometric,
  title={Geometric thermodynamics for the Fokker--Planck equation: stochastic thermodynamic links between information geometry and optimal transport},
  author={Ito, Sosuke},
  journal={Information Geometry},
  volume={7},
  number={Suppl 1},
  pages={441--483},
  year={2024},
  publisher={Springer}
}

@article{nakazato2021geometrical,
  title={Geometrical aspects of entropy production in stochastic thermodynamics based on Wasserstein distance},
  author={Nakazato, Muka and Ito, Sosuke},
  journal={Physical Review Research},
  volume={3},
  number={4},
  pages={043093},
  year={2021},
  publisher={APS}
}

@article{ito2018stochastic,
  title={Stochastic thermodynamic interpretation of information geometry},
  author={Ito, Sosuke},
  journal={Physical review letters},
  volume={121},
  number={3},
  pages={030605},
  year={2018},
  publisher={APS}
}

@article{ito2020stochastic,
  title={Stochastic time evolution, information geometry, and the Cram{\'e}r-Rao bound},
  author={Ito, Sosuke and Dechant, Andreas},
  journal={Physical Review X},
  volume={10},
  number={2},
  pages={021056},
  year={2020},
  publisher={APS}
}

@book{villani2009optimal,
  title={Optimal transport: old and new},
  author={Villani, C{\'e}dric and others},
  volume={338},
  year={2009},
  publisher={Springer}
}

@article{josephson1962possible,
	title = {Possible new effects in superconductive tunnelling},
	volume = {1},
	issn = {0031-9163},
	url = {https://www.sciencedirect.com/science/article/pii/0031916362913690},
	doi = {10.1016/0031-9163(62)91369-0},
	number = {7},
	urldate = {2024-11-11},
	journal = {Physics Letters},
	author = {Josephson, B. D.},
	month = jul,
	year = {1962},
	pages = {251--253},
}

@article{josephson1974discovery,
	title = {The discovery of tunnelling supercurrents},
	volume = {46},
	url = {https://link.aps.org/doi/10.1103/RevModPhys.46.251},
	doi = {10.1103/RevModPhys.46.251},
	number = {2},
	urldate = {2024-11-11},
	journal = {Reviews of Modern Physics},
	author = {Josephson, B. D.},
	month = apr,
	year = {1974},
	note = {Publisher: American Physical Society},
	pages = {251--254},
}

@phdthesis{quintana2017superconducting,
	address = {Santa Barbara, California, USA},
	type = {Ph.{D}.},
	title = {Superconducting flux qubits for high-connectivity quantum annealing without lossy dielectrics},
	url = {https://escholarship.org/uc/item/9844c3h3},
	urldate = {2023-02-01},
	school = {UC Santa Barbara},
	author = {Quintana, Christopher},
	year = {2017},
}

@misc{novikov2018exploring,
	title = {Exploring {More}-{Coherent} {Quantum} {Annealing}},
	url = {http://arxiv.org/abs/1809.04485},
	urldate = {2024-10-24},
	publisher = {arXiv},
	author = {Novikov, Sergey and Hinkey, Robert and Disseler, Steven and Basham, James I. and Albash, Tameem and Risinger, Andrew and Ferguson, David and Lidar, Daniel A. and Zick, Kenneth M.},
	month = sep,
	year = {2018},
	note = {arXiv:1809.04485},
}

@article{harris2010experimental,
	title = {Experimental demonstration of a robust and scalable flux qubit},
	volume = {81},
	url = {https://link.aps.org/doi/10.1103/PhysRevB.81.134510},
	doi = {10.1103/PhysRevB.81.134510},
	number = {13},
	urldate = {2023-01-11},
	journal = {Physical Review B},
	author = {Harris, R. and Johansson, J. and Berkley, A. J. and Johnson, M. W. and Lanting, T. and Han, Siyuan and Bunyk, P. and Ladizinsky, E. and Oh, T. and Perminov, I. and Tolkacheva, E. and Uchaikin, S. and Chapple, E. M. and Enderud, C. and Rich, C. and Thom, M. and Wang, J. and Wilson, B. and Rose, G.},
	month = apr,
	year = {2010},
	note = {Publisher: American Physical Society},
	pages = {134510},
}

@misc{khezri2021anneal,
	title = {Anneal-path correction in flux qubits},
	url = {http://arxiv.org/abs/2002.11217},
	urldate = {2024-10-24},
	publisher = {arXiv},
	author = {Khezri, Mostafa and Grover, Jeffrey A. and Basham, James I. and Disseler, Steven M. and Chen, Huo and Novikov, Sergey and Zick, Kenneth M. and Lidar, Daniel A.},
	month = feb,
	year = {2021},
	note = {arXiv:2002.11217},
}

@article{han1992effect,
	title = {Effect of a two-dimensional potential on the rate of thermally induced escape over the potential barrier},
	volume = {46},
	url = {https://link.aps.org/doi/10.1103/PhysRevB.46.6338},
	doi = {10.1103/PhysRevB.46.6338},
	number = {10},
	urldate = {2023-02-18},
	journal = {Physical Review B},
	author = {Han, Siyuan and Lapointe, J. and Lukens, J. E.},
	month = sep,
	year = {1992},
	note = {Publisher: American Physical Society},
	pages = {6338--6345},
}

@article{pratt2025extracting,
	title = {Extracting equations of motion from superconducting circuits},
	volume = {7},
	url = {https://link.aps.org/doi/10.1103/PhysRevResearch.7.013014},
	doi = {10.1103/PhysRevResearch.7.013014},
	number = {1},
	urldate = {2025-01-10},
	journal = {Physical Review Research},
	author = {Pratt, Christian Z. and Ray, Kyle J. and Crutchfield, James P.},
	month = jan,
	year = {2025},
	note = {Publisher: American Physical Society},
	pages = {013014},
}

@inproceedings{nussbaum2019ising,
  title={Ising models with latent conditional gaussian variables},
  author={Nussbaum, Frank and Giesen, Joachim},
  booktitle={Algorithmic Learning Theory},
  pages={669--681},
  year={2019},
  organization={PMLR}
}

@article{dunn2013learning,
  title={Learning and inference in a nonequilibrium Ising model with hidden nodes},
  author={Dunn, Benjamin and Roudi, Yasser},
  journal={Physical Review E—Statistical, Nonlinear, and Soft Matter Physics},
  volume={87},
  number={2},
  pages={022127},
  year={2013},
  publisher={APS}
}

@article{lokhov2018optimal,
  title={Optimal structure and parameter learning of Ising models},
  author={Lokhov, Andrey Y and Vuffray, Marc and Misra, Sidhant and Chertkov, Michael},
  journal={Science advances},
  volume={4},
  number={3},
  pages={e1700791},
  year={2018},
  publisher={American Association for the Advancement of Science}
}

@inproceedings{bresler2015efficiently,
  title={Efficiently learning Ising models on arbitrary graphs},
  author={Bresler, Guy},
  booktitle={Proceedings of the forty-seventh annual ACM symposium on Theory of computing},
  pages={771--782},
  year={2015}
}

@article{nishikawa1976continuous,
  title={A Continuous Ising Model Exhibiting Phase Transitions of First or Second Order},
  author={Nishikawa, Ken-ichi and Nakano, Huzio},
  journal={Progress of Theoretical Physics},
  volume={56},
  number={3},
  pages={773--785},
  year={1976},
  publisher={Oxford University Press}
}

@article{van1978phase,
  title={Phase transitions for continuous-spin Ising ferromagnets},
  author={van Beijeren, Henk and Sylvester, Garrett S},
  journal={Journal of Functional Analysis},
  volume={28},
  number={2},
  pages={145--167},
  year={1978},
  publisher={Elsevier}
}

@article{bayong1999effect,
  title={Effect of long-range interactions on the critical behavior of the continuous Ising model},
  author={Bayong, E and Diep, HT},
  journal={Physical Review B},
  volume={59},
  number={18},
  pages={11919},
  year={1999},
  publisher={APS}
}

@misc{dynamax,
author = {Linderman, Scott W. and Chang, Peter and Harper-Donnelly, Giles and Kara, Aleyna and Li, Xinglong and Duran-Martin, Gerardo and Murphy, Kevin},
doi = {10.21105/joss.07069},
journal = {Journal of Open Source Software},
month = apr,
number = {108},
pages = {7069},
title = {{Dynamax: A Python package for probabilistic state space modeling with JAX}},
url = {https://joss.theoj.org/papers/10.21105/joss.07069},
volume = {10},
year = {2025}
}

@article{virtanen2020scipy,
  title={SciPy 1.0: fundamental algorithms for scientific computing in Python},
  author={Virtanen, Pauli and Gommers, Ralf and Oliphant, Travis E and Haberland, Matt and Reddy, Tyler and Cournapeau, David and Burovski, Evgeni and Peterson, Pearu and Weckesser, Warren and Bright, Jonathan and others},
  journal={Nature methods},
  volume={17},
  number={3},
  pages={261--272},
  year={2020},
  publisher={Nature Publishing Group}
}

@phdthesis{kidger2021on,
    title={{O}n {N}eural {D}ifferential {E}quations},
    author={Patrick Kidger},
    year={2021},
    school={University of Oxford},
}

@misc{jax2018github,
  author = {James Bradbury and Roy Frostig and Peter Hawkins and Matthew James Johnson and Chris Leary and Dougal Maclaurin and George Necula and Adam Paszke and Jake Vander{P}las and Skye Wanderman-{M}ilne and Qiao Zhang},
  title = {{JAX}: composable transformations of {P}ython+{N}um{P}y programs},
  url = {http://github.com/jax-ml/jax},
  version = {0.3.13},
  year = {2018},
}

@article{kidger2021equinox,
    author={Patrick Kidger and Cristian Garcia},
    title={{E}quinox: neural networks in {JAX} via callable {P}y{T}rees and filtered transformations},
    year={2021},
    journal={Differentiable Programming workshop at Neural Information Processing Systems 2021}
}

@misc{deepmind2020jax,
  title = {The {D}eep{M}ind {JAX} {E}cosystem},
  author = {DeepMind and Babuschkin, Igor and Baumli, Kate and Bell, Alison and Bhupatiraju, Surya and Bruce, Jake and Buchlovsky, Peter and Budden, David and Cai, Trevor and Clark, Aidan and Danihelka, Ivo and Dedieu, Antoine and Fantacci, Claudio and Godwin, Jonathan and Jones, Chris and Hemsley, Ross and Hennigan, Tom and Hessel, Matteo and Hou, Shaobo and Kapturowski, Steven and Keck, Thomas and Kemaev, Iurii and King, Michael and Kunesch, Markus and Martens, Lena and Merzic, Hamza and Mikulik, Vladimir and Norman, Tamara and Papamakarios, George and Quan, John and Ring, Roman and Ruiz, Francisco and Sanchez, Alvaro and Sartran, Laurent and Schneider, Rosalia and Sezener, Eren and Spencer, Stephen and Srinivasan, Srivatsan and Stanojevi\'{c}, Milo\v{s} and Stokowiec, Wojciech and Wang, Luyu and Zhou, Guangyao and Viola, Fabio},
  url = {http://github.com/google-deepmind},
  year = {2020},
}

@misc{lockwood2024distreqx,
 title        = {distreqx: Distributions and Bijectors in Jax},
  author       = {Owen Lockwood},
  howpublished = {\url{https://github.com/lockwo/distreqx}},
  doi = {10.5281/zenodo.13764512},
  year         = {2024}
}

@article{aifer2024thermodynamic2,
  title={Thermodynamic Bayesian Inference},
  author={Aifer, Maxwell and Duffield, Samuel and Donatella, Kaelan and Melanson, Denis and Klett, Phoebe and Belateche, Zach and Crooks, Gavin and Martinez, Antonio J and Coles, Patrick J},
  journal={arXiv preprint arXiv:2410.01793},
  year={2024}
}

@article{donatella2024thermodynamic,
  title={Thermodynamic Natural Gradient Descent},
  author={Donatella, Kaelan and Duffield, Samuel and Aifer, Maxwell and Melanson, Denis and Crooks, Gavin and Coles, Patrick J},
  journal={arXiv preprint arXiv:2405.13817},
  year={2024}
}

@article{melanson2023thermodynamic,
  title={Thermodynamic computing system for AI applications},
  author={Melanson, Denis and Khater, Mohammad Abu and Aifer, Maxwell and Donatella, Kaelan and Gordon, Max Hunter and Ahle, Thomas and Crooks, Gavin and Martinez, Antonio J and Sbahi, Faris and Coles, Patrick J},
  journal={arXiv preprint arXiv:2312.04836},
  year={2023}
}

@article{martens2020new,
  title={New insights and perspectives on the natural gradient method},
  author={Martens, James},
  journal={Journal of Machine Learning Research},
  volume={21},
  number={146},
  pages={1--76},
  year={2020}
}

@misc{debos2025learningmultifieldcoherentising,
      title={Learning in a Multifield Coherent Ising Machine}, 
      author={Daan de Bos and Marc Serra-Garcia},
      year={2025},
      eprint={2502.12020},
      archivePrefix={arXiv},
      primaryClass={cond-mat.mes-hall},
      url={https://arxiv.org/abs/2502.12020}, 
}

@article{grabert1984crossover,
	title = {Crossover from {Thermal} {Hopping} to {Quantum} {Tunneling}},
	volume = {53},
	url = {https://link.aps.org/doi/10.1103/PhysRevLett.53.1787},
	doi = {10.1103/PhysRevLett.53.1787},
	number = {19},
	urldate = {2023-04-03},
	journal = {Physical Review Letters},
	author = {Grabert, Hermann and Weiss, Ulrich},
	month = nov,
	year = {1984},
	note = {Publisher: American Physical Society},
	pages = {1787--1790},
}

@article{hanggi1985quantum,
	title = {Quantum {Theory} of {Activated} {Events} in {Presence} of {Long}-{Time} {Memory}},
	volume = {55},
	url = {https://link.aps.org/doi/10.1103/PhysRevLett.55.761},
	doi = {10.1103/PhysRevLett.55.761},
	number = {7},
	urldate = {2023-04-03},
	journal = {Physical Review Letters},
	author = {Hanggi, Peter and Grabert, Hermann and Ingold, Gert-Ludwig and Weiss, Ulrich},
	month = aug,
	year = {1985},
	note = {Publisher: American Physical Society},
	pages = {761--764},
}

@article{devoret1985measurements,
	title = {Measurements of {Macroscopic} {Quantum} {Tunneling} out of the {Zero}-{Voltage} {State} of a {Current}-{Biased} {Josephson} {Junction}},
	volume = {55},
	url = {https://link.aps.org/doi/10.1103/PhysRevLett.55.1908},
	doi = {10.1103/PhysRevLett.55.1908},
	number = {18},
	urldate = {2024-12-17},
	journal = {Physical Review Letters},
	author = {Devoret, Michel H. and Martinis, John M. and Clarke, John},
	month = oct,
	year = {1985},
	note = {Publisher: American Physical Society},
	pages = {1908--1911},
}

@article{li2002quantitative,
	title = {Quantitative {Study} of {Macroscopic} {Quantum} {Tunneling} in a dc {SQUID}: {A} {System} with {Two} {Degrees} of {Freedom}},
	volume = {89},
	shorttitle = {Quantitative {Study} of {Macroscopic} {Quantum} {Tunneling} in a dc {SQUID}},
	url = {https://link.aps.org/doi/10.1103/PhysRevLett.89.098301},
	doi = {10.1103/PhysRevLett.89.098301},
	number = {9},
	urldate = {2024-12-17},
	journal = {Physical Review Letters},
	author = {Li, Shao-Xiong and Yu, Yang and Zhang, Yu and Qiu, Wei and Han, Siyuan and Wang, Zhen},
	month = aug,
	year = {2002},
	note = {Publisher: American Physical Society},
	pages = {098301},
}

@article{massarotti2012escape,
	title = {Escape dynamics in moderately damped {Josephson} junctions ({Review} {Article})},
	volume = {38},
	issn = {1063-777X},
	url = {https://doi.org/10.1063/1.3699625},
	doi = {10.1063/1.3699625},
	number = {4},
	urldate = {2024-12-17},
	journal = {Low Temperature Physics},
	author = {Massarotti, D. and Longobardi, L. and Galletti, L. and Stornaiuolo, D. and Montemurro, D. and Pepe, G. and Rotoli, G. and Barone, A. and Tafuri, F.},
	month = apr,
	year = {2012},
	pages = {263--272},
}

@article{affleck1981quantum,
	title = {Quantum-{Statistical} {Metastability}},
	volume = {46},
	url = {https://link.aps.org/doi/10.1103/PhysRevLett.46.388},
	doi = {10.1103/PhysRevLett.46.388},
	number = {6},
	urldate = {2024-12-17},
	journal = {Physical Review Letters},
	author = {Affleck, Ian},
	month = feb,
	year = {1981},
	note = {Publisher: American Physical Society},
	pages = {388--391},
}

@article{martinis1987experimental,
	title = {Experimental tests for the quantum behavior of a macroscopic degree of freedom: {The} phase difference across a {Josephson} junction},
	volume = {35},
	shorttitle = {Experimental tests for the quantum behavior of a macroscopic degree of freedom},
	url = {https://link.aps.org/doi/10.1103/PhysRevB.35.4682},
	doi = {10.1103/PhysRevB.35.4682},
	number = {10},
	urldate = {2025-01-28},
	journal = {Physical Review B},
	author = {Martinis, John M. and Devoret, Michel H. and Clarke, John},
	month = apr,
	year = {1987},
	note = {Publisher: American Physical Society},
	pages = {4682--4698},
}

@article{anferov2024improved,
	title = {Improved coherence in optically defined niobium trilayer-junction qubits},
	volume = {21},
	url = {https://link.aps.org/doi/10.1103/PhysRevApplied.21.024047},
	doi = {10.1103/PhysRevApplied.21.024047},
	number = {2},
	urldate = {2024-11-27},
	journal = {Physical Review Applied},
	author = {Anferov, Alexander and Lee, Kan-Heng and Zhao, Fang and Simon, Jonathan and Schuster, David I.},
	month = feb,
	year = {2024},
	note = {Publisher: American Physical Society},
	pages = {024047},
}

@article{han1989thermal,
	title = {Thermal activation in a two-dimensional potential},
	volume = {63},
	url = {https://link.aps.org/doi/10.1103/PhysRevLett.63.1712},
	doi = {10.1103/PhysRevLett.63.1712},
	number = {16},
	urldate = {2024-10-24},
	journal = {Physical Review Letters},
	author = {Han, S. and Lapointe, J. and Lukens, J. E.},
	month = oct,
	year = {1989},
	note = {Publisher: American Physical Society},
	pages = {1712--1715},
}

@article{dai2021calibration,
	title = {Calibration of {Flux} {Crosstalk} in {Large}-{Scale} {Flux}-{Tunable} {Superconducting} {Quantum} {Circuits}},
	volume = {2},
	url = {https://link.aps.org/doi/10.1103/PRXQuantum.2.040313},
	doi = {10.1103/PRXQuantum.2.040313},
	number = {4},
	urldate = {2024-12-03},
	journal = {PRX Quantum},
	author = {Dai, X. and Tennant, D.M. and Trappen, R. and Martinez, A.J. and Melanson, D. and Yurtalan, M.A. and Tang, Y. and Novikov, S. and Grover, J.A. and Disseler, S.M. and Basham, J.I. and Das, R. and Kim, D.K. and Melville, A.J. and Niedzielski, B.M. and Weber, S.J. and Yoder, J.L. and Lidar, D.A. and Lupascu, A.},
	month = oct,
	year = {2021},
	note = {Publisher: American Physical Society},
	pages = {040313},
}

@article{vandamme2024advanced,
	title = {Advanced {CMOS} manufacturing of superconducting qubits on 300 mm wafers},
	volume = {634},
	copyright = {2024 The Author(s)},
	issn = {1476-4687},
	url = {https://www.nature.com/articles/s41586-024-07941-9},
	doi = {10.1038/s41586-024-07941-9},
	language = {english},
	number = {8032},
	urldate = {2024-11-03},
	journal = {Nature},
	author = {Van Damme, J. and Massar, S. and Acharya, R. and Ivanov, Ts and Perez Lozano, D. and Canvel, Y. and Demarets, M. and Vangoidsenhoven, D. and Hermans, Y. and Lai, J. G. and Vadiraj, A. M. and Mongillo, M. and Wan, D. and De Boeck, J. and Potočnik, A. and De Greve, K.},
	month = oct,
	year = {2024},
	note = {Publisher: Nature Publishing Group},
	pages = {74--79},
}

@article{rosenberg20173d,
	title = {{3D} integrated superconducting qubits},
	volume = {3},
	copyright = {2017 The Author(s)},
	issn = {2056-6387},
	url = {https://www.nature.com/articles/s41534-017-0044-0},
	doi = {10.1038/s41534-017-0044-0},
	language = {english},
	number = {1},
	urldate = {2024-11-04},
	journal = {npj Quantum Information},
	author = {Rosenberg, D. and Kim, D. and Das, R. and Yost, D. and Gustavsson, S. and Hover, D. and Krantz, P. and Melville, A. and Racz, L. and Samach, G. O. and Weber, S. J. and Yan, F. and Yoder, J. L. and Kerman, A. J. and Oliver, W. D.},
	month = oct,
	year = {2017},
	note = {Publisher: Nature Publishing Group},
	pages = {1--5},
}

@article{yost2020solid,
	title = {Solid-state qubits integrated with superconducting through-silicon vias},
	volume = {6},
	copyright = {2020 The Author(s)},
	issn = {2056-6387},
	url = {https://www.nature.com/articles/s41534-020-00289-8},
	doi = {10.1038/s41534-020-00289-8},
	language = {english},
	number = {1},
	urldate = {2024-11-03},
	journal = {npj Quantum Information},
	author = {Yost, D. R. W. and Schwartz, M. E. and Mallek, J. and Rosenberg, D. and Stull, C. and Yoder, J. L. and Calusine, G. and Cook, M. and Das, R. and Day, A. L. and Golden, E. B. and Kim, D. K. and Melville, A. and Niedzielski, B. M. and Woods, W. and Kerman, A. J. and Oliver, W. D.},
	month = jul,
	year = {2020},
	note = {Publisher: Nature Publishing Group},
	pages = {1--7},
}

@misc{mallek2021fabrication,
	title = {Fabrication of superconducting through-silicon vias},
	url = {http://arxiv.org/abs/2103.08536},
	doi = {10.48550/arXiv.2103.08536},
	urldate = {2024-11-04},
	publisher = {arXiv},
	author = {Mallek, Justin L. and Yost, Donna-Ruth W. and Rosenberg, Danna and Yoder, Jonilyn L. and Calusine, Gregory and Cook, Matt and Das, Rabindra and Day, Alexandra and Golden, Evan and Kim, David K. and Knecht, Jeffery and Niedzielski, Bethany M. and Schwartz, Mollie and Sevi, Arjan and Stull, Corey and Woods, Wayne and Kerman, Andrew J. and Oliver, William D.},
	month = mar,
	year = {2021},
	note = {arXiv:2103.08536},
}

@misc{vahidpour2017superconducting,
	title = {Superconducting {Through}-{Silicon} {Vias} for {Quantum} {Integrated} {Circuits}},
	url = {http://arxiv.org/abs/1708.02226},
	urldate = {2024-11-04},
	publisher = {arXiv},
	author = {Vahidpour, Mehrnoosh and O'Brien, William and Whyland, Jon Tyler and Angeles, Joel and Marshall, Jayss and Scarabelli, Diego and Crossman, Genya and Yadav, Kamal and Mohan, Yuvraj and Bui, Catvu and Rawat, Vijay and Renzas, Russ and Vodrahalli, Nagesh and Bestwick, Andrew and Rigetti, Chad},
	month = aug,
	year = {2017},
	note = {arXiv:1708.02226},
}

@article{acharya2023multiplexed,
	title = {Multiplexed superconducting qubit control at millikelvin temperatures with a low-power cryo-{CMOS} multiplexer},
	volume = {6},
	copyright = {2023 The Author(s), under exclusive licence to Springer Nature Limited},
	issn = {2520-1131},
	url = {https://www.nature.com/articles/s41928-023-01033-8},
	doi = {10.1038/s41928-023-01033-8},
	language = {en},
	number = {11},
	urldate = {2025-01-20},
	journal = {Nature Electronics},
	author = {Acharya, R. and Brebels, S. and Grill, A. and Verjauw, J. and Ivanov, Ts and Lozano, D. Perez and Wan, D. and Van Damme, J. and Vadiraj, A. M. and Mongillo, M. and Govoreanu, B. and Craninckx, J. and Radu, I. P. and De Greve, K. and Gielen, G. and Catthoor, F. and Potočnik, A.},
	month = nov,
	year = {2023},
	note = {Publisher: Nature Publishing Group},
	pages = {900--909},
}

@misc{gupta2024low,
	title = {Low loss lumped-element inductors made from granular aluminum},
	url = {http://arxiv.org/abs/2411.12611},
	doi = {10.48550/arXiv.2411.12611},
	urldate = {2024-12-06},
	publisher = {arXiv},
	author = {Gupta, Vishakha and Winkel, Patrick and Thakur, Neel and Vlaanderen, Peter van and Wang, Yanhao and Ganjam, Suhas and Frunzio, Luigi and Schoelkopf, Robert J.},
	month = nov,
	year = {2024},
	note = {arXiv:2411.12611 [quant-ph]},
}

@article{strandberg2024digital,
	title = {Digital {Homodyne} and {Heterodyne} {Detection} for {Stationary} {Bosonic} {Modes}},
	volume = {133},
	url = {https://link.aps.org/doi/10.1103/PhysRevLett.133.063601},
	doi = {10.1103/PhysRevLett.133.063601},
	number = {6},
	urldate = {2025-01-20},
	journal = {Physical Review Letters},
	author = {Strandberg, Ingrid and Eriksson, Axel M. and Royer, Baptiste and Kervinen, Mikael and Gasparinetti, Simone},
	month = aug,
	year = {2024},
	note = {Publisher: American Physical Society},
	pages = {063601},
}

@article{potts2001novel,
	title = {Novel fabrication methods for submicrometer {Josephson} junction qubits},
	volume = {12},
	issn = {1573-482X},
	url = {https://doi.org/10.1023/A:1011279908265},
	doi = {10.1023/A:1011279908265},
	language = {english},
	number = {4},
	urldate = {2024-11-11},
	journal = {Journal of Materials Science: Materials in Electronics},
	author = {Potts, A. and Routley, P. R. and Parker, G. J. and Baumberg, J. J. and de Groot, P. A. J.},
	month = jun,
	year = {2001},
	pages = {289--293},
}

@misc{muthusubramanian2023wafer,
	title = {Wafer-scale uniformity of {Dolan}-bridge and bridgeless {Manhattan}-style {Josephson} junctions for superconducting quantum processors},
	url = {http://arxiv.org/abs/2304.09111},
	doi = {10.48550/arXiv.2304.09111},
	urldate = {2024-11-11},
	publisher = {arXiv},
	author = {Muthusubramanian, N. and Duivestein, P. and Zachariadis, C. and Finkel, M. and Meer, S. L. M. van der and Veen, H. M. and Beekman, M. W. and Stavenga, T. and Bruno, A. and DiCarlo, L.},
	month = apr,
	year = {2023},
	note = {arXiv:2304.09111},
}

@inproceedings{gangloff2021general,
  title={A general parametrization framework for pairwise Markov models: An application to unsupervised image segmentation},
  author={Gangloff, Hugo and Morales, Katherine and Petetin, Yohan},
  booktitle={2021 IEEE 31st International Workshop on Machine Learning for Signal Processing (MLSP)},
  pages={1--6},
  year={2021},
  organization={IEEE}
}

@article{gangloff2023deep,
  title={Deep parameterizations of pairwise and triplet Markov models for unsupervised classification of sequential data},
  author={Gangloff, Hugo and Morales, Katherine and Petetin, Yohan},
  journal={Computational Statistics \& Data Analysis},
  volume={180},
  pages={107663},
  year={2023},
  publisher={Elsevier}
}

@article{yedidia2000generalized,
  title={Generalized belief propagation},
  author={Yedidia, Jonathan S and Freeman, William and Weiss, Yair},
  journal={Advances in neural information processing systems},
  volume={13},
  year={2000}
}

@article{kumar2025evaluation,
  title={Evaluation of fluxon synapse device based on superconducting loops for energy efficient neuromorphic computing},
  author={Kumar, Ashwani and Goteti, Uday S and Cubukcu, Ertugrul and Dynes, Robert C and Kuzum, Duygu},
  journal={Frontiers in Neuroscience},
  volume={19},
  pages={1511371},
  year={2025},
  publisher={Frontiers Media SA}
}

@inproceedings{zhu2025transformers,
  title={Transformers without normalization},
  author={Zhu, Jiachen and Chen, Xinlei and He, Kaiming and LeCun, Yann and Liu, Zhuang},
  booktitle={Proceedings of the Computer Vision and Pattern Recognition Conference},
  pages={14901--14911},
  year={2025}
}

@inproceedings{reilly2025physical,
  title={Physical complexity and black hole quantum computers},
  author={Reilly, Michele and Lloyd, Seth},
  booktitle={Journal of Physics: Conference Series},
  volume={3017},
  pages={012010},
  year={2025},
  organization={IOP Publishing}
}

@article{klinger2025minimally,
  title={Minimally dissipative multi-bit logical operations},
  author={Klinger, J{\'e}r{\'e}mie and Rotskoff, Grant M},
  journal={arXiv preprint arXiv:2506.24021},
  year={2025}
}

@article{lopez2023self,
  title={Self-learning machines based on Hamiltonian echo backpropagation},
  author={Lopez-Pastor, Victor and Marquardt, Florian},
  journal={Physical Review X},
  volume={13},
  number={3},
  pages={031020},
  year={2023},
  publisher={APS}
}

@article{bosch2025local,
  title={Local Learning Rules for Out-of-Equilibrium Physical Generative Models},
  author={B{\"o}sch, Cyrill and Roeder, Geoffrey and Serra-Garcia, Marc and Adams, Ryan P},
  journal={arXiv preprint arXiv:2506.19136},
  year={2025}
}

@article{whitelam2025generative,
  title={Generative thermodynamic computing},
  author={Whitelam, Stephen},
  journal={arXiv preprint arXiv:2506.15121},
  year={2025}
}

@article{sajeeb2025scalable,
  title={Scalable connectivity for Ising machines: Dense to sparse},
  author={Sajeeb, M Mahmudul Hasan and Aadit, Navid Anjum and Chowdhury, Shuvro and Wu, Tong and Smith, Cesely and Chinmay, Dhruv and Raut, Atharva and Camsari, Kerem Y and Delacour, Corentin and Srimani, Tathagata},
  journal={Physical Review Applied},
  volume={24},
  number={1},
  pages={014005},
  year={2025},
  publisher={APS}
}

@inproceedings{choi2014algorithmic,
  title={Algorithmic time, energy, and power on candidate HPC compute building blocks},
  author={Choi, Jee and Dukhan, Marat and Liu, Xing and Vuduc, Richard},
  booktitle={2014 IEEE 28th international parallel and distributed processing symposium},
  pages={447--457},
  year={2014},
  organization={IEEE}
}

@article{garcia2019estimation,
  title={Estimation of energy consumption in machine learning},
  author={Garc{\'\i}a-Mart{\'\i}n, Eva and Rodrigues, Crefeda Faviola and Riley, Graham and Grahn, H{\aa}kan},
  journal={Journal of Parallel and Distributed Computing},
  volume={134},
  pages={75--88},
  year={2019},
  publisher={Elsevier}
}

@inproceedings{shankar2023energy,
  title={Energy estimates across layers of computing: from devices to large-scale applications in machine learning for natural language processing, scientific computing, and cryptocurrency mining},
  author={Shankar, Sadasivan},
  booktitle={2023 IEEE High Performance Extreme Computing Conference (HPEC)},
  pages={1--6},
  year={2023},
  organization={IEEE}
}

@article{radford2019language,
  title={Language models are unsupervised multitask learners},
  author={Radford, Alec and Wu, Jeffrey and Child, Rewon and Luan, David and Amodei, Dario and Sutskever, Ilya and others},
  journal={OpenAI blog},
  volume={1},
  number={8},
  pages={9},
  year={2019}
}

@article{freitas2021stochastic,
  title={Stochastic thermodynamics of nonlinear electronic circuits: A realistic framework for computing around k T},
  author={Freitas, Nahuel and Delvenne, Jean-Charles and Esposito, Massimiliano},
  journal={Physical Review X},
  volume={11},
  number={3},
  pages={031064},
  year={2021},
  publisher={APS}
}

@article{foster2023high,
    title={High order splitting methods for SDEs satisfying a commutativity
           condition},
    author={James Foster and Goncalo dos Reis and Calum Strange},
    year={2023},
    journal={arXiv:2210.17543},
}

@article{yang2025250,
  title={250 Magnetic Tunnel Junctions-Based Probabilistic Ising Machine},
  author={Yang, Shuhan and Grimaldi, Andrea and Bao, Youwei and Raimondo, Eleonora and Si, Jia and Finocchio, Giovanni and Yang, Hyunsoo},
  journal={arXiv preprint arXiv:2506.14590},
  year={2025}
}

@article{camsari2017implementing,
  title={Implementing p-bits with embedded MTJ},
  author={Camsari, Kerem Yunus and Salahuddin, Sayeef and Datta, Supriyo},
  journal={IEEE Electron Device Letters},
  volume={38},
  number={12},
  pages={1767--1770},
  year={2017},
  publisher={IEEE}
}

@article{rhee2023probabilistic,
  title={Probabilistic computing with NbOx metal-insulator transition-based self-oscillatory pbit},
  author={Rhee, Hakseung and Kim, Gwangmin and Song, Hanchan and Park, Woojoon and Kim, Do Hoon and In, Jae Hyun and Lee, Younghyun and Kim, Kyung Min},
  journal={Nature communications},
  volume={14},
  number={1},
  pages={7199},
  year={2023},
  publisher={Nature Publishing Group UK London}
}

@article{shalf2020future,
  title={The future of computing beyond Moore’s Law},
  author={Shalf, John},
  journal={Philosophical Transactions of the Royal Society A},
  volume={378},
  number={2166},
  pages={20190061},
  year={2020},
  publisher={The Royal Society Publishing}
}

@misc{chen2025strongernormalizationfreetransformers,
      title={Stronger Normalization-Free Transformers}, 
      author={Mingzhi Chen and Taiming Lu and Jiachen Zhu and Mingjie Sun and Zhuang Liu},
      year={2025},
      eprint={2512.10938},
      archivePrefix={arXiv},
      primaryClass={cs.LG},
      url={https://arxiv.org/abs/2512.10938}, 
}

@article{jelinvcivc2025efficient,
  title={An efficient probabilistic hardware architecture for diffusion-like models},
  author={Jelin{\v{c}}i{\v{c}}, Andra{\v{z}} and Lockwood, Owen and Garlapati, Akhil and Schillinger, Peter and Chuang, Isaac L and Verdon, Guillaume and McCourt, Trevor},
  journal={npj Unconventional Computing},
  volume={3},
  number={1},
  pages={30},
  year={2026},
  publisher={Nature Publishing Group UK London}
}

@article{aimone2025neuromorphic,
  title={Neuromorphic Computing: A Theoretical Framework for Time, Space, and Energy Scaling},
  author={Aimone, James B},
  journal={arXiv preprint arXiv:2507.17886},
  year={2025}
}

@article{whitelam2025training,
  title={Training thermodynamic computers by gradient descent},
  author={Whitelam, Stephen},
  journal={arXiv preprint arXiv:2509.15324},
  year={2025}
}

@article{hnybida2025minimal,
  title={Minimal-Dissipation Learning for Energy-Based Models},
  author={Hnybida, Jeff and Verret, Simon},
  journal={arXiv preprint arXiv:2510.03137},
  year={2025}
}

@article{lloyd2025thermodynamics+,
  title={Thermodynamics+ Natural Selection= Bayesian Inference},
  author={Lloyd, Seth},
  journal={arXiv preprint arXiv:2511.17641},
  year={2025}
}

@inproceedings{borle2018analyzing,
  title={Analyzing the quantum annealing approach for solving linear least squares problems},
  author={Borle, Ajinkya and Lomonaco, Samuel J},
  booktitle={International Workshop on Algorithms and Computation},
  pages={289--301},
  year={2018},
  organization={Springer}
}

@article{hooker2021hardware,
  title={The hardware lottery},
  author={Hooker, Sara},
  journal={Communications of the ACM},
  volume={64},
  number={12},
  pages={58--65},
  year={2021},
  publisher={ACM New York, NY, USA}
}

@article{fedus2022switch,
  title={Switch transformers: Scaling to trillion parameter models with simple and efficient sparsity},
  author={Fedus, William and Zoph, Barret and Shazeer, Noam},
  journal={Journal of Machine Learning Research},
  volume={23},
  number={120},
  pages={1--39},
  year={2022}
}

@article{shazeer2017outrageously,
  title={Outrageously large neural networks: The sparsely-gated mixture-of-experts layer},
  author={Shazeer, Noam and Mirhoseini, Azalia and Maziarz, Krzysztof and Davis, Andy and Le, Quoc and Hinton, Geoffrey and Dean, Jeff},
  journal={arXiv preprint arXiv:1701.06538},
  year={2017}
}

@article{rolandi2026energy,
  title={Energy-Time-Accuracy Tradeoffs in Thermodynamic Computing},
  author={Rolandi, Alberto and Abiuso, Paolo and Lipka-Bartosik, Patryk and Aifer, Maxwell and Coles, Patrick J and Perarnau-Llobet, Mart{\'\i}},
  journal={arXiv preprint arXiv:2601.04358},
  year={2026}
}

@misc{Freitas2026NEATRN,
  title={Taming nonequilibrium thermal fluctuations in subthreshold CMOS circuits},
  author={Freitas, Nahuel and Massarelli, Geremia and Rothschild, Jeremy and Keane, Dylan and Dawe, Ethan and Hwang, Sewook and Garlapati, Akhil and McCourt, Trevor},
  journal={Physical Review Applied},
  volume={25},
  number={3},
  pages={034061},
  year={2026},
  publisher={APS}
}

@article{scellier2017equilibrium,
  title={Equilibrium propagation: Bridging the gap between energy-based models and backpropagation},
  author={Scellier, Benjamin and Bengio, Yoshua},
  journal={Frontiers in computational neuroscience},
  volume={11},
  pages={24},
  year={2017},
  publisher={Frontiers Media SA}
}

@article{kendall2020training,
  title={Training end-to-end analog neural networks with equilibrium propagation},
  author={Kendall, Jack and Pantone, Ross and Manickavasagam, Kalpana and Bengio, Yoshua and Scellier, Benjamin},
  journal={arXiv preprint arXiv:2006.01981},
  year={2020}
}

@article{stern2021supervised,
  title={Supervised learning in physical networks: From machine learning to learning machines},
  author={Stern, Menachem and Hexner, Daniel and Rocks, Jason W and Liu, Andrea J},
  journal={Physical Review X},
  volume={11},
  number={2},
  pages={021045},
  year={2021},
  publisher={APS}
}

@article{de2021materials,
  title={Materials challenges and opportunities for quantum computing hardware},
  author={De Leon, Nathalie P and Itoh, Kohei M and Kim, Dohun and Mehta, Karan K and Northup, Tracy E and Paik, Hanhee and Palmer, BS and Samarth, Nitin and Sangtawesin, Sorawis and Steuerman, David W},
  journal={Science},
  volume={372},
  number={6539},
  pages={eabb2823},
  year={2021},
  publisher={American Association for the Advancement of Science}
}

@article{wendin2017quantum,
  title={Quantum information processing with superconducting circuits: a review},
  author={Wendin, G{\"o}ran},
  journal={Reports on Progress in Physics},
  volume={80},
  number={10},
  pages={106001},
  year={2017},
  publisher={IOP Publishing}
}

@article{leroux2025analog,
  title={Analog in-memory computing attention mechanism for fast and energy-efficient large language models},
  author={Leroux, Nathan and Manea, Paul-Philipp and Sudarshan, Chirag and Finkbeiner, Jan and Siegel, Sebastian and Strachan, John Paul and Neftci, Emre},
  journal={Nature Computational Science},
  volume={5},
  number={9},
  pages={813--824},
  year={2025},
  publisher={Nature Publishing Group US New York}
}

@article{massar2025equilibrium,
  title={Equilibrium propagation: the quantum and the thermal cases},
  author={Massar, Serge and Mognetti, Bortolo Matteo},
  journal={Quantum Studies: Mathematics and Foundations},
  volume={12},
  number={1},
  pages={6},
  year={2025},
  publisher={Springer}
}

@misc{gen_1_5_USPTO,
  author = {Chamberland, Christopher and Verdon-Akzam, Guillaume},
  title = {Thermodynamic computing system configured to use natural gradient descent techniques to determine updated weights and biases},
  howpublished = {US Patent Application Publication US 2025/0238670 A1},
  year = {2025},
  note = {
      {Assignee}:\ Extropic Corp.
      Status:\ pending.
  },
  url = {https://patents.google.com/patent/US20250238670A1/en?oq=US+2025%2f0238670+A1},
}

@misc{gen_1_75_USPTO,
  author = {Chamberland, Christopher and Verdon-Akzam, Guillaume},
  title = {Thermodynamic computing system configured to update weights and biases based on gradient values obtained by relay oscillators},
  howpublished = {US Patent Application Publication US 2025/0390737 A1},
  year = {2025},
  month = {12},
  note = {
      {Assignee}:\ Extropic Corp.
      Status:\ pending.
  },
  url = {https://patents.google.com/patent/US20250390737A1/en?oq=US+2025%2f0390737+A1},
}

@misc{gen_2_USPTO,
  author = {Chamberland, Christopher and Verdon-Akzam, Guillaume},
  title = {Self-learning thermodynamic computing system},
  howpublished = {US Patent Application Publication US 2025/0165761 A1},
  year = {2025},
  month = {05},
  note = {
      {Assignee}:\ Extropic Corp.
      Status:\ pending.
  },
  url = {https://patents.google.com/patent/US20250165761A1/en?oq=US+2025%2f0165761+A1},
}

@misc{gibbs_sampling_USPTO,
  author = {Chamberland, Christopher and Verdon-Akzam, Guillaume},
  title = {Gibbs sampling methods using thermodynamic computing},
  howpublished = {US Patent Application Publication US 2025/0284562 A1},
  year = {2025},
  month = {09},
  note = {
      {Assignee}:\ Extropic Corp.
      Status:\ pending.
  },
  url = {https://patents.google.com/patent/US20250284562A1/en?oq=US-2025-0284562-A1},
}

@misc{mixture_of_experts_USPTO,
  author = {Chamberland, Christopher and Verdon-Akzam, Guillaume},
  title = {Mixture of experts energy based model gadget},
  howpublished = {US Patent Application Publication US 2025/0284998 A1},
  year = {2025},
  month = {09},
  note = {
      {Assignee}:\ Extropic Corp.
      Status:\ pending.
  },
  url = {https://patents.google.com/patent/US20250284998A1/en?oq=US-2025-0284998-A1},
}

@misc{relay_gadget_analo_USPTO,
  author = {Chamberland, Christopher and Verdon-Akzam, Guillaume},
  title = {Thermodynamic computing relay gadget},
  howpublished = {US Patent Application Publication US 2025/0284867 A1},
  year = {2025},
  month = {09},
  note = {
      {Assignee}:\ Extropic Corp.
      Status:\ pending.
  },
  url = {https://patents.google.com/patent/US20250284867A1/en?oq=US+2025%2f0284867+A1},
}

@misc{relay_gadget_multi_well_USPTO,
  author = {Chamberland, Christopher and Verdon-Akzam, Guillaume},
  title = {Thermodynamic computing relay gadget for multi-well potentials},
  howpublished = {US Patent Application Publication US 2025/0373202 A1},
  year = {2025},
  month = {12},
  note = {
      {Assignee}:\ Extropic Corp.
      Status:\ pending.
  },
  url = {https://patents.google.com/patent/US20250373202A1/en?oq=US+2025%2f0373202+A1},
}

@misc{mean_field_USPTO,
  author = {Chamberland, Christopher and Verdon-Akzam, Guillaume},
  title = {Thermodynamic computing mean-field forwards and backwards propagation},
  howpublished = {US Patent Application Publication US 2025/0284959 A1},
  year = {2025},
  month = {09},
  note = {
      {Assignee}:\ Extropic Corp.
      Status:\ pending.
  },
  url = {https://patents.google.com/patent/US20250284959A1/en?oq=US+2025%2f0284959+A1},
}

@misc{mean_field_PTC,
  author = {Chamberland, Christopher and Verdon-Akzam, Guillaume},
  title = {Thermodynamic computing mean-field forwards and backwards propagation},
  howpublished = {PCT Application Publication WO 2025/189010 A8},
  year = {2025},
  month = {09},
  note = {
      {Assignee}:\ Extropic Corp.
      Status:\ pending.
  },
  url = {https://patents.google.com/patent/WO2025189010A8/en?oq=WO+2025%2f189010+A8},
}

@misc{selection_of_experts_USPTO,
  author = {Chamberland, Christopher and Verdon-Akzam, Guillaume},
  title = {Selection of experts energy based model gadget},
  howpublished = {US Patent Application Publication US 2025/0284999 A1},
  year = {2025},
  month = {09},
  note = {
      {Assignee}:\ Extropic Corp.
      Status:\ pending.
  },
  url = {https://patents.google.com/patent/US20250284999A1/en?oq=US+2025%2f0284999+A1},
}

@misc{swish_gadget_USPTO,
  author       = {Chamberland, Christopher and Verdon-Akzam, Guillaume},
  title        = {Thermodynamic computing swish gadget},
  howpublished = {US Patent Application US 18/937,670},
  year         = {2024},
  month = {11},
  note         = {
      {Assignee}:\ Extropic Corp. 
      Filed: Nov.\ 5, 2024.
      Status: pending; yet to be published.
  },
}

@misc{thermo_transformer_USPTO,
  author = {Chamberland, Christopher and Verdon-Akzam, Guillaume},
  title = {Thermodynamic computing system configured to implement transformer based architecture},
  howpublished = {US Patent Application Publication US 2025/0284949 A1},
  year = {2025},
  month = {09},
  note = {
      {Assignee}:\ Extropic Corp.
      Status:\ pending.
  },
  url = {https://patents.google.com/patent/US20250284949A1/en?oq=US+2025%2f0284949+A1},
}

@misc{sc_neuron_USPTO,
  author = {Chamberland, Christopher and Verdon-Akzam, Guillaume},
  title = {Superconducting thermodynamic neuron},
  howpublished = {US Patent Application Publication US 2025/0284924 A1},
  year = {2025},
  month = {09},
  note = {
      {Assignee}:\ Extropic Corp.
      Status:\ pending.
  },
  url = {https://patents.google.com/patent/US20250284924A1/en?oq=US+2025%2f0284924+A1},
}

\appendix

\clearpage

\onecolumngrid

\section{Dimensionless Langevin equations} \label{app:dimensionless}

This appendix records a convenient nondimensionalization that connects the standard dimensionful underdamped Langevin equation used in this work (cf.~Eq.~\eqref{eq:underdamped_langevin}) to the dimensionless form used in Sec.~\ref{sec:selflearning}.

Consider a set of coordinates $\tilde{x}_i(\tilde{t})$ with conjugate momenta $\tilde{p}_i(\tilde{t})$ evolving in a potential energy $\tilde{U}(\tilde{x})$ at inverse temperature $\beta=(k_B T)^{-1}$. The dimensionful SDEs may be written as
\begin{align}
    \begin{split}
    d\tilde{x}_i &= \frac{\tilde{p}_i}{m_i} d\tilde{t}\\
    d\tilde{p}_i &= -\Bigl(\partial_{\tilde{x}_i}\tilde{U}(\tilde{x}) + \frac{\tilde{\gamma}_i}{m_i}\tilde{p}_i\Bigr)d\tilde{t}
    + \sqrt{\frac{2\tilde{\gamma}_i}{\beta}}\,d\tilde{W}_i(\tilde{t}),
    \end{split}
\end{align}
with independent Wiener processes satisfying $\langle d\tilde{W}_i\,d\tilde{W}_j\rangle=\delta_{ij}\,d\tilde{t}$.

To convert to dimensionless, define the thermal velocity scale $u_i\coloneq \sqrt{k_B T/m_i} = \sqrt{1/(\beta m_i)}$ and pick an arbitrary inverse-time scale $\Gamma$. We introduce dimensionless variables
\begin{equation}
\begin{aligned}
    t &\coloneq \Gamma \tilde{t},
    \qquad
    x_i \coloneq \Gamma \frac{\tilde{x}_i}{u_i},
    \qquad
    p_i \coloneq \frac{\tilde{p}_i}{m_i u_i},
    \\
    \zeta_i &\coloneq \frac{\tilde{\gamma}_i}{m_i\Gamma},
    \qquad
    U(x) \coloneq \beta\,\tilde{U}(\tilde{x}),
    \qquad
    W_i(t) \coloneq \sqrt{\Gamma} \tilde{W}_i(t/\Gamma).
\end{aligned}
\end{equation}
Equivalently,
\begin{equation}
    dW_i(t) = \xi_i(t) dt,
\end{equation}
the independent Gaussian white noises satisfy
\begin{equation}
    \mathbbm{E}[\xi_i(t)] = 0, \qquad
    \mathbbm{E} \left[\xi_i(t) \xi_j(t^\prime) \right] = \delta_{ij} \delta(t-t^\prime).
\end{equation}
Then the SDEs become
\begin{align}
    \begin{split}
    d x_i &= p_i dt  \\
    d p_i &= \Bigl[-\partial_{x_i}U(x) - \zeta_i p_i\Bigr]dt + \sqrt{2\zeta_i}\xi_i(t) dt.
    \end{split}
\end{align}
This is the prototype form used in Sec.~\ref{sec:selflearning}. The same construction applies componentwise when enlarging the state space to include additional (latent or parameter) variables.

\section{Further subleading-order terms in the kinematics} \label{app:full_expressions}

In the following, we assume that the initial velocity vanishes, which is a sensible assumption for this driving protocol; a more general scenario is considered later.
Let us define the following stochastic observable:
\begin{equation}\label{eq:theta_observable_definition}
    \Phi^{(\theta)}_i(t) \coloneq \frac{2\bigl(\theta_i(t) - \theta^0_i\bigr)}{t^2}.
\end{equation}
To leading order in time, its expectation value and variance go as
\begin{subequations}
\begin{align}\label{eq:theta_observable_mean_var}
\left\langle \Phi^{(\theta)}_i(t) \right\rangle &= F^0_i - \frac{t}{3} \zeta_i F^0_i + \cO(t^2),
\\
\text{var}\Bigl(\Phi^{(\theta)}_i(t)\Bigr) &= \frac{8\zeta_i}{3t} - 2 \zeta_i^2 + \cO(t).
\end{align}
\end{subequations}

We see that as \(t \rightarrow 0\), \(\langle \Phi^{(\theta)}_i(t) \rangle\) tends to \(F^0_i\).
Naïvely, this suggests that repeated observations of \(\Phi^{(\theta)}_i(t)\) at the earliest times will yield the best estimates of \(F^0_i\), so that the subleading-order term remains small.
However, the standard deviation grows as \(t^{-1/2}\) at early times, and the number of repeated measurements needed to capture the mean with some fixed uncertainty grows as \(N \propto t^{-1}\) for small \(t\).
Hence, in practice, a tradeoff must be struck between earlier times at which statistical errors dominate, and later times at which systematic errors may become sizeable.

There are other stochastic observables that can play the same role as \(\Phi^{(\theta)}_i(t)\). In the following, we also define \(\Phi^{(p)}_i(t)\), which involves measurements of \(p_i\) rather than \(\theta_i\), and \(\Phi^{(\theta p)}_i(t)\), which involves measurements of both, but has the advantage of converging faster to \(F^0_i\) for small \(t\).

The effective equations of motion for \(\theta\) are Eq.~\eqref{eq:langevin-nonlinear}.
Whereas in the text, we assumed that the different Gaussian white noise channels were uncorrelated, here we allow correlation:
\begin{equation}
    \mathbbm{E} [ \xi_i(t) \xi_j(t') ] = q_{ij} \delta(t - t').
\end{equation}
The early-time solution for the means is
\begin{subequations}
\begin{multline}
    \mathbbm{E} [ \theta_i(t) ] = \theta^0_i + tp_i^0 + \frac{t^2}{2} \left( F^0_i - \zeta_i p_i^0 \right) + \frac{t^3}{3!} \left( \sum_j J^0_{ij} p^0_j - \zeta_i F^0_i + \zeta_i^2 p_i^0 \right)
    \\
    + \frac{t^4}{4!} \left( \zeta_i^2 (F^0_i - \zeta_i p_i^0) + \sum_j J^0_{ij} \bigl(F^0_j - \zeta_j p^0_j - \zeta_i p^0_j \bigr) \right)
    + \cO(t^5),
\end{multline}
\begin{multline}
    \mathbbm{E}[ p_i(t) ] = p_i^0 + t \left( F^0_i - \zeta_i p_i^0 \right) + \frac{t^2}{2} \left( \sum_j J^0_{ij} p^0_j  - \zeta_i F^0_i + \zeta_i^2 p_i^0 \right)   
    \\
    + \frac{t^3}{3!} \left( \zeta_i^2 (F^0_i - \zeta_i p_i^0) + \sum_j J^0_{ij} \bigl(F^0_j - \zeta_j p^0_j - \zeta_i p^0_j \bigr) \right)
    + \cO(t^4),
\end{multline}
\end{subequations}

while for the variances we have
\begin{subequations}
\begin{gather}
    \text{var}\bigl(\theta_i(t)\bigr) = \frac{2}{3} q_{ii} \zeta_i t^3 - \frac{1}{2} q_{ii} \zeta_i^2 t^4 + \cO(t^5),
    \\
    \text{var}\bigl(p_i(t)\bigr) = 2 q_{ii} \zeta_i t - 2 q_{ii} \zeta_i^2 t^2 + \cO(t^3).
\end{gather}
\end{subequations}

As mentioned in the main text, we consider two additional stochastic observables:
\begin{subequations} \label{eq:observables_definition}
    \begin{align}
        \Phi^{(\theta)}_i(t) &\coloneq \frac{2}{t} \Biggl( \frac{\theta_i(t) - \theta^0_i}{t} - p_i^0 \Biggr) + \zeta_i p_i^0,
        \\
        \Phi^{(p)}_i(t) &\coloneq \vphantom{\Biggl(} \frac{p_i(t) - p_i^0}{t} + \zeta_i p_i^0 \vphantom{\Biggr)},
        \\
        \Phi^{(\theta p)}_i(t) &\coloneq \frac{p_i(t) - p_i^0 + \zeta_i \bigl(\theta_i(t) - \theta^0_i\bigr)}{t}.
    \end{align}
\end{subequations}
Any of these could potentially be used to measure the force \(F^0\).
To sub-leading order, their early-time statistical behavior is given by
\begin{subequations} \label{eq:observables_mean}
    \begin{align}
        \mathbbm{E} \left [ \Phi^{(\theta)}_i(t) \right] &= F^0_i - \frac{t}{3} \left( \zeta_i F^0_i - \zeta_i^2 p_i^0 - \sum_j J^0_{ij} p^0_j \right)
        + \frac{t^2}{12} \left( \zeta_i^2 F^0_i - \zeta_i^3 p_i^0 + \sum_j J^0_{ij} \Bigl( F^0_j - (\zeta_i+\zeta_j) p^0_j \Bigr) \right)
        + \cO(t^3),
        \\
        \mathbbm{E} \left [ \Phi^{(p)}_i(t) \right] &= F^0_i - \frac{t}{2} \left( \zeta_i F^0_i - \zeta_i^2 p_i^0 - \sum_j J^0_{ij} p^0_j \right)
        + \frac{t^2}{3!} \left( \zeta_i^2 F^0_i - \zeta_i^3 p_i^0 + \sum_j J^0_{ij} \Bigl( F^0_j - (\zeta_i+\zeta_j) p^0_j \Bigr) \right)
        + \cO(t^3),
        \\
        \mathbbm{E} \left [ \Phi^{(\theta p)}_i(t) \right] &= F^0_i + \frac{t}{2} \sum_j J^0_{ij} p^0_j + \frac{t^2}{3!}\sum_j J^0_{ij} \left( F^0_j - \zeta_j p^0_j \right) + \cO(t^3),
    \end{align}
\end{subequations}
together with
\begin{subequations} \label{eq:observables_variance}
    \begin{align}
        \text{var}\Bigl(\Phi^{(\theta)}_i(t)\Bigr) &= \frac{8}{3} q_{ii} \zeta_i \frac{1}{t} - 2 q_{ii} \zeta_i^2 + \cO(t),
        \\
        \text{var}\Bigl(\Phi^{(p)}_i(t)\Bigr) &= 2 q_{ii} \zeta_i \frac{1}{t} - 2 q_{ii} \zeta_i^2 + \cO(t),
        \\
        \text{var}\Bigl(\Phi^{(\theta p)}_i(t)\Bigr) &= 2 q_{ii} \zeta_i \frac{1}{t} + \frac{2}{3} \ t \sum_j J^0_{ij} q_{ij} \sqrt{\zeta_i\zeta_j} + \cO(t^2).
    \end{align}
\end{subequations}

Here, we see the advantage of using \(\Phi^{(\theta p)} (t)\): presuming the initial condition satisfies \(p^0 = 0\), the subleading-order contributions to both \( \mathbbm{E} [\Phi^{(\theta p)} (t) ] \) and \( \text{var}\bigl(\Phi^{(\theta p)} (t)\bigr) \) vanish, making it easier to strike a balance between statistical noise and systematic errors.

\section{HMM derivations}
\label{ap:hmm}

In the following we denote a set of samples at the steps $1:T$ with the superscript $x^{(1:T)}$. As a reminder the probability density of an HMM is given by:
\begin{align}
    \pi(x^{(1:T)}, z^{(1:T)}) =
    \pi(z^{(1)}) \prod_{t=2}^T \pi(z^{(t)} | z^{(t-1)}) \prod_{t=1}^T \pi(x^{(t)} | z^{(t)}).
\end{align}
If we use EBMs to parameterize the transition and emission probabilities we get:
\begin{align}
\label{eq:HMM joint prob with EBMs}
\pi(x^{(1:T)}, z^{(1:T)}) = \frac{e^{-E_{\pi}(z^{(1)})}}{Z_{\pi}} \prod_{t=2}^T \frac{e^{-E_{\theta}(z^{(t)}, z^{(t-1)})}}{Z_{\theta}(z^{(t-1)})} \prod_{t=1}^T \frac{e^{-E_{\phi}(x^{(t)}, z^{(t)})}}{Z_{\phi}(z^{(t)})},
\end{align}
where the partition functions are:
\begin{align}
\begin{split}
\label{eq: partial partition functions HMM}
Z_{\theta}(z^{(t-1)}) &= \sum_{z^{(t)}} e^{-E_{\theta}(z^{(t)}, z^{(t-1)})} \\  
Z_{\phi}(z^{(t)}) &= \sum_{x^{(t)}} e^{-E_{\phi}(x^{(t)}, z^{(t)})} \\ 
Z_{\pi} &= \sum_{z^{(1)}} e^{-E_{\pi}(z^{(1)})}.
\end{split}
\end{align}
To avoid confusion, we briefly show on the example of $\pi(x^{(t)} \vert z^{(t)})$ how it can be rewritten in the form of Eq.~\eqref{eq:HMM joint prob with EBMs}. Basic probability states that:
\begin{align}
\label{eq: conditional p from joint and marginal}
    \pi(x^{(t)} \vert z^{(t)}) = \frac{\pi(x^{(t)}, z^{(t)})}{\pi(z^{(t)})} = \frac{\pi(x^{(t)}, z^{(t)})}{\sum_{x^{(t)}} \pi(x^{(t)}, z^{(t)})}.
\end{align}
The joint probability is parameterized by an EBM, hence:
\begin{align}
    \pi(x^{(t)}, z^{(t)}) = \frac{e^{-E_{\phi} (x^{(t)}, z^{(t)})}}{Z_{\phi}}.
\end{align}
The partition function without argument $Z_{\phi} = \sum_{x^{(t)}, z^{(t)}} e^{-E_{\phi} (x^{(t)}, z^{(t)})}$ is traced out over both variables.
Dividing $\pi(x^{(t)}, z^{(t)}) /\pi(z^{(t)})$ in Eq.~\eqref{eq: conditional p from joint and marginal} cancels the partition functions $Z_{\phi}$. So we are left with:
\begin{align}
    \pi(x^{(t)} \vert z^{(t)}) = \frac{e^{-E_{\phi} (x^{(t)}, z^{(t)})}}{\sum_{x^{(t)}} e^{-E_{\phi} (x^{(t)}, z^{(t)})}} = \frac{e^{-E_{\phi} (x^{(t)}, z^{(t)})}}{Z_{\phi} (z^{(t)})},
\end{align}
using the shorthand defined in Eq.~\eqref{eq: partial partition functions HMM}.

As for any latent variable model, we will have to trace out latent variables to obtain the marginal distribution over our data. The optimization objective is:
\begin{equation}
    \log \pi(x^{(1:T)}) = \log \sum_{z^{(1:T)}} \pi(x^{(1:T)}, z^{(1:T)}).
\end{equation}

After taking the gradient with respect to any of the parameters, we obtain the well-known form from latent variable EBMs
\begin{align}
    \nabla \log \pi(x^{(1:T)}) = \frac{1}{\pi(x^{(1:T)})} \nabla \sum_{z^{(1:T)}} \frac{e^{-E_{\pi}(z^{(1)})}}{Z_{\pi}}  \prod_{t=1}^T \frac{e^{-E_{\phi}(x^{(t)}, z^{(t)})}}{Z_{\phi}(z^{(t)})}  \prod_{t=2}^T \frac{e^{-E_{\theta}(z^{(t)}, z^{(t-1)})}}{Z_{\theta}(z^{(t-1)})}.
\end{align}

The main difference to the latent variable case, is that the partition functions $Z_{\phi}$ and $Z_{\theta}$ depend on $z^{(1:T)}$.

First, let's look at the gradient with respect to $\theta$. For simplicity we denote everything in the gradient function that does not depend on $\theta$ as $\pi_{\text{rest}}$.

\begin{align}
\begin{split}
    \nabla_\theta \log \pi(x^{(1:T)})  &= \frac{1}{\pi(x^{(1:T)})} \sum_{z^{(1:T)}} \pi_{\text{rest}} \nabla_{\theta} \prod_{t=2}^T \frac{e^{-E_{\theta}(z^{(t)}, z^{(t-1)})}}{Z_{\theta}(z^{(t-1)})} \\
    &= \frac{1}{\pi(x^{(1:T)})} \sum_{z^{(1:T)}} \pi_{\text{rest}} \sum_t \nabla_{\theta} \pi(z^{(t)} \vert z^{(t-1)}) \prod_{s \neq t} \pi(z_s \vert z_{s-1}) \\ 
    &= \frac{1}{\pi(x^{(1:T)})} \sum_{z^{(1:T)}} \pi_{\text{rest}} \sum_t \prod_{s \neq t} \pi(z_s \vert z_{s-1})  \\ 
    & \left[ \frac{-\nabla_{\theta} E_{\theta}(z^{(t)}, z^{(t-1)}) e^{-E_{\theta}(z^{(t)}, z^{(t-1)})} Z_{\theta}(z^{(t-1)}) - e^{-E_{\theta}(z^{(t)}, z^{(t-1)})} \nabla_{\theta} Z_{\theta}(z^{(t-1)})}{Z_{\theta}(z^{(t-1)}) ^ 2} \right] \\
    &= 
    \frac{1}{\pi(x^{(1:T)})} \sum_{z^{(1:T)}} \pi_{\text{rest}} \sum_t \prod_{s \neq t} \pi(z_s \vert z_{s-1})   \\
    & \Biggl[ -\nabla_{\theta} E_{\theta}(z^{(t)}, z^{(t-1)}) \pi(z^{(t)} \vert z^{(t-1)})  - \pi(z^{(t)} \vert  z^{(t-1)}) \sum_{z'^{(t)}} - \nabla_{\theta} E_{\theta}(z'^{(t)}, z^{(t-1)}) \pi(z'^{(t)} \vert  z^{(t-1)}) \Biggr] \\
    & = 
    \frac{1}{\pi(x^{(1:T)})} \sum_{z^{(1:T)}} \pi_{\text{rest}} \prod_{s} \pi(z_s \vert z_{s-1})  \sum_t 
    \\
    & \left[ -\nabla_{\theta} E_{\theta}(z^{(t)}, z^{(t-1)}) - \sum_{z'^{(t)}} - \nabla_{\theta} E_{\theta}(z'^{(t)}, z^{(t-1)}) \pi(z'^{(t)} \vert  z^{(t-1)}) \right] \\
    & = 
    - \sum_{z^{(1:T)}} \pi(z^{(1:T)} \vert x^{(1:T)}) 
    \sum_t  \left[ \nabla_{\theta} E_{\theta}(z^{(t)}, z^{(t-1)}) - \sum_{z'^{(t)}}  \nabla_{\theta} E_{\theta}(z'^{(t)}, z^{(t-1)}) \pi(z'^{(t)} \vert  z^{(t-1)}) \right]
    \end{split}
\end{align}
Averaged over many data trajectories from a batch $\mathcal{B}$, we obtain:


\begin{equation}
\sum_{x^{(1:T)}\in\mathcal B} \nabla_\theta\log\pi(x^{(1:T)}) = \sum_{x^{(1:T)}\in\mathcal B} \mathbbm{E}_{z^{(1:T)}\sim \pi(\cdot\mid x^{(1:T)})} \Biggl[ \sum_{t=2}^T \Biggl(-\nabla_\theta E_\theta(z^{(t)},z^{(t-1)}) + \mathbbm{E}_{z'^{(t)}\sim \pi_\theta(\cdot\mid z^{(t-1)})} \left[ \nabla_\theta E_\theta(z'^{(t)},z^{(t-1)}) \right] \Biggr) \Biggr].
\end{equation}

Now, let's calculate the gradient with respect to $\phi$. For simplicity, we rewrite:
\begin{align}
    \prod_{t=1}^T \frac{e^{-E_{\phi}(x^{(t)}, z^{(t)})}}{Z_{\phi}(z^{(t)})}  = \frac{e^{- \sum_t E_{\phi}(x^{(t)}, z^{(t)})}}{\sum_{x^{(1:T)}} e^{- \sum_t E_{\phi}(x^{(t)}, z^{(t)})}} = \frac{e^{-\mathcal{E}_{\phi}(x^{(1:T)}, z^{(1:T)})}}{\mathcal{Z}_{\phi}(z^{(1:T)})},
\end{align}
where we denote $\mathcal{E}_{\phi}(x^{(1:T)}, z^{(1:T)}) = \sum_t E_{\phi}(x^{(t)}, z^{(t)})$ and $\mathcal{Z}_{\phi}(z^{(1:T)}) = \sum_{x^{(1:T)}} e^{- \sum_t E_{\phi}(x^{(t)}, z^{(t)})}$.
Taking the gradient of this returns:
\begin{align}
      \nabla_\phi \frac{e^{-\mathcal{E}_{\phi}(x^{(1:T)}, z^{(1:T)})}}{\mathcal{Z}_{\phi}(z^{(1:T)})}  =  \pi(x^{(1:T)} \vert z^{(1:T)}) \left ( -\nabla_\phi \mathcal{E}_{\phi}  - \frac{e^{-\mathcal{E}_{\phi}}}{\mathcal{Z}_{\phi}} \sum_{x^{(1:T)}} -\nabla_\phi \mathcal{E}_{\phi} \pi(x^{(1:T)} \vert z^{(1:T)}) \right),
\end{align}
where we used that
\begin{align}
    \pi(x^{(1:T)} \vert z^{(1:T)}) &= \frac{e^{-\mathcal{E}_{\phi}(x^{(1:T)}, z^{(1:T)})}}{\mathcal{Z}_{\phi}(z^{(1:T)})}.
\end{align}
Also, if we now denote 
\begin{align}
    \pi_{\text{rest}}  = \frac{1}{\pi(x^{(1:T)})} \sum_{z^{(1:T)}} \frac{e^{-E_{\pi}(z^{(1)})}}{Z_{\pi}}  \prod_{t=2}^T \frac{e^{-E_{\theta}(z^{(t)}, z^{(t-1)})}}{Z_{\theta}(z^{(t-1)})},
\end{align}
we have:
\begin{align}
    \pi_{\text{rest}} \pi(x^{(1:T)} \vert z^{(1:T)}) &= \sum_{z^{(1:T)}}\frac{\pi(z^{(1)}) \prod_{t=2}^T \pi(z^{(t)} | z^{(t-1)}) \pi(x^{(1:T)} \vert z^{(1:T)})}{\pi(x^{(1:T)})} \notag \\ &= \sum_{z^{(1:T)}} \frac{\pi(x^{(1:T)}, z^{(1:T)})}{\pi(x^{(1:T)})} = \sum_{z^{(1:T)}} \pi(z^{(1:T)} \vert x^{(1:T)}).
\end{align}
Using this we get:
\begin{align}
     \nabla_\phi \log \pi(x^{(1:T)})  =  - \sum_{z^{(1:T)}} \pi(z^{(1:T)} \vert x^{(1:T)}) \big( \nabla_\phi \mathcal{E}_{\phi}  - \sum_{x^{(1:T)}} \nabla_\phi \mathcal{E}_{\phi} \pi(x^{(1:T)} \vert z^{(1:T)}) \big).
\end{align}
The complete loss is an average over a batch $\mathcal{B}$ of trajectories $x^{(1:T)}$ that we sample from a data distribution, hence the complete gradient reads:
\begin{align}
     \sum_{x^{(1:T)} \in \mathcal{B}} \nabla_\phi \log \pi(x^{(1:T)})  &= -\sum_{x^{(1:T)} \in \mathcal{B}} \sum_{z^{(1:T)}} \pi(z^{(1:T)} \vert x^{(1:T)}) \big( \nabla_\phi \mathcal{E}_{\phi}  - \sum_{x'^{(1:T)}} \nabla_\phi \mathcal{E}_{\phi} \pi(x'^{(1:T)} \vert z^{(1:T)}) \big) \notag \\ &= - \sum_{x^{(1:T)} \in \mathcal{B}} \left \{ \mathbbm{E}_{z^{(1:T)} \sim \pi(\cdot \vert x^{(1:T)})} \left [ \nabla_\phi \mathcal{E}_{\phi} \right ]  - \mathbbm{E}_{\substack{z^{(1:T)} \sim \pi(\cdot \vert x^{(1:T)}) \\ x'^{(1:T)} \sim \pi(\cdot \vert z^{(1:T)})}} \left [ \nabla_\phi \mathcal{E}_{\phi} \right ]  \right \}.
\end{align}

\section{Crosstalk and scale calibration}\label{ap:crosstalk}

Two effects complicate our calibration goals: First, there is significant crosstalk between the two on-chip flux lines and the \tnode{} loops, meaning the flux induced by current through the barrier (tilt) line partially couples into the tilt (barrier) loop. Second, there is almost always some residual flux trapped in the loop causing an effective flux offset, and therefore moving the zero flux point to some non-zero line voltage.

We can compensate for these two effects by finding the affine relationship between the control voltages and the true flux coordinates~\cite{quintana2017superconducting, novikov2018exploring, dai2021calibration, khezri2021anneal}.
\begin{equation}
\begin{bmatrix}
\phi_\text{bar} \\
\phi_\text{tilt}
\end{bmatrix} =
\begin{bmatrix}
m_{11} & m_{12} \\
m_{21} & m_{22}
\end{bmatrix}
\begin{bmatrix}
V_\text{bar} \\
V_\text{tilt}
\end{bmatrix}
+\begin{bmatrix}
\phi_\text{bar}^0 \\
\phi_\text{tilt}^0
\end{bmatrix}
\label{eq:voltage-flux-transformation}
\end{equation}
where $V_\text{bar}$ ($V_\text{tilt}$) is the control voltage applied to the barrier (tilt) line, and $\phi_\text{bar}^0$ ($\phi_\text{tilt}^0$) is the offset flux in the barrier (tilt) loop that is set during the normal-superconducting transition. The parameters $m_{ij}$ can be found by using the periodicity in $\phi_\text{bar}$ and $\phi_\text{tilt}$ of the readout resonator frequency. It is thus possible to measure the resonator frequency for a wide range of control voltages and correct for the crosstalk. The offset does not need to be compensated for, we simply need to know it. In experiments, we immediately compensate for crosstalk in the applied voltages and we rescale the compensated voltages to flux  in a post-processing step.

To facilitate the rescaling, we implement a simple method to measure and automatically detect the edges of the triangular features that appear in the $S_{21}$ map (Figure~\ref{fig:honeycomb}) via peak finding. We then fit the two edges of each triangle to lines and determine the location where they intercept, i.e., at the apex of the triangle. Because this triangle feature is repeated at regular interval on the $\phi_\text{bar}$ and $\phi_\text{tilt}$ flux map, given enough apex points, it is possible to linearly fit for the volt-to-flux scale and offset.

\begin{figure}[t] 
    \centering
    \includegraphics[width=0.6\linewidth]{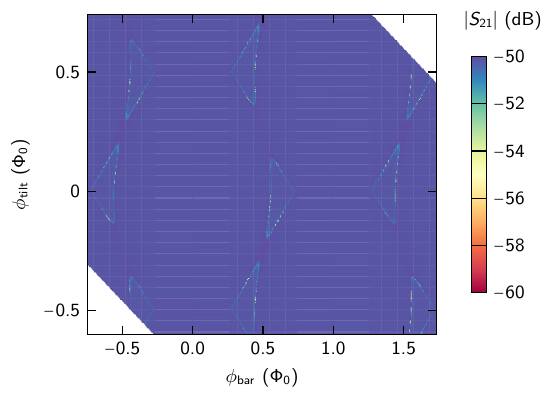} 
    \caption{Heatmap of $|S_{21}|$ data versus $\phi_\text{bar}$ and $\phi_\text{tilt}$ for readout frequency $f_\text{probe}=\qty{11.790}{\GHz}$. The mapping between control voltage and flux was done using \cref{eq:voltage-flux-transformation}. The data on the corners is missing due to voltage amplitude limits of the AWG.}
    \label{fig:honeycomb}
\end{figure}

The device displays an unexpected non-affine relationship between the flux and voltage in the temperature region between 60 and \qty{100}{\milli\kelvin}. We suspect that this is due to contamination of the base layer film, see \cref{ap:contamination}. Outside of this temperature region, we still observe some mild drift in the crosstalk and scale. We thus re-measure the triangle scale at each temperature point to ensure a proper calibration of the barrier flux.

\section{Relaxation time experiment}\label{ap:relax}

\begin{figure}[t]
    \includegraphics[width=0.45\linewidth]{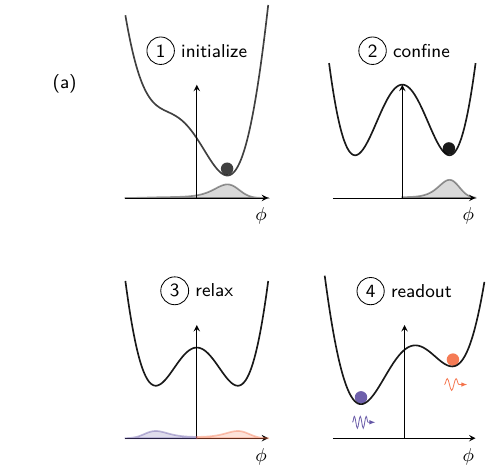} 
    \includegraphics[width=0.45\linewidth]{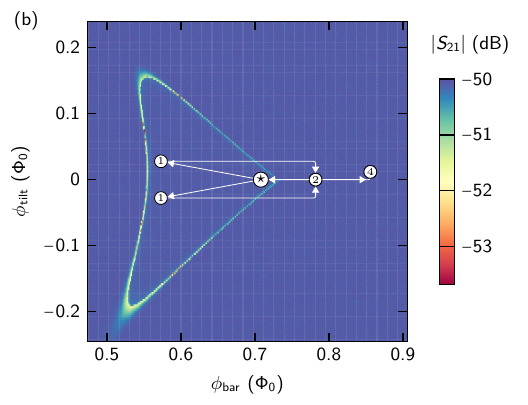} 
    \caption{(a) Experimental sequence of the relaxation experiment. The system is initialized in either the left or right well by tilting the potential and then raising the barrier. Next, the system is held at a bias point $({\phi}_{\text{bar}}, {\phi}_{\text{tilt}})$ and allowed to relax for a variable time $\tau$. The system is then brought to the readout point by raising the barrier and then introducing asymmetry through the tilt control. Lastly, the transmission line is probed to detect the \tnode{} position into the left or right well. (b) Control parameter trajectory in flux-space overlaid on $|S_{21}|$ transmission data. The star designates both the starting point of the sequence and the relaxation point.}
    \label{fig:exp-diagram}
\end{figure}

Figure~\ref{fig:exp-diagram} presents the procedure for the main relaxation experiment. It is similar to the initialization/readout experiment, but with an added wait time prior to fully raising the barrier for readout. We use the dc offset voltage functionality of our arbitrary waveform generator (AWG) to set the dc flux at the $(\phi_\text{bar},\phi_\text{tilt})$ point used during the relaxation time, denoted by the star in Figure~\ref{fig:exp-diagram}~(b), and use pulses to initialize the population and readout. After initialization \mbox{(step 1)}, but before going to the relaxation point, we raise the barrier, center the tilt flux, and wait \qtyrange[range-units=single, range-phrase=--]{1.75}{3}{\us}, \mbox{(step 2)}. This ensures that the population is fully confined in one well. Then we only need to lower the barrier to the relaxation point and wait some time $t$ \mbox{(step 3)}. Finally we raise the barrier for readout \mbox{(step 4)}.

We measure relaxation curves for a range of barrier settings, but always in the symmetric double-well configuration. To calibrate the tilt, we measure the population $P$ at a long wait time $t = \qty{1}{\ms}$ over a range of tilt flux settings and choose the value resulting in a 50-50 population. This calibration is done for every barrier flux point. We also ensure that the measured barrier range is as similar as possible for the different temperatures by choosing barrier voltages relative to the triangle apex identified for the scale calibration. We choose a voltage range between $-40$ and \qty{-10}{\mV} at the AWG, corresponding to $\phi_\text{bar}$ between 0.680 and 0.715 $\Phi_0$. To properly characterize the exponential decay of the population, we pick wait times logarithmically distributed between \qty{40}{\ns} and \qty{1}{\ms}. Finally, we measure these relaxation curves over temperatures ranging from \qty{9}{\milli\kelvin}, the base temperature, to \qty{250}{\milli\kelvin}. Due to the control issue appearing between 60 and \qty{100}{\milli\kelvin} forcing us to recalibrate the crosstalk compensation, we acquire two separate datasets, one for \qtyrange[range-units=single]{9}{60}{\milli\kelvin}, and another for \qtyrange[range-units=single]{105}{250}{\milli\kelvin}.

\section{Setup and sample}\label{ap:setup}

The sample is a chip comprising three independent \tnode{s} that are fabricated with evaporated aluminum on a silicon wafer. Each \tnode{} is coupled to a readout resonator. Additionally, the device hosts four coplanar waveguide resonators for design and fabrication verification. The three readout resonators are designed with frequencies between 11.8 and \qty{12}{\GHz}, spaced \qty{100}{\MHz} apart. Analogously, the four test resonators are designed between 11.4 and \qty{11.7}{\GHz}. We use the \tnode{} designed with the \qty{12}{\GHz} resonator for the experiments presented in this work, though because of the unaccounted coupling to the \tnode{}, the effective frequency ends up near \qty{11.8}{\GHz}. Flux pulses to control the dc-SQUID and the tilt of a \tnode{} are made via two independent coplanar waveguide (CPW) lines terminating in a short. Measurement is performed by means of a high-power readout scheme using a resonator inductively coupled to the \tnode{}, see \cref{ap:readout} for more details. Images of an identically-fabricated chip can be seen in Figure~\ref{fig:sample}. The package-to-chip connections are made with \qty{25}{\um} aluminum wire-bonds. The chip is wire-bonded to a commercially available package, the \product{QCage.24}, and shielded with a combination mu-metal and aluminum can made by QDevil.

\begin{figure}[ht]
  \centering
  \resizebox{0.5\linewidth}{!}{
  \tikzsetnextfilename{SampleImages}
  \includegraphics[width=\linewidth]{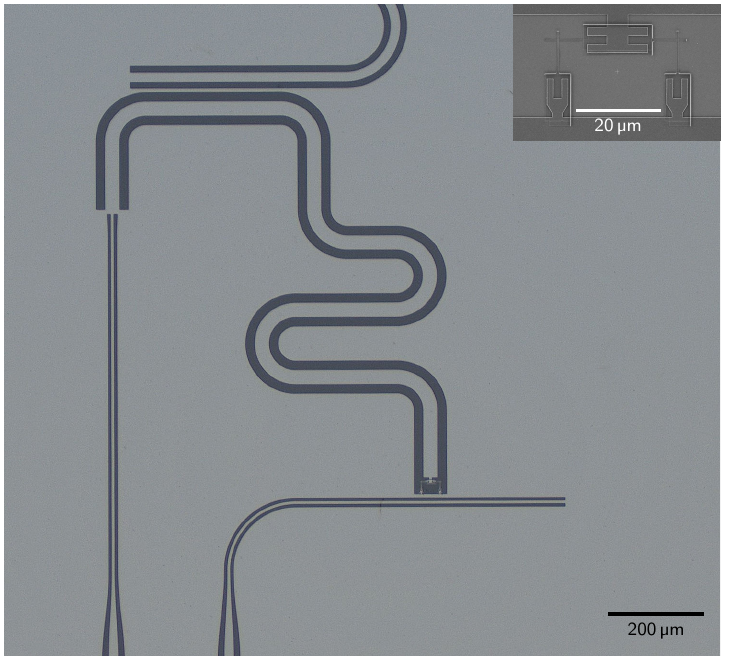} 
  }
  \caption{Optical image of the \tnode{}, with the tilt flux line on the left at the shorted end, readout resonator above (only the coupler is visible), and SQUID flux bias lines below. The $\lambda/4$ CPW segment of the \tnode{} effectively functions as a parallel LC-circuit, while the SQUID acts as a tunable nonlinear inductor. (Inset) Scanning electron microscope (SEM) image of the dc-SQUID, which is located at the open end of the CPW segment of the \tnode{}.}
  \label{fig:sample}
\end{figure}

The setup schematics of the dilution refrigerator, including instrument and wiring details can be found in Figure~\ref{fig:fridgewiring}. The two lines used to bias the \tnode{} and the readout input line are filtered with Eccosorb-based coaxial filters made by Quantum Microwave. The readout line output is filtered with a low-loss IR filter made by Sweden Quantum and a Low Noise Factory triple-junction isolator/circulator. We use the Keysight Quantum Control System to send microwave pulses for readout and DC pulses for flux control.

\section{Device fabrication}
\label{ap:fabrication}
The device is fabricated using an undoped high-resistivity (\textgreater20~k$\Omega \cdot$cm) silicon substrate using a two-layer process. Firstly, alignment markers are patterned using masked lithography and etched into the silicon with a Bosch etch using an \product{STS} plasma etcher. Following the etch and a solvent clean, the silicon wafer is cleaned using piranha solution, followed by an HF-dip to remove native oxides, after which it is transferred into an \product{Angstrom} electron-beam (e-beam) evaporator system to deposit the \qty{100}{\nm} aluminum base layer that makes up the ground plane and large-scale features. These features are patterned with a 100~kV e-beam lithography system \product{EBPG5200} into \product{ZEP520A} e-beam resist and developed with \product{ZED-N50}. Consequently, the pattern is plasma-etched with an \ce{Ar}/\ce{Cl2}/\ce{BCl_3} chemistry in an \product{STS} etcher. After stripping the remaining resist with 1-methyl-2-pyrrolidinone (NMP), the wafer is diced into $10\times 10$ $\text{mm}^2$ chips. The Manhattan-style Josephson junctions~\cite{potts2001novel, muthusubramanian2023wafer} are patterned with e-beam lithography in a PMMA/MMA (\qty{780}{\nm}/\qty{430}{\nm}) bilayer after dicing. Development was done cold with an IPA:\ce{H2O} (9:1) solution. The \ce{Al}/\ce{AlOx}/\ce{Al} junctions are then deposited at a 45$^\circ$ angle in a \product{Plassys} e-beam evaporator, using a double-angle technique. This evaporation was performed directly from the hearth, without the use of a crucible liner. We perform liftoff using NMP and sonication in acetone and isopropyl alcohol. Lastly, the sample is wirebonded to a printed circuit board and packaged.

\section{Hamiltonian fit}
\label{ap:hamiltonian-fit}

To determine the \tnode{} parameters, we perform a Hamiltonian fit to measured readout resonator frequency data. We use the \emph{1D no caps} model presented in Ref.~\cite{quintana2017superconducting}. This model, based on a Born-Oppenheimer approximation, accounts for the inline inductances and capacitances in the dc-SQUID, along with potential Josephson junction asymmetry. It is given by
\begin{align}
H_{\text{BO}} = \: & \frac{q^2}{2C_\text{eff}} + \frac{1}{2L} (\phi - \phi_{\text{tilt}})^2 \\
& - E_{J_1} \cos_{\beta_1} (\phi + \tfrac{1}{2}\phi_{\text{bar}}) - E_{J_2} \cos_{\beta_2}(\phi - \tfrac{1}{2}\phi_{\text{bar}}), \nonumber
\end{align}
where $C_{\text{eff}} = C + C_1 + C_2$ incorporates the inline capacitances of the two junctions that are designated by the subscript $i \in \{1,2\}$, $L$ is the inductance of the main loop, $E_{J_1}$ and $E_{J_2}$ are the Josephson energies, and
\begin{equation}
    \cos_{\beta_i}(\phi) = 1 + \sum_{k \in \mathbb{N}} \frac{2J_k(k \beta_i)}{k^2\beta_i} \left (\cos \left (\tfrac{2\pi}{\Phi_0} k\phi \right )-1 \right )
\end{equation}
where $\beta_i \equiv E_{J_i}/E_{L_i}$ is the fraction between junction $i$'s Josephson energy and its inductive energy $E_{L_i} \equiv \tfrac{\Phi_0^2}{8\pi^2L_i}$ associated with the inline inductance $L_i$, and $J_k(x)$ is the $k$th Bessel function. The junction asymmetry is characterized by  $\chi = (I_{c1}-I_{c2})/(I_{c1}+I_{c2})$, with $I_c = I_{c1} + I_{c2}$.

We add the Hamiltonian of the readout resonator, along with an inductive coupling between the \tnode{} and resonator. The resonator Hamiltonian is given by
\begin{equation}
    H_\mathrm{r} = \frac{q_\mathrm{r}^2}{2C_\mathrm{r}} + \frac{\phi_\mathrm{r}^2}{2L_\mathrm{r}}, 
\end{equation}
where $q_\mathrm{r}$ and $\phi_\mathrm{r}$ are the conjugate charge and flux of the resonator, $C_\mathrm{r}$ and $L_\mathrm{r}$ are the corresponding capacitance and inductance. The coupling term is written as
\begin{equation}
    H_\mathrm{c} = -\frac{M}{\alpha L_\mathrm{r}L} \phi_\mathrm{r}\phi
\end{equation}
where $M$ is the mutual inductance between the resonator and the main loop of the \tnode{}, and $\alpha = 1 - {M^2}/(L_\mathrm{r}L)$.

We calculate the theoretical resonator frequency in the quantum ground state by diagonalizing the $\phi$-space discretized Hamiltonian for all measured combinations of $\phi_{\text{bar}}$ and $\phi_{\text{tilt}}$, and fit with the Levenberg–Marquardt algorithm. Due to the large number of free parameters (11), the lack of direct measurement of the \tnode{} frequency, and potential residual crosstalk or control imperfections, the fitting results encompass a very large region of the $L$, $C$, and $I_c$ parameter space, over which equally-good convergence is achieved. To bypass this problem, we choose to fix the capacitance to a realistic value determined via simulations. With this added constraint, we achieve a robust convergence over the 10 other parameters.

The parameters of the device used in the experiment are specified in \cref{tbl:deviceparams}. The resonator frequencies measured at 10~mK are shown in Figure~\ref{fig:resonator-frequencies} together with the frequencies based on the Hamiltonian fit.

\begin{table}[hb]
\centering
\caption{Device parameters for the \tnode{} used in the experiment. $M$ is the mutual inductance between the \tnode{} and its readout resonator. The Hamiltonian parameters are determined via a fit of the resonator frequency over a large range of bias voltages. The uncertainty of the $C$ and $L$ \tnode{} parameters should be considered fairly large since we only fit the resonator frequency spectrum, and not that of the \tnode{}. Note that the resonator frequency shown is that of the bare resonator.}
\label{tbl:deviceparams}
\begin{minipage}{2in}
\begin{ruledtabular}
\begin{tabular}[t]{l r}
  Parameter & Value \\
  \hline\\[-3mm]
  $C$ \footnotesize(\si{\femto\farad}) & $120$ \\
  $L$ \footnotesize(\si{\pico\henry}) & $750$ \\
  $L_{1,2}$ \footnotesize(\si{\pico\henry}) & $50$ \\
  $I_c$ \footnotesize(\si{\micro\ampere}) & $0.997$ \\
  $\chi$ & 0.0104 \\
  $f_\text{res}$ \footnotesize(\si{\giga\hertz}) & $11.804856$ \\
  $M$ \footnotesize(\si{\pico\henry}) & $24.7$ \\
\end{tabular}
\end{ruledtabular}
\end{minipage}

\end{table}

\begin{figure*}[t] 
    \centering
    \includegraphics[width=0.8\linewidth]{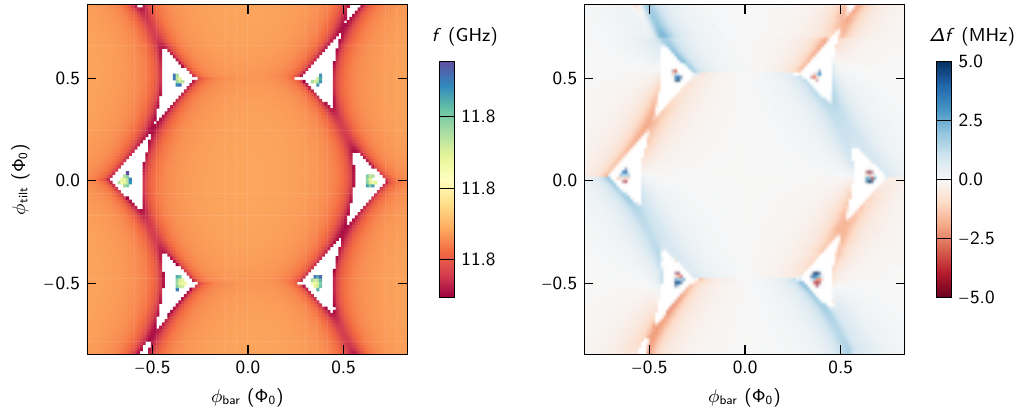}
    \caption{Readout resonator frequency as function of $\phi_\text{bar}$ and $\phi_\text{tilt}$. (left) Extracted resonator frequency from $S_{21}$ transmission measurements. (right) Residuals between the experimental data and the Hamiltonian fit with the parameters listed in \cref{tbl:deviceparams}.}
    \label{fig:resonator-frequencies}
\end{figure*}

\section{Readout}
\label{ap:readout}

Readout is performed by utilizing the dispersive shift allowing the system to be projected in the left or right well, see Ref.~\cite{quintana2017superconducting}. To this end, the \tnode{} is inductively coupled to a coplanar $LC$-oscillator, the readout resonator. Occupancy in either well of the \tnode{} will induce a shift in the readout resonator frequency. These two dispersive shifts differ when an asymmetry between the two wells is introduced through the tilt control. The readout resonator is capacitively coupled to the transmission line from which it can be probed. The \tnode{} state can be inferred using homodyne detection of the output field at an appropriate readout frequency $f_\text{probe}$. We use a readout pulse with a flattop Gaussian shape of \qty{5}{\us}, with a \qty{200}{\ns} risetime. After down-conversion and time averaging, the measured in-phase and quadrature (IQ) points are used to discriminate the two states. For the experiments presented in the main text, we calibrate the readout by preparing the flux particle in the left (right) well, pulse to an asymmetric flux coordinate with sufficiently deep wells and finally probe the transmission line. We find a triplet $(V_\text{bar}, V_{\text{tilt}}, f_\text{r})$ of control voltages and readout frequency that optimizes the visibility, defined as $\mathcal{V} = P_{L|L} + P_{R|R} - 1$, with $P_{L|L}$ being the proportion of measurement shots classified as $L$ when the \tnode{} was prepared in the $L$ state and equivalently for $P_{R|R}$. When characterizing readout at optimized settings over 1000 single-shot measurements, we find the proportion in either target well with a $100.0\%$ visibility. Figure~\ref{fig:readout-calibration-resonator-tracking} shows the visibility against the barrier and tilt voltage, where the readout frequency is selected per barrier voltage.

\begin{figure}[ht] 
    \centering
    \includegraphics[width=0.48\linewidth]{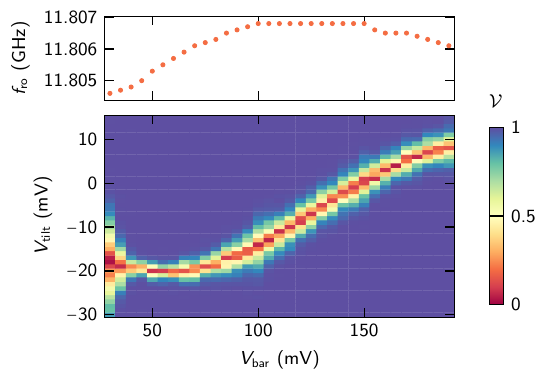}
    \caption{(bottom) Visibility of the readout procedure at \qty{120}{\milli\kelvin} for variable flux and tilt control voltages where the readout frequency is chosen per barrier voltage (top). At the tilt-symmetric curve the visibility reduces to zero as the dispersive shifts of both wells become nearly identical. This curve is not a straight line both due to nonlinear crosstalk caused by junction asymmetry~\cite{khezri2021anneal, quintana2017superconducting}.}
    \label{fig:readout-calibration-resonator-tracking}
\end{figure}

\section{Non-affine flux-voltage relationship}
\label{ap:contamination}

In the temperature region between \qty{60}{\milli\kelvin} and \qty{100}{\milli\kelvin} we observed a non-affine relationship between the applied control voltages and the flux coordinates, see Figure~\ref{fig:contamination}. This inhibits performing experiments in this temperature region as we do not have a faithful mapping between control voltages and fluxes. Below and beyond this range the voltage-flux relationship was found to be adequately characterized by \cref{eq:voltage-flux-transformation}, though still with a small drift that must be corrected for.

We suspect that this unexpected behavior could be due to Boron contamination that was introduced from an intermetallic crucible in the evaporator during the base layer deposition. The Josephson junction deposition was performed in a different evaporator from the hearth without the use of a crucible liner, see \cref{ap:fabrication}.

\begin{figure}[ht!] 
    \centering
    \resizebox{0.48\linewidth}{!}{
    \includegraphics[width=\linewidth]{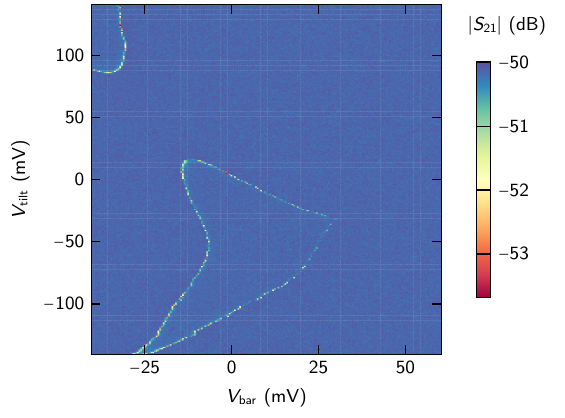} 
    }
    \caption{Non-affine voltage-flux relationship at \qty{80}{\milli\kelvin}.}
    \label{fig:contamination}
\end{figure}

\clearpage

\begin{figure*}[p]
    \centering
    \includegraphics[width=0.7\linewidth]{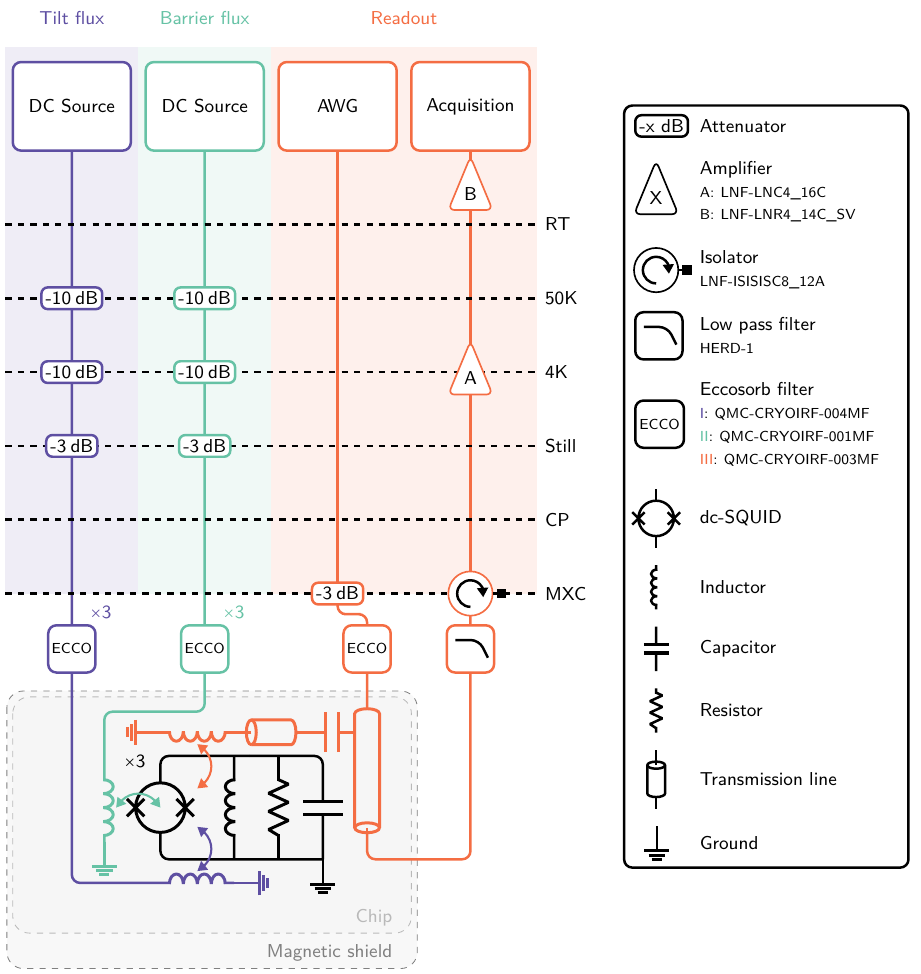} 
    \caption{Experimental setup diagram of the Bluefors LD400 dilution refrigerator and the wiring. The refrigerator has a room temperature (RT), 50K, 4K, still and mixing chamber (MXC) flange. The cold plate (CP) is situated between the still and mixing chamber flanges. Placement on a flange means that the component is thermally anchored to it. The triple-junction isolator is mounted on top of the MXC flange and includes a mu-metal shield. The cold amplifier is mounted to the bottom of the 4K flange. The sample chip is packaged and placed inside a magnetic shield that is mounted underneath the MXC flange. The chip hosts three superconducting \tnode{}s, each with dedicated tilt and barrier flux lines that are controlled by DC sources at room temperature. Mutual inductances are indicated by an arc between an inductor and the relevant loop. Readout of a \tnode{} is performed through its inductively coupled readout resonator that is capacitively coupled to a single shared transmission line. The readout pulses are supplied by an arbitrary waveform generator (AWG) that has an integrated acquisition system for down-conversion and digitization of the transmitted signal. Four coplanar waveguide verification resonators that are also capacitively coupled to the transmission line are omitted from the diagram.}
    \label{fig:fridgewiring}
\end{figure*}

\end{document}